\documentclass[10pt,journal,compsoc]{IEEEtran}

\ifCLASSOPTIONcompsoc
\else
\fi

\ifCLASSINFOpdf

\else

\fi

\ifCLASSOPTIONcompsoc
\usepackage[normalsize,sf]{subfigure}
\else
\usepackage[footnotesize]{subfigure}
\fi

\usepackage{times}
\usepackage{epsfig}
\usepackage{graphicx}
\usepackage{amsmath}
\usepackage{amssymb}
\usepackage{color}
\usepackage{epsfig}
\usepackage{graphicx}
\usepackage{amsmath}
\usepackage{amssymb}

\usepackage{amsmath}
\usepackage{gensymb}

\usepackage[utf8]{inputenc} 
\usepackage[T1]{fontenc}    
\usepackage{url}            
\usepackage{booktabs}       
\usepackage{amsfonts}       
\usepackage{nicefrac}       
\usepackage{microtype}      
\usepackage{color}
\usepackage{multirow}
\usepackage{booktabs}
\usepackage{siunitx}
\usepackage{comment}
\usepackage[hidelinks]{hyperref}
\usepackage{xcolor}
\usepackage{xspace}

\makeatletter
\DeclareRobustCommand\onedot{\futurelet\@let@token\@onedot}
\def\@onedot{\ifx\@let@token.\else.\null\fi\xspace}

\def\eg{\emph{e.g}\onedot} 
\def\ie{\emph{i.e}\onedot}

\def\etal{\emph{et al}\onedot}
\makeatother

\newcommand{\ourmodelshortoriginal}{GeoNet}
\newcommand{\ourmodel}{Geometric Neural Network with Edge-Aware Refinement}
\newcommand{\ourmodelshort}{GeoNet++}

\begin{document}

\title{GeoNet++: Iterative Geometric Neural Network with Edge-Aware Refinement for Joint Depth and Surface Normal Estimation}

\author{Xiaojuan Qi,~
        Zhengzhe Liu,~
        Renjie Liao,~
        Philip H. S. Torr,~
        Raquel Urtasun,~
        and~Jiaya~Jia
\IEEEcompsocitemizethanks{
\IEEEcompsocthanksitem X. Qi  is with the Department
of Electrical and Electronic Engineering, University of Hong Kong, Hong Kong.\protect\\
\vspace{-0.1in}
\IEEEcompsocthanksitem Z. Liu is with DJI corporation, Shenzhen, China.\protect\\
\vspace{-0.1in}
\IEEEcompsocthanksitem R. Liao and R. Urtasun are with Uber ATG, University of Toronto, and Vector Institute, Toronto, Canada.\protect\\
\vspace{-0.1in}
\IEEEcompsocthanksitem P. Torr is with the Department of Engineering Science, University of Oxford, Oxford, United Kingdom.\protect\\
\vspace{-0.1in}
\IEEEcompsocthanksitem J. Jia is with the Department of Computer Science and Engineering, The Chinese University of Hong Kong, Hong Kong.\protect\\
\vspace{-0.1in}
\IEEEcompsocthanksitem  X. Qi and Z. Liu share the first-authorship.\protect
}
}

\markboth{IEEE TRANSACTIONS ON PATTERN ANALYSIS AND MACHINE INTELLIGENCE}%
{Shell \MakeLowercase{\textit{et al.}}: Bare Demo of IEEEtran.cls for Computer Society Journals}

\IEEEcompsoctitleabstractindextext{%
\begin{abstract}
In this paper, we propose a geometric neural network with edge-aware refinement (\ourmodelshort) to jointly predict both depth and surface normal maps from a single image. Building on top of two-stream CNNs, 
{\ourmodelshort} captures the geometric relationships between depth and surface normals with the proposed depth-to-normal and normal-to-depth modules. In particular, the ``depth-to-normal'' module exploits the least square solution of estimating surface normals from depth to improve their quality, while the ``normal-to-depth'' module refines the depth map based on the constraints on surface normals through kernel regression. Boundary information is exploited via an edge-aware refinement module.
{\ourmodelshort} effectively predicts depth and surface normals with strong 3D consistency and sharp boundaries resulting in better reconstructed 3D scenes. Note that {\ourmodelshort} is generic and can be used in other depth/normal prediction frameworks to improve the quality of 3D reconstruction and pixel-wise accuracy of depth and surface normals. 
Furthermore, we propose a new 3D geometric metric (3DGM) for evaluating depth prediction in 3D. In contrast to current metrics that focus on evaluating pixel-wise error/accuracy, 3DGM measures whether the predicted depth can reconstruct high-quality 3D surface normals. This is a more natural metric for many 3D application domains. Our experiments on NYUD-V2 \cite{silberman2012indoor} and KITTI \cite{geiger2013vision} datasets verify that {\ourmodelshort} produces fine boundary details, and the predicted depth can be used to reconstruct high-quality 3D surfaces. Code has been made publicly available.
\end{abstract}

}

\maketitle

\IEEEdisplaynotcompsoctitleabstractindextext

%
\IEEEpeerreviewmaketitle

\IEEEraisesectionheading{\section{Introduction}\label{sec:introduction}}

\IEEEPARstart{W}{e} tackle the important problem of jointly estimating depth and surface normals from a single RGB image. This 2.5D geometric information is beneficial to various computer vision tasks, including structure from motion (SfM), 3D reconstruction, pose estimation, object recognition, and scene classification. Depth and surface normals are typically employed in many application domains that require 3D understanding of the scene, {\eg}, robotics, virtual reality, and human-computer interactions, to name a few.

\begin{figure}[t]
\centering
\includegraphics[width=0.5\textwidth]{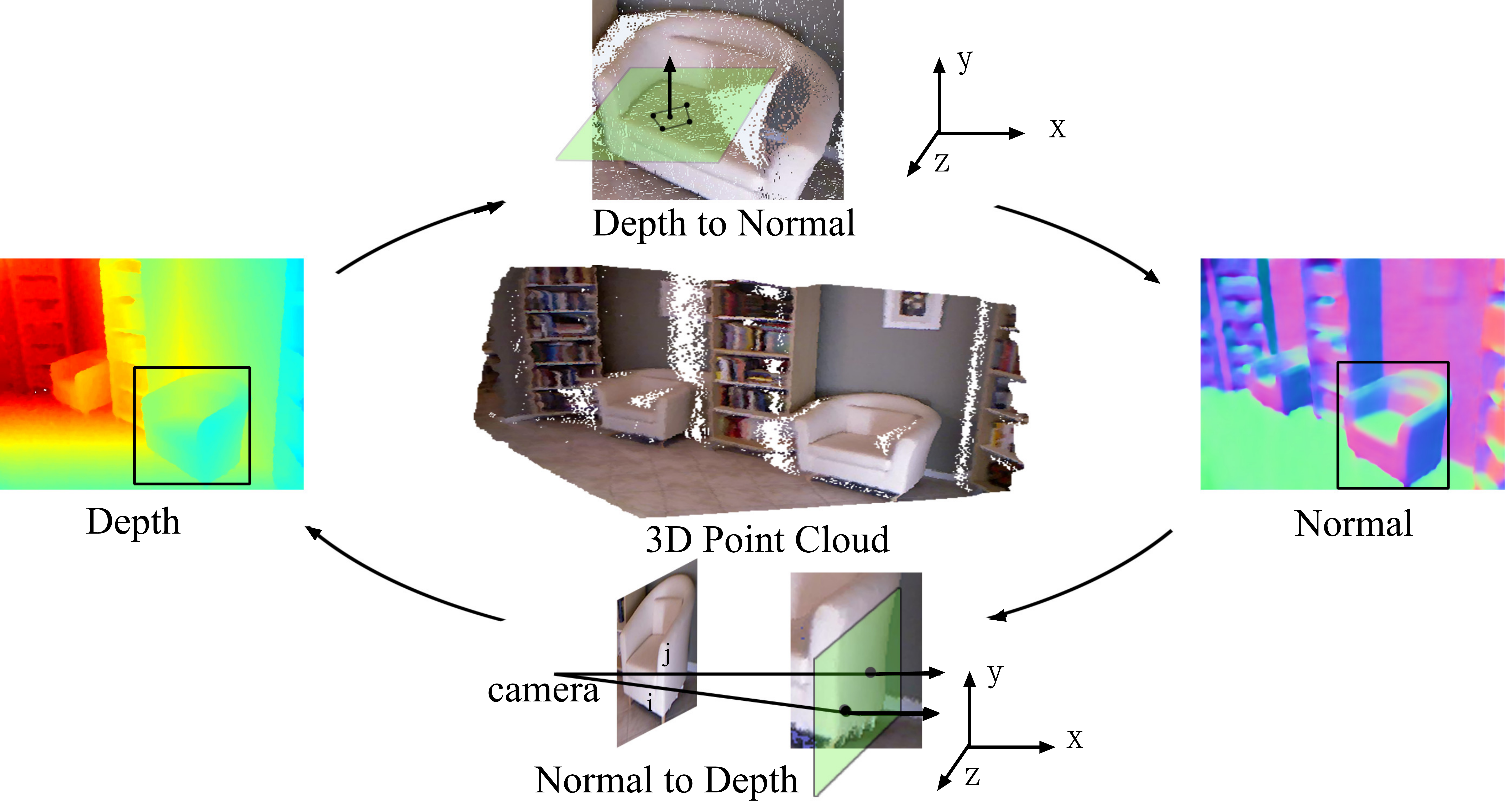}
\caption{The geometric relationship between depth and surface normals. The point cloud is obtained by casting depth values into 3D via the pinhole camera model. Surface normals are estimated from the point cloud by solving a system of linear equations; depth is constrained by the local plane determined by neighboring points and their surface normals.} 
\label{fig:intro}
\end{figure}

\begin{figure*}
\centering
\begin{tabular}{@{\hspace{0.1mm}}c@{\hspace{0.1mm}}c@{\hspace{0.1 mm}}c@{\hspace{0.1 mm}}c}
\includegraphics[width=0.24\linewidth]{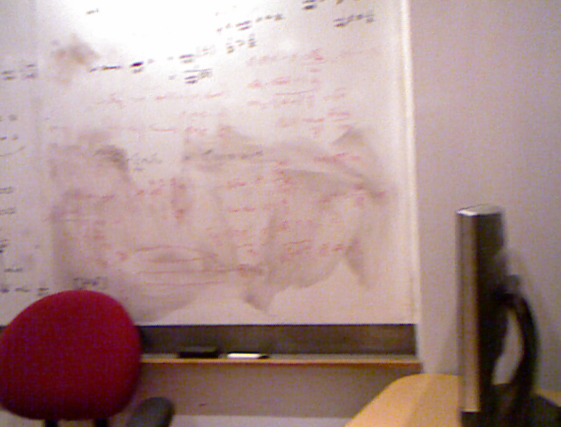} &
\includegraphics[width=0.24\linewidth]{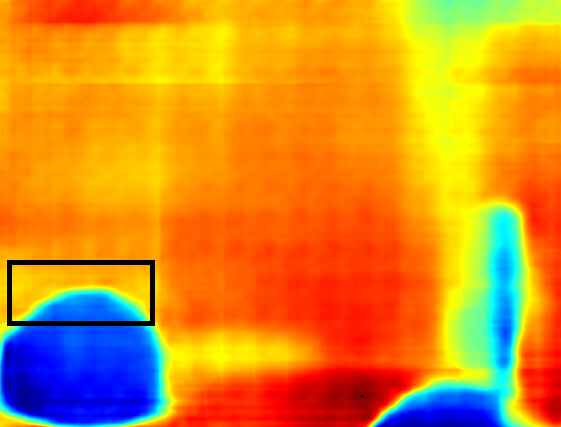}&
\includegraphics[width=0.24\linewidth]{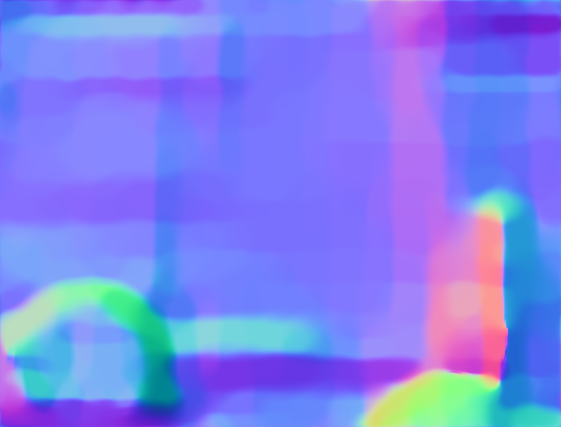}&
\includegraphics[width=0.24\linewidth]{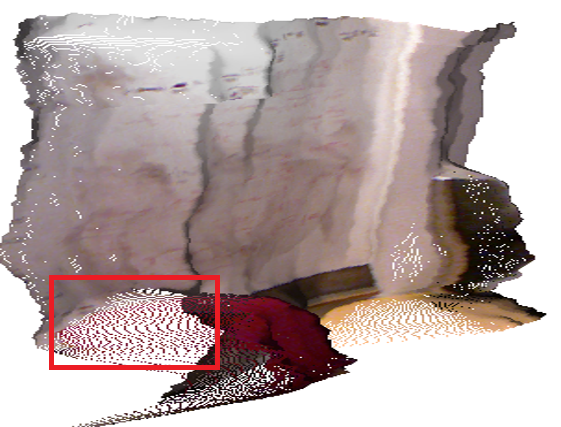}\\
\footnotesize{ (a) Input } &\footnotesize{(b) Depth (\textcolor{black}{DORN~\cite{fu2018deep}})} &\footnotesize{(c) Normal (\textcolor{black}{DORN~\cite{fu2018deep}})}&\footnotesize{ (d) 3D (\textcolor{black}{DORN~\cite{fu2018deep}})}\\
\includegraphics[width=0.24\linewidth]{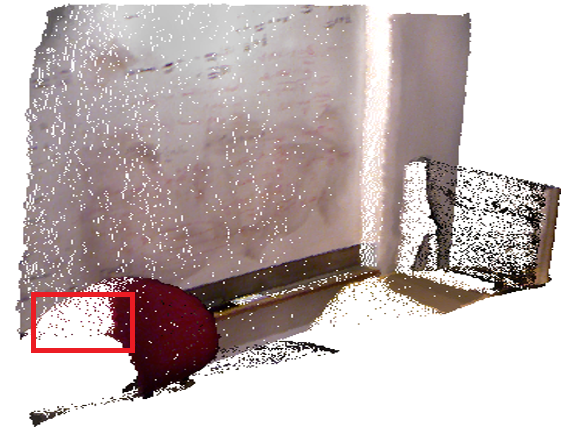}&
\includegraphics[width=0.24\linewidth]{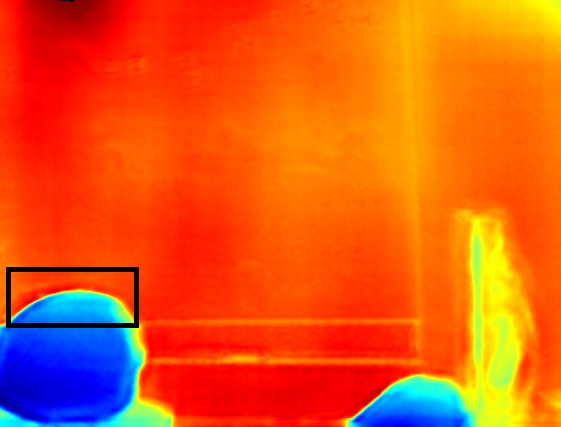} &
\includegraphics[width=0.24\linewidth]{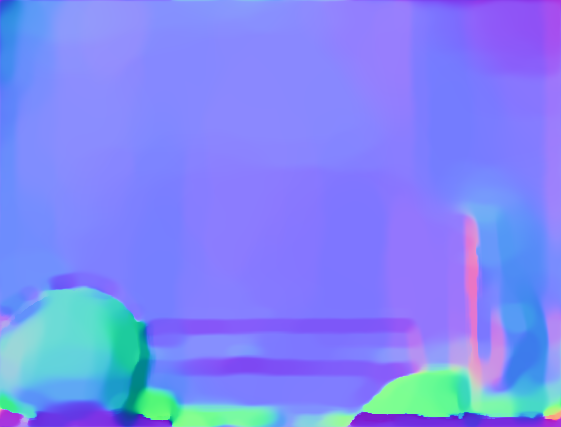} &
\includegraphics[width=0.24\linewidth]{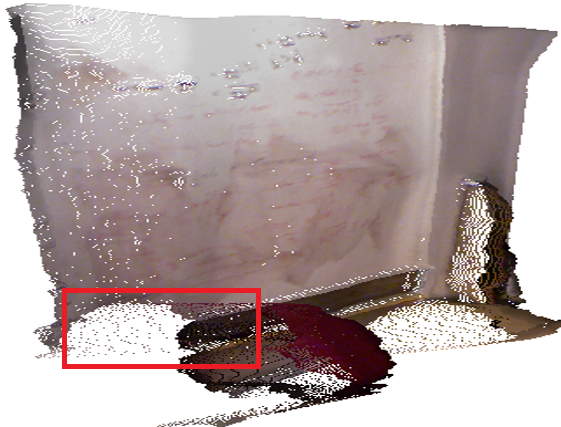}\\
\footnotesize{ (e) 3D (GT) } &\footnotesize{(f) Depth (Ours)} &\footnotesize{(g) Normal (Ours)}&\footnotesize{ (h) 3D (Ours)}\\

\end{tabular}
\caption{Visual illustrations of depth/normal maps and the reconstructed 3D point cloud.
(a) is the input image. (b) is the depth map from a state-of-the-art approach \textcolor{black}{DORN~\cite{fu2018deep}}. (c) is the surface normal derived from (b).
(d) is the corresponding point cloud visualization of (b). (e) shows the ground-truth point cloud. (f) is the depth map from our approach. (g) shows the surface normal derived from (f). (h) shows the corresponding point cloud visualization of (f). \textcolor{black}{ The normal maps (DORN~\cite{fu2018deep} and Ours) are computed from the corresponding point cloud shown in (d) and (h) respectively using the least square fitting provided in~\cite{silberman2012indoor} followed by TV-denoising.}}
\label{fig:vis-teaser}
\end{figure*}

There exists a large body of work on depth ~\cite{saxena2006learning,liu2010single,eigen2014depth,eigen2015predicting,liu2014discrete,wang2015towards,roy2016monocular,laina2016deeper,liu2016learning,xu2017multi,li2015depth} and surface normal estimation~\cite{eigen2015predicting,wang2015designing,bansal2016marr,bansal2017pixelnet,li2015depth} from a single image.
Most previous methods independently perform depth and normal estimation, potentially leading to inconsistent predictions and poor 3D surface reconstructions. 
As shown in Fig.~\ref{fig:vis-teaser} (d)-- wall regions, the predicted depth map could be distorted in planar regions. 
Utilizing the fact that the surface normal does not change in such regions could help denoise planar surfaces.

This motivates us to exploit the geometric relationship between depth and surface normals.
We use the example in Fig.~\ref{fig:intro} as an illustration. On the one hand, the surface normal is determined by the tangent plane to the 3D points, which can be estimated from their depth; on the other hand, depth is constrained by the local surface of the tangent plane determined by the surface normal. 
Several approaches have tried to incorporate geometric relationships into traditional models via hand-crafted features~\cite{saxena2006learning,barron2015shape}.
However, little research has been done in the context of neural networks. This is the focus of our paper.  

One possible design is to build a convolutional neural network (CNN) to directly learn such geometric relationships from data.
However, our experiments in Sec.~\ref{sect:naive-solution} demonstrate that existing CNN architectures ({\eg}, VGG-16) can not predict good normals from depth. We found that the training always converges to very poor local minima, even with carefully tuned architectures and hyperparameters.
Another challenge stems from pooling operations and large receptive fields, which makes current architectures perform poorly near object boundaries. 
We refer the reader to Fig.~\ref{fig:vis-teaser} (b) where the results are blurry on the black bounding box. 
This phenomenon is amplified when viewing the results in 3D. 
As shown in Fig.~\ref{fig:vis-teaser} (d), the points inside the red bounding box are scattered in 3D due to the blurry boundaries. 
It is therefore problematic for robotic applications where obstacle detection and avoidance are needed for safety.

The above facts motivate us to design a new architecture that explicitly incorporates and enforces 3D geometric constraints considering object boundaries. 
Towards this goal, we propose  {\ourmodel} ({\ourmodelshort}), which integrates geometric constraints and boundary information into a CNN. 
This contrasts with previous works~\cite{bansal2016marr,eigen2014depth,fu2018deep}, which focus on designing new network architectures~\cite{eigen2014depth} or loss functions more tailored for the task~\cite{fu2018deep}.

The overall system (see Fig.~\ref{fig:backbone}) has a two-stream backbone CNN, which predicts initial depth and surface normals from a single image respectively. With initial depth and surface normals, 
{\ourmodelshort} (see Fig.~\ref{fig:overview}) is utilized to incorporate geometric constraints by modeling depth-to-normal (see Fig.~\ref{fig:normal-depth} (a) -- (L)) and normal-to-depth (see Fig.~\ref{fig:normal-depth} (b) -- (L)) mapping, and introduce boundary information with edge-aware refinement (see Fig.~\ref{fig:normal-depth} (c)).
Our ``depth-to-normal'' module relies on least-square and residual sub-modules, while the ``normal-to-depth'' module updates the depth estimates via kernel regression.
Guided by the learned propagation weights, our ``edge-aware refinement module'' sharpens boundary predictions and smooths out noisy estimations.
Our framework enforces the final depth and surface normal prediction to follow the underlying 3D constraints, which directly improves 3D surface reconstruction quality.
Note that {\ourmodelshort} can be integrated into other CNN backbones for depth or surface normal prediction.
Importantly, the overall system can be trained end-to-end.

Our final contribution is a new geometric-related evaluation metric for depth prediction which measures the quality of the 3D surface reconstruction. This metric directly measures the local 3D surface quality by casting the predicted depth into 3D point clouds.
It is better correlated with the end goals of the 3D tasks.
Experimental results on NYUD-V2~\cite{silberman2012indoor} and KITTI~\cite{geiger2013vision} datasets show that our {\ourmodelshort} achieves decent performance, while being more efficient.

\vspace{0.05in}
\noindent \textbf{Difference from our Conference Paper:~~~}
This manuscript significantly improves the conference version \cite{qi2018geonet}: (i) we introduce an edge-aware propagation network to improve the prediction at boundaries, facilitating the generation of better point clouds; (ii) we develop an iterative scheme to progressively improve the quality of predicted depth and surface normals; (iii) we propose a new evaluation metric to measure 3D surface reconstruction accuracy; (iv) we conduct additional experiments and analysis on KITTI~\cite{geiger2013vision} dataset;
(v) we empirically show that {\ourmodelshort} can be incorporated into previous methods~\cite{eigen2014depth,xu2017multi,fu2018deep} to further improve the results especially the 3D reconstruction quality; (vi) from both qualitative and quantitative perspectives, our results are significantly better compared to~\cite{qi2018geonet}, especially in 3D metrics. 

The rest of the paper is organized as follows. Sec.~\ref{sect:relate_work} reviews the literature on depth and surface normal prediction. In Sec.~\ref{sect:model}, we elaborate on our {\ourmodelshort} model. 
We conduct experiments and show more detailed analysis in Sections~\ref{sect:experiments} -- \ref{sect:naive-solution}. We draw our conclusion in Sec.~\ref{sect:conclusion}.

\section{Related Work}\label{sect:relate_work}

The 2.5D geometry estimation from a single image has been intensively studied.
Previous works can be roughly divided into two categories based on whether deep neural networks have been used.

Traditional methods do not use deep neural networks and mainly focus on exploiting low-level image cues and geometric constraints. For example, \cite{torralba2002depth} estimates the mean depth of the scene by recognizing the structures presented in the image, and inferring the scene scale. Based on Markov random fields (MRF), Saxena {\etal} \cite{saxena2006learning} predicted a depth map given  hand-crafted features of a single image.
Vanishing points and lines are utilized in~\cite{hoiem2007recovering} for recovering the surface layout~\cite{schwing2013box}.
Liu {\etal}~\cite{liu2010single} leveraged predicted  semantic segmentation to incorporate geometric constraints. A scale-dependent classifier was proposed in \cite{ladicky2014pulling} to jointly learn semantic segmentation and depth estimation.
Shi {\etal}~\cite{shi2015break} showed that estimating defocus blur is beneficial for recovering the depth map. 
Favaro {\etal}~\cite{favaro2005geometric} proposed to learn a set of projection operators from blurred images, which are further utilized to estimate the 3D geometry of the scene from novel blurred images.
In~\cite{barron2015shape}, a unified optimization problem was formed aiming at recovering the intrinsic scene properties, {\eg}, shape, illumination, and reflectance from shading. Relying on specially designed features, above methods directly incorporate geometric constraints.

Many deep learning methods were recently proposed for single-image depth and/or surface normal prediction.
Eigen {\etal}~\cite{eigen2014depth} directly predicted the depth map by feeding the image to a CNN. Shelhamer {\etal}~\cite{shelhamer2015scene} proposed a fully convolutional network (FCN) to learn the intrinsic decomposition of a single image, which involves inferring the depth map as the first intermediate step.
Recently, Ma {\etal}~\cite{ma2018single} incorporated the physical rule in multi-image intrinsic decomposition for single image intrinsic decomposition.
In~\cite{eigen2015predicting}, a unified coarse-to-fine hierarchical network was adopted for
depth/normal prediction.
Continuous conditional random fields (CRFs) were proposed in~\cite{xu2017multi} to fuse information derived from CNN outputs.
Fu~{\etal}~\cite{fu2018deep} introduced ordinal regression loss to help the optimization process and achieve better performance.
In~\cite{liu2016learning}, continuous CRFs were built on top of CNN to smooth super-pixel-based depth prediction.
For predicting single-image surface normals, Wang {\etal}~\cite{wang2015designing} incorporated local, global, and vanishing point information in designing the network architecture.
Reconstruction loss has been exploited in~\cite{garg2016unsupervised} for unsupervised depth estimation from a single image.
Following work~\cite{godard2017unsupervised} introduced left-right consistency constraint.
A skip-connected architecture has also been proposed in \cite{bansal2016marr} to fuse hidden representations of different layers for surface normal estimation.

All these methods regard depth and surface normal predictions as independent tasks, thus ignoring their basic geometric relationship that also influences the quality of the reconstructed surface.
Recently, a few works~\cite{wang2016surge,xu2018pad,zhang2018joint,wang2015towards} jointly reason multiple tasks. Wang~{\etal}~\cite{wang2016surge} designed CRFs to fuse semantic segmentation prediction and depth estimation. Xu {\etal}~\cite{xu2018pad} proposed a hierarchical framework to first predict depth, surface normal, edge maps, and semantic segmentation, and then fuse them together for the final depth and semantic map prediction. Zhang~{\etal}~\cite{zhang2018joint} learned depth prediction and semantic segmentation with the recursive refinement to progressively refine the predicted depth and semantic segmentation.
All these approaches focus on modifying either CNN architectures or loss functions to make the model better fit the data without explicitly considering the geometric property.
In contrast, our approach explicitly incorporates geometric constraints by designing edge-aware geometric modules, which is orthogonal to previous works.

The most related work to ours is that of~\cite{wang2016surge}, which has a CRF with a 4-stream CNN, considering the consistency of predicted depth and surface normal in planar regions. Nevertheless, it may fail when planar regions are uncommon in images. Moreover, the iterative inference in CRF and the Monte Carlo sampling strategy make the approach suffer from the heavy computational cost.
In comparison, our {\ourmodelshort} exploits the geometric relationship between depth and surface normal for {\it general} situations without making any planar or curvature assumption. Our model is more efficient compared to the iterative inference of CRF in~\cite{wang2016surge}.

Recently, the depth-normal consistency was utilized for depth completion~\cite{zhang2018deep} from a single image and unsupervised depth-normal estimation~\cite{yang2018unsupervised} from monocular videos.
In this paper, we focus on deploying geometric constraints to improve depth and surface normal estimation from a single image and analyzing its influence on 3D surface reconstruction.

\section{\ourmodelshort}\label{sect:model}
\begin{figure*}[t]
\centering
\includegraphics[width=0.9\textwidth]{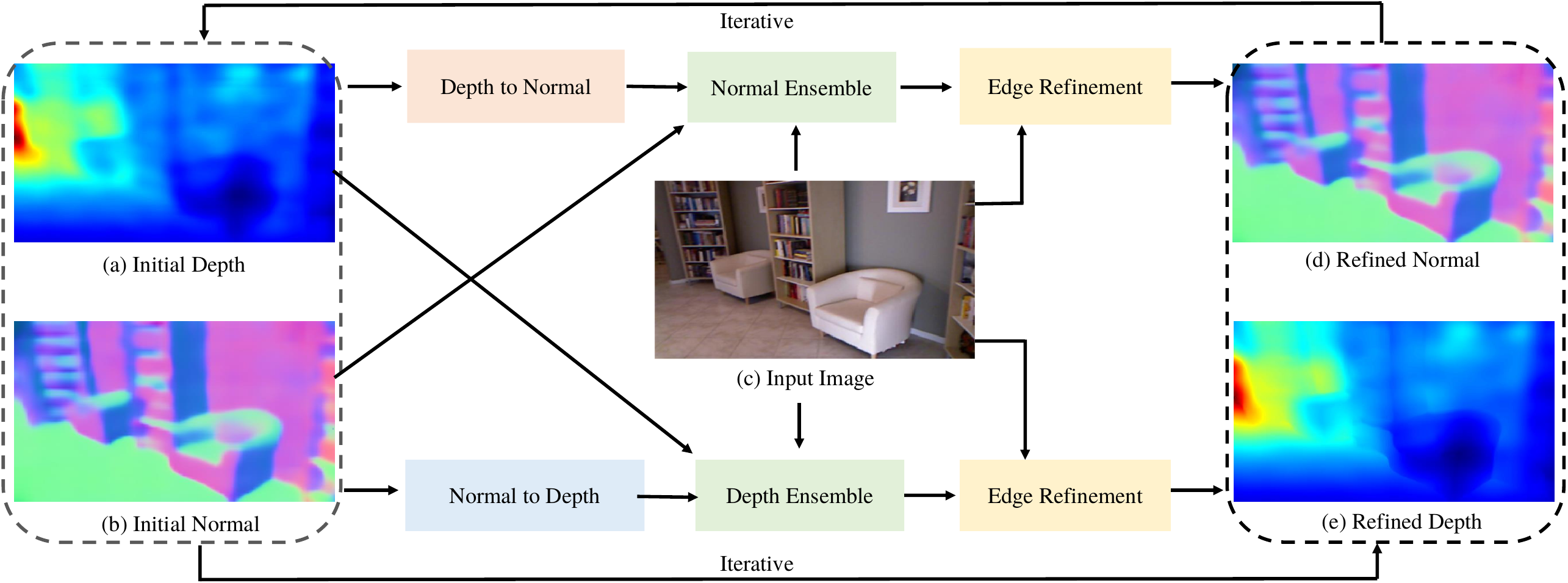}
\caption{The overall structure of \ourmodelshort. {\ourmodelshort} takes as inputs initial depth estimation (a), initial normal estimation (b), and an input image (c). The initial depth and surface normal are firstly refined with depth-to-normal and normal-to-depth modules. Then, the depth (normal) ensemble module is adopted to combine results from the initial estimation. It is followed by the edge refinement module, which reduces noise and refines boundaries. {\ourmodelshort} can be applied for multiple times by iteratively taking the refined normal and depth as inputs.}
\label{fig:overview}
\end{figure*}

\begin{figure*}[t]
\centering
\begin{subfigure}[\footnotesize{Depth-to-normal (L) and normal ensemble (R) modules. L: left shaded box, R: right shaded box.}]
   {\vspace{-0.1in}\includegraphics[width=0.95\textwidth]{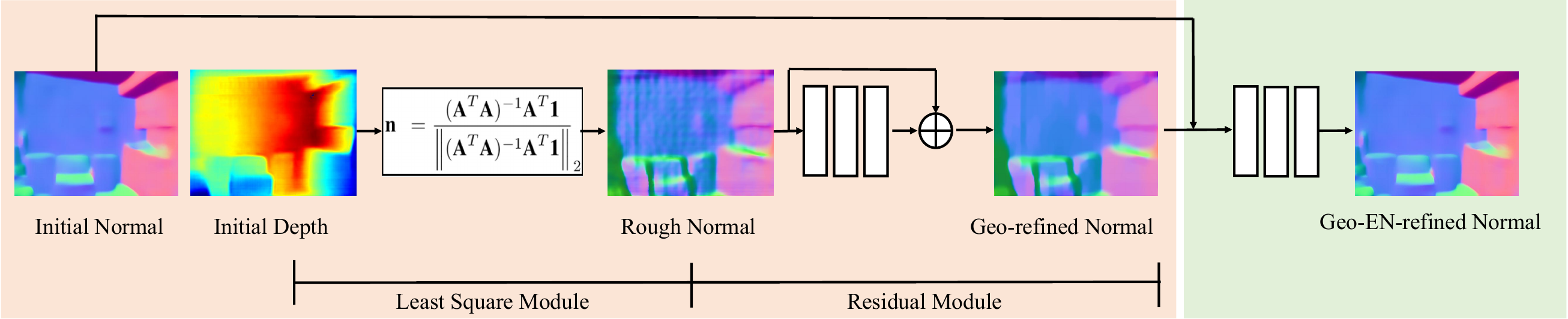}}\\
\end{subfigure}
\begin{subfigure}[\footnotesize{Normal-to-depth (L) and depth ensemble (R) modules. L: left shaded box, R: right shaded box.}]
   {\vspace{-0.1in}\includegraphics[width=0.95\textwidth]{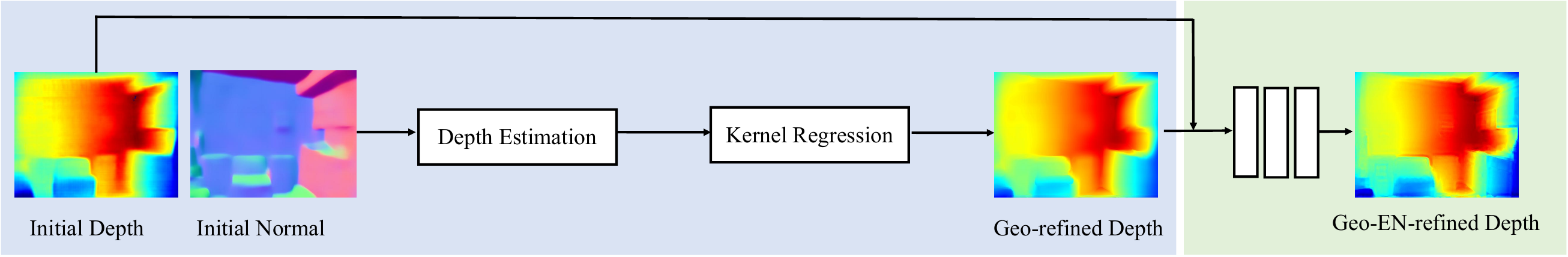}}\\
\end{subfigure}
\begin{subfigure}[\footnotesize{Edge-aware refinement module for depth. Residual (weight) maps include ``left to right'', ``right to left'',  ``top to bottom'',  ``bottom to top''. }]
{\vspace{-0.1in}\includegraphics[width=0.95\textwidth]{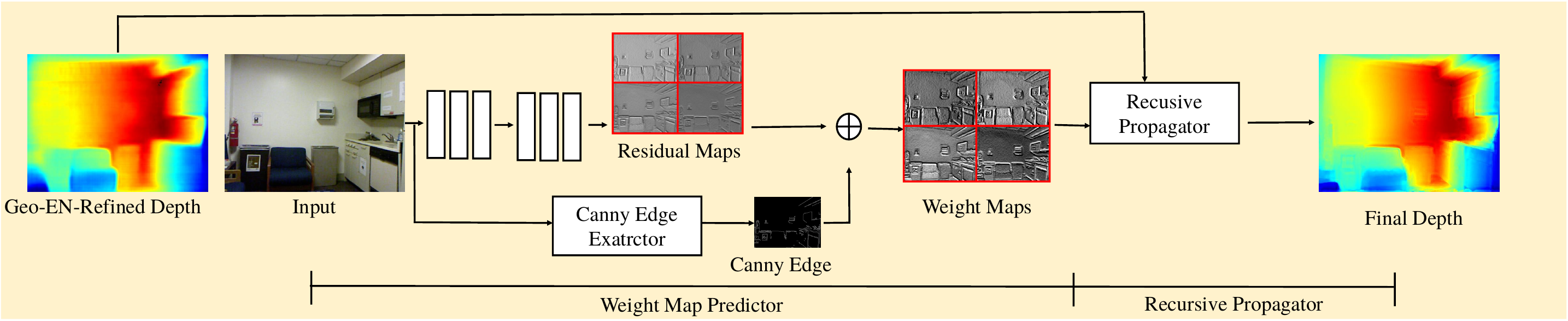}}
\end{subfigure}
\begin{subfigure}[\footnotesize{\textcolor{black}{
Edge-aware refinement module for surface normal. Residual (weight) maps include ``left to right'', ``right to left'',  ``top to bottom'',  ``bottom to top''. 
}}]
{\vspace{-0.1in}\includegraphics[width=0.95\textwidth]{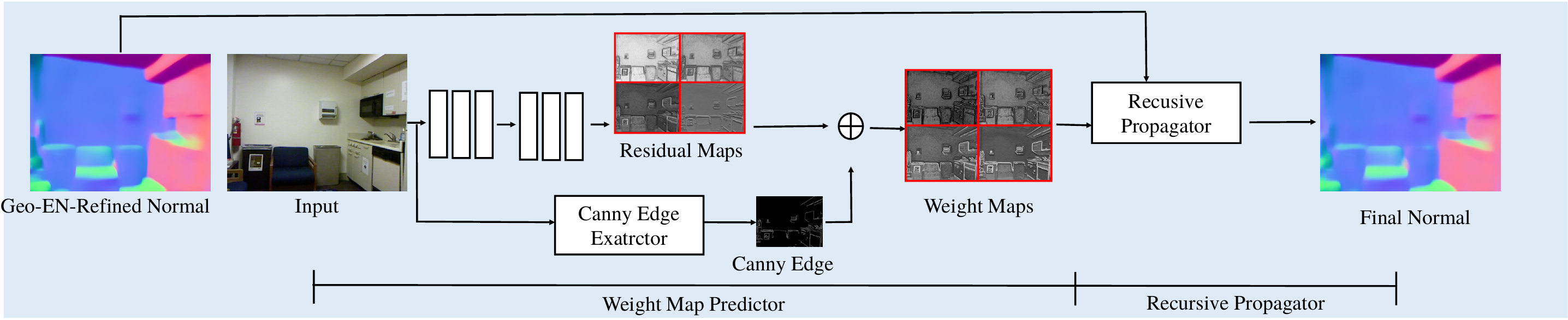}}
\end{subfigure}
\vspace{-0.1in}\caption{\textcolor{black}{{\ourmodelshort} components.
(a) 
The depth-to-normal module (L) first estimates ``Rough Normal'' from the ``Initial Depth'' with least square fitting; 
normals are then refined by the residual module producing ``Geo-refined Normal''; 
a normal ensemble network (R) is utilized to fuse the initial and Geo-refined normals generating ``Geo-EN-refined normal''.
(b) 
The normal-to-depth module (L) takes the ``Initial Depth'' and ``Initial Normal'' as inputs;
the normal map helps propagate the initial depth prediction to neighbors;
depth estimates are aggregated by the kernel regression module producing ``Geo-refined Depth''.
The depth ensemble module (R) taking ``Geo-refined Depth'' and ``Initial Depth'' as inputs further improves prediction generating ``Geo-EN-refined Depth''.
(c) 
The edge-aware refinement module first constructs direction-aware propagation ``Weight Maps'' by combining low-level edges with ``Residual Maps''; the recursive propagator utilizes the learned weight maps to refine ``Geo-EN-refined Depth'' producing ``Final Depth'';  (d) the edge-aware refinement module for surface normal. Please zoom in to see more details.}
}
\label{fig:normal-depth}
\end{figure*}

In this section, we first introduce the overall architecture of our {\ourmodelshort}, and then elaborate on its components.

\subsection{{Overall Architecture}}\label{subsect:full-model}

The overall architecture of {\ourmodelshort} is illustrated in Fig.~\ref{fig:overview}.
Based on the initial depth map predicted by the backbone CNNs (Sec.~\ref{sec:backbone}), we apply the depth-to-normal module (Sec.~\ref{subsect:depth-to-normal}) to transfer the initial depth map to the normal map as shown in Fig.~\ref{fig:normal-depth} (a) -- (L). This module refines the surface normals with the initial depth map considering  geometric constraints. 
Similarly, given the initial surface normal estimation, we generate the depth using the normal-to-depth module (Sec.~\ref{subsect:normal-to-depth}). 
This enhances the depth prediction by incorporating the inherent geometric constraints to the estimation of depth from normals.
The depth/normal maps generated with the above components are then  adjusted via the depth (normal) ensemble module (Sec.~\ref{subsect:ensemble}).
Furthermore, by removing noisy results and refining boundary predictions, the edge refinement network as described in Sec.~\ref{subsect:refine} further improves the predictions and generates the refined results as shown in Fig.~\ref{fig:overview} (d)-(e).
Finally, {\ourmodelshort} can be applied iteratively by taking the refined results from previous iteration as inputs as described in Sec.~\ref{sec:inferandtrain}.

\subsection{Depth-to-Normal Module}\label{subsect:depth-to-normal}
Learning geometrically consistent surface normals from depth via directly applying neural networks is surprisingly hard as discussed in Sec.~\ref{sect:naive-solution}.
\textcolor{black}{To this end, we propose a depth to normal transformation module that explicitly incorporates depth-normal consistency into deep neural networks.}
We start our discussion with the least square module, viewed as a fix-weight neural network.
We then  describe  the residual sub-module that aims at smoothing and combining the results with initial normals as in Fig.~\ref{fig:normal-depth} (a) -- (L).

\vspace{0.1in}
\noindent\textbf{Pinhole camera model. }
As a common practice, we adopt  the pinhole camera model.
We denote $\left(u_i, v_i\right)$ as the location of pixel $i$ in the 2D image.
Its corresponding location in 3D space is $\left(x_i, y_i, z_i\right)$, where $z_i$ is the depth.
Based on the geometry of perspective projection, we have
\begin{align}\label{eq:camera-model}
x_i = (u_i - c_x) * z_i / f_x, \nonumber \\
y_i = (v_i - c_y) * z_i / f_y.
\end{align}
where $f_x$ and $f_y$ are the focal length along the $x$ and $y$ directions respectively.
$c_x$ and $c_y$ are coordinates of the principal points.

\vspace{0.1in}
\noindent\textbf{Least square sub-module.} 
We formulate the inference of surface normals from a depth map as a least-square problem.
Specifically, for any pixel $i$, given its depth $z_i$, we first compute its 3D coordinates $\left(x_i, y_i, z_i\right)$ from its 2D coordinates $(u_i, v_i)$ relying on the pinhole camera model.
To compute the surface normal of pixel $i$, we need to determine the tangent plane, which crosses pixel $i$ in 3D space.
We follow the traditional assumption that pixels within a local neighborhood lie on the same tangent plane.
In particular, we define the set of neighboring pixels, including pixel $i$ itself, as
{\small
\begin{align}
\mathcal{N}_i = \left\{\left(x_j, y_j, z_j\right) \middle\vert \vert u_i - u_j \vert < \beta, \vert v_i - v_j \vert < \beta, \vert z_i - z_j \vert < \gamma z_i\right\}, \nonumber
\end{align}
}%
where $\beta$ and $\gamma$ are hyperparameters controlling the size of neighborhood along $x$, $y$, and depth axes respectively.
With these pixels on the tangent plane, the surface normal estimate $\mathbf{n} ={\begin{bmatrix} n_x, n_y, n_z \end{bmatrix}}$ should satisfy the over-determined linear system of equations 
\begin{align}\label{eq:linear-sys}
\mathbf{An} = \mathbf{b}, \qquad \text{subject to} \quad {\left\lVert \mathbf{n}\right\rVert}_2^2 = 1.
\end{align}
where
\begin{align}\mathbf{A} = \begin{bmatrix}
    x_{1} & y_{1} & z_{1} \\
    x_{2} & y_{2} & z_{2} \\
    \vdots & \vdots & \vdots \\
    x_{K} & y_{K} & z_{K}
\end{bmatrix} \in \mathbb{R}^{K\times3},
\end{align}
and $\mathbf{b}\in \mathbb{R}^{K\times1}$ is a constant vector. {$K$} is the size of $\mathcal{N}_i$, \ie, the set of neighboring points.
The least square solution of this problem, which minimizes $\Vert \mathbf{An}- \mathbf{b} \Vert^{2}$, can be computed in closed form as
\begin{align}\label{eq:least-square}
\mathbf{n} = \frac{{(\mathbf{A}^{\top}\mathbf{A})^{-1}\mathbf{A}^{\top}\mathbf{1}}}{\left\lVert
{(\mathbf{A}^{\top}\mathbf{A})^{-1}\mathbf{A}^{\top}\mathbf{1}}\right\rVert}_2,
\end{align}
where $\mathbf{1}\in \mathbb{R}^k$ is a vector with all-one elements.
It is not surprising that Eq. \eqref{eq:least-square} can be regarded as a fix-weight neural network, which predicts surface normals given the depth map.

\vspace{0.1in}
\noindent\textbf{Residual sub-module.} 
This least-square module occasionally produces noisy surface normal estimation (see Fig.~\ref{fig:normal-depth} (a): ``Rough Normal'') due to issues like noise and improper neighborhood size. 
{To further improve the quality, we propose a residual module, which consists of a $3$-layer CNN with skip-connections as shown in Fig.~\ref{fig:normal-depth} (a).} The goal is to smooth the  noisy estimation from the least square module.

\vspace{0.1in}
\noindent\textbf{Overall architecture.} 
The architecture of the depth-to-normal module is illustrated in Fig.~\ref{fig:normal-depth} (a) -- (L).
By explicitly leveraging the geometric relationship between depth and surface normals, our network circumvents the aforementioned difficulty in learning geometrically consistent depth and surface normals.
Note that the module can be incorporated and jointly fine-tuned with other networks that predict depth maps from raw images.

\subsection{Normal-to-Depth Module}\label{subsect:normal-to-depth}

Now we turn our attention to the normal-to-depth module.
For any pixel $i$, given its surface normal $(n_{ix}, n_{iy}, n_{iz})$ and an initial estimate of depth $z_i$, the goal is to refine its depth.

First, note that given the 3D point $\left(x_i, y_i, z_i\right)$ and its surface normal $\left(n_{ix}, n_{iy}, n_{iz}\right)$, we can uniquely determine the tangent plane $\mathcal{P}_i$, which satisfies the following equation
\begin{align}\label{eq:tangent-plane}
n_{ix}(x - x_i) + n_{iy}(y - y_i) + n_{iz}(z - z_i) = 0.
\end{align}
As explained in Sec. \ref{subsect:depth-to-normal}, we  assume that pixels within a small neighborhood  of $i$ lie on this tangent plane $\mathcal{P}_i$. This neighborhood $\mathcal{M}_i$ is defined as
{\small
\begin{align}
\mathcal{M}_i = \left\{(x_j, y_j, z_j) \middle\vert \mathbf{n}_{j}^{\top}\mathbf{n}_i > \alpha, \vert u_i - u_j \vert < \beta, \vert v_i - v_j \vert < \beta \right\}, \nonumber
\end{align}
}%
where $\beta$ is a hyperparameter controlling the size of the neighborhood along the $x$ and $y$ axes, 
$\alpha$ is a threshold to rule out spatially close points, which are not approximately coplanar, and 
$(u_i, v_i)$ are the coordinates of pixel $i$ in the 2D image.

{For any pixel $j \in \mathcal{M}_i$, if the depth $z_j$ is given, }we can compute the depth estimate of pixel $i$ as $z_{ji}^{\prime}$ relying on Eqs.~\eqref{eq:camera-model} and \eqref{eq:tangent-plane}  as
\begin{align}\label{eq:depth_coplanar}
z_{ji}^{\prime} = \frac{n_{jx}x_j + n_{jy}y_j + n_{jz}z_j}{{(u_i -c_x)n_{jx}}/{f_x} +{(v_i - c_y)n_{jy}}/{f_y}+ n_{jz}}.
\end{align}
To refine the depth of pixel $i$, we then use kernel regression to aggregate the estimation from all pixels in the neighborhood as 
\begin{align}\label{eq:kernel-regression}
\hat{z}_i = \frac{\sum_{j\in{\mathcal{M}_i}} \mathbf{K}(\mathbf{n}_j, \mathbf{n}_i) z_{ji}^{\prime}} {\sum_{j\in{\mathcal{M}_i}} \mathbf{K}(\mathbf{n}_j, \mathbf{n}_i)},
\end{align}
where $\hat{z}_i$ is the refined depth, $\mathbf{n}_i ={\begin{bmatrix} n_{ix}, n_{iy}, n_{iz} \end{bmatrix}}$ and $\mathbf{K}$ is the kernel function. We use linear kernels ({\ie}, cosine similarity) {to measure the similarity between $n_i$ and $n_j$}, {\ie}, $\mathbf{K}(\mathbf{n}_j, \mathbf{n}_i) = \mathbf{n}_{j}^{\top}\mathbf{n}_i$.
In this case, the smaller the angle between normals $\mathbf{n}_i$ and $\mathbf{n}_j$ is, which means the higher probability that pixels $i$ and $j$ are in the same tangent plane, the more contribution the estimate $z_{ji}^{\prime}$ makes to the estimate of $\hat{z}_i$.

The above process is illustrated in Fig.~\ref{fig:normal-depth} (b) -- (L).
It can be viewed as a voting process where every pixel $j \in \mathcal{M}_i$ gives a ``vote'' to determine the depth of pixel $i$.
By utilizing the geometric relationship between surface normal and depth, we efficiently improve the quality of depth estimate without any weights to learn. 

\subsection{Depth (Normal) Ensemble Module}\label{subsect:ensemble}

To further enhance the prediction quality,  the ``Initial Depth (Normal)'' from the backbone network and the ``Geo-refined Depth'' from the geometric refinement are combined together with the ensemble module illustrated in Fig.~\ref{fig:normal-depth} (a) -- (R) for surface normal and Fig.~\ref{fig:normal-depth} (b) -- (R) for depth. In the following, we detail the depth ensemble module. The normal ensemble module shares a similar architecture.

The depth ensemble module takes as inputs ``Initial Depth'' from the backbone network and ``Geo-refined Depth'' (Fig.~\ref{fig:normal-depth} (b)) from the geometric module, and produces a refined depth -- ``Geo-EN-refined Depth'', as shown in Fig.~\ref{fig:normal-depth} (b).
To enlarge the receptive field of the ensemble module, the input is firstly processed with $3$ convolution layers with a dilation rate of $2$, kernel size $3\times3$, and channel number $128$. This is followed by another $2$ dilation-free convolution layers with kernel size $3\times3$ and channel number $128$.

\subsection{Edge-aware Refinement Module}\label{subsect:refine}
We design an edge-aware refinement module (see Fig.~\ref{fig:normal-depth} (c) \textcolor{black}{\& Fig.~\ref{fig:normal-depth} (d)}) to further enhance the prediction inspired by~\cite{chen2016semantic}.
This module enhances the boundary prediction and removes noisy predictions by gradually aggregating the information from neighboring pixels. This process is guided by a set of learned weight maps (see Fig.~\ref{fig:normal-depth}: ``Weight Maps'').
The edge-aware refinement contains two sub-modules, \ie, the weight map predictor and the recursive propagator.
This module uses the same architecture for both the depth and normal prediction so that we only elaborate on the details for the depth as below.

\vspace{0.1in}
\noindent\textbf{Weight map predictor. }
The weight map predictor contains a canny edge operator extracting low-level edge information and a residual network to learn edge-aware propagation weights.
The output is fused by element-wise summation as shown in Fig.~\ref{fig:normal-depth} (c).
The residual network contains $3$ convolutional layers with ReLU nonlinearity, dilation rate $2$, kernel size $3\times3$, and channel number $32$, followed by $3$ convolution layers without dilation using the same parameter setting as above.
Finally, a convolution layer with kernel size $1\times1$ and channel number $4$ is adopted to produce the edge weight maps ${\mathcal{W}\in \mathbb{R}^{H\times W\times 4}}$, where ${H}$ and ${W}$ are image spatial size, and the $4$ channels represent propagation weights in four directions, {\ie}, left to right ($L{\rightarrow}R$), right to left ($R{\rightarrow}L$), top to bottom ($T{\rightarrow}B$), and bottom to top ($B{\rightarrow}T$).
The learned edge weight maps for depth refinement are illustrated in Fig.~\ref{fig:normal-depth} (c). The corresponding maps for surface normal are shown in Fig.~\ref{fig:normal-depth} (d), where a larger value means a higher chance to be near boundaries.

\vspace{0.1in}
\noindent\textbf{Recursive propagator. }
The recursive propagator takes as inputs the edge weight maps ${\mathcal{W}}$ and the input signal ${X}$ (depth or normal maps in our case), and recursively refines the input signal for ${T}$ times by applying the following operations
{\small
\begin{align}
L{\rightarrow}R: {\quad}{S}_{(i,j)}^{1,t} & = (1-\mathcal{W}_{(i,j,1)}){X}_{i-1,j}^t + {\mathcal{W}}_{(i,j,1)}{X}_{i,j}^t, \nonumber \\
R{\rightarrow}L: {\quad}{S}_{(i,j)}^{2,t} & = (1-\mathcal{W}_{(i,j,2)}){S}_{i+1,j}^{1,t} + \mathcal{W}_{(i,j,2)}{S}_{i,j}^{1,t}, \nonumber \\
T{\rightarrow}B: {\quad}{S}_{(i,j)}^{3,t} & = (1-\mathcal{W}_{(i,j,3)}){S}_{i,j-1}^{2,t} + \mathcal{W}_{(i,j,3)}S_{i,j}^{2,t}, \nonumber \\
B{\rightarrow}T:{\quad} {S}_{(i,j)}^{4,t} & = (1-\mathcal{W}_{(i,j,4)}){S}_{i,j+1}^{3,t} + \mathcal{W}_{(i,j,4)}{S}_{i,j}^{3,t}, \nonumber \\
{\quad}{X}_{(i,j)}^{(t+1)} & = S_{(i,j)}^{4,t},
\end{align}
}%
where ${X^{0}=X}$. ${S^{1,t}}$, ${S^{2,t}}$, ${S^{3,t}}$, and ${S^{4,t}}$ represent intermediate results after ``$L{\rightarrow}R$'', ``$R{\rightarrow}L$'', ``$T{\rightarrow}B$'', and ``$B{\rightarrow}T$'' propagation at step $t$ respectively.

At each step, the recursive propagator takes as input the previous iteration result ${X^t}$ and produces a new estimation ${X^{t+1}}$ employing a weighted summation of depth diffused from neighboring pixels and the current depth values. The weights are determined by the learned edge-aware weight maps ${\mathcal{W}}$.
In regions near boundaries, the learned weights ${\mathcal{W}_{i,j}}$ are large to avoid blurring and preserve sharp results. 
On the other hand, the learned weights are small in non-boundary regions  removing noisy predictions.
The edge-aware weights enable us to separate computational intensive two-dimensional propagation into four one-dimensional propagations, which is more efficient without sacrificing the quality~\cite{chen2016semantic,gastal2011domain}.
The above propagation procedure is executed for ${T}$ times to incorporate long-range dependencies. We use ${T=3}$ in our experiments.

\subsection{Iterative Inference and Training Details}\label{sec:inferandtrain}
\vspace{0.1in}
\noindent
\textbf{Iterative inference. }
{\ourmodelshort} can be applied iteratively to further improve the results as shown in Fig.~\ref{fig:overview}. 
The refined depth and normal maps from previous iterations can further serve as the inputs to {\ourmodelshort} for iterative refinement.
Note that we only apply this iteratively during the inference. In the training phase, {\ourmodelshort} is only applied once to reduce the memory consumption and improve the training efficiency.

\begin{figure}[t]
\centering
\includegraphics[width=0.5\textwidth]{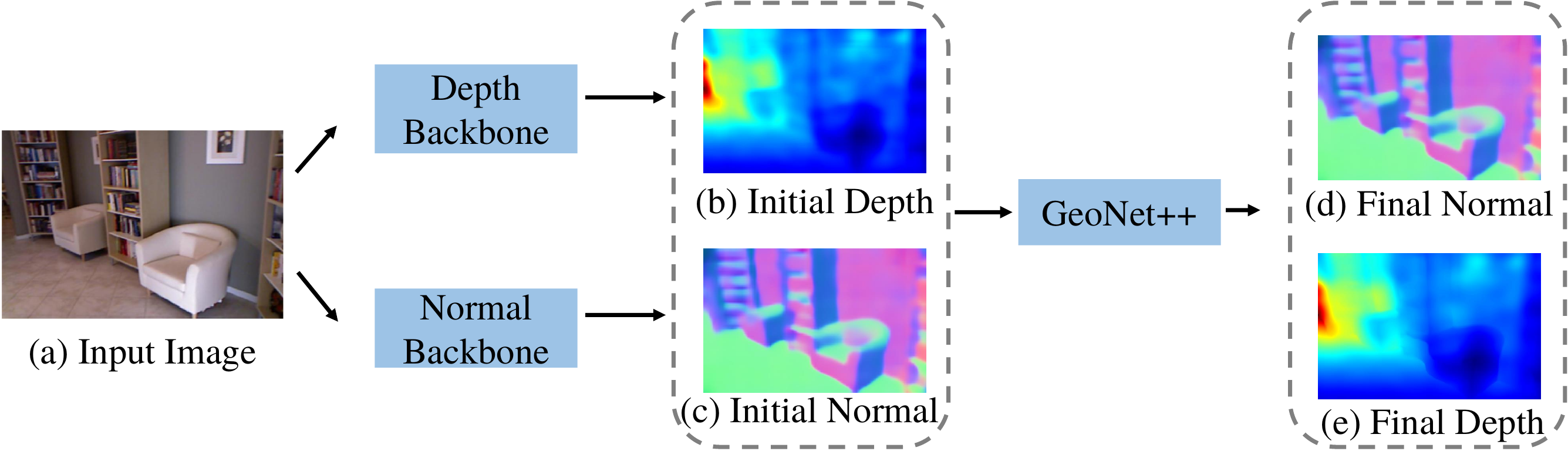}
\caption{Our end-to-end trainable full system with backbone architecture and {\ourmodelshort}.}
\label{fig:backbone}
\end{figure}

\vspace{0.1in}
\noindent
\textbf{End-to-end network training. }
Our full system  is shown in Fig.~\ref{fig:backbone}. The backbone network produces the initial depth and surface normal maps, which are further refined with {\ourmodelshort} by incorporating the geometric constraint and the edge information. The whole system can be trained end-to-end. 
In the following, we explain our loss function for training the full system.
We denote the initial, refined, and ground-truth depth of pixel $i$ as $z_i$, $\hat{z}_i$ and $z^{\text{gt}}_i$ respectively.
Similarly, we denote $\mathbf{n}_i$, $\hat{\mathbf{n}}_i$, and $\mathbf{n}^{\text{gt}}_i$ initial, refined, and ground-truth surface normals respectively. 
The overall loss function is the summation of two losses, one for the depth and one for the normals, $L = l_{\text{depth}} + l_{\text{normal}}$. The depth loss $l_{\text{depth}}$ is expressed as
\begin{align}
l_{\text{depth}} = \frac{1}{M}\left(\sum_i{\left\lVert {z_i} - {z^{\text{gt}}_i}\right\rVert}_2^2 + \eta \sum_i{\left\lVert {\hat{z}_i} - {z^{\text{gt}}_i} \right\rVert}_2^2 \right), 
\end{align}
with ${M}$ the total number of pixels.
The surface normal loss is
\begin{align}
l_{\text{normal}} = \frac{1}{M}\left( \sum_i{\left\lVert \mathbf{n}_i -  \mathbf{n}^{\text{gt}}_i \right\rVert}_2^2 + \lambda \sum_i {\left\lVert \hat{\mathbf{n}}_i -  \mathbf{n}^{\text{gt}}_i \right\rVert}_2^2 \right). 
\end{align}
Here $\lambda$ and $\eta$ are hyperparameters which balance the contribution of individual terms.
We will explain their values in the following section.

\section{Experimental Setup}\label{sect:experiments}

\subsection{Backbone Networks}\label{sec:backbone}
We validate the effectiveness of our proposed {\ourmodelshort} on top of several baseline architectures.

\vspace{0.1in}
\noindent{\textbf{Baseline network.}} In most experiments, we utilize a modified VGG-16~\cite{simonyan2014very}, {\ie}, deeplab-LargeFOV~\cite{chen2014semantic} with dilated convolution~\cite{chen2014semantic} and global pooling~\cite{liu2015parsenet,zhao2016pyramid} for initial depth and surface normal prediction. This is our baseline backbone network for comparison with VGG-based methods. It is also adopted for ablation studies. We utilize this baseline network to produce initial depth and/or surface normal in the experiments if not specified.

\vspace{0.1in}
\noindent\textbf{State-of-the-art approaches. } 
To further evaluate the effectiveness of the system, we also adopt state-of-the-art methods to produce the initial prediction of depth and surface normal. We experiment with Multi-scale CNN V1~\cite{eigen2014depth}, Multi-scale CNN V2~\cite{eigen2015predicting}, FCRN~\cite{laina2016deeper}, Multi-scale CRF~\cite{xu2017multi}, and DORN~\cite{fu2018deep} for initial depth estimation. For initial normal estimation, we employ the initial normal map from SkipNet~\cite{bansal2016marr}.
\begin{table}
\setlength{\tabcolsep}{2pt}
\caption{Performance of surface normal prediction on NYUD-V2 test set. ``Baseline'' refers to using VGG-16 network with global pooling to directly predict surface normal from raw images. ``SkipNet~\cite{bansal2016marr} + {\ourmodelshort}'' means building {\ourmodelshort} on top of the normal result of~\cite{bansal2016marr}. 
\textcolor{black}{``Baseline + Loss'' indicates that we only use a geometry-aware loss function as~\cite{yang2018unsupervised}.}
}
\label{tab:com-normal}
\centering
\scalebox{0.95}{
\begin{tabular}{c|ccc|ccc}
\toprule
\multirow{1}{*}{} &
\multicolumn{3}{c|}{{Error}}  &
\multicolumn{3}{c}{{Accuracy}} \\
& {{mean}} & {{median}} & {{rmse}} & {{$11.25^\circ$}} & {{$22.5^\circ$}} & {{$30^\circ$}} \\
\midrule
\midrule
3DP~\cite{fouhey2013data} & 35.3 & 31.2 & - & 16.4 & 36.6 & 48.2 \\
3DP (MW)~\cite{fouhey2013data} & 36.3 & 19.2 & - & 39.2 & 52.9 & 57.8 \\
UNFOLD~\cite{fouhey2014unfolding} & 35.2 & 17.9 & - & 40.5 & 54.1 & 58.9 \\
Discr.~\cite{zeisl2014discriminatively} & 33.5 & 23.1 & - & 27.7 & 49.0 & 58.7 \\
Multi-scale CNN V2~\cite{eigen2015predicting}& {23.7} & {15.5} & {-} & {39.2} & {62.0} & {71.1} \\
Deep3D~\cite{wang2015designing} & 26.9 & 14.8 & - & 42.0 & 61.2 & 68.2 \\
SURGE~\cite{wang2016surge}&{20.6}&{12.2}&{-}&{47.3}&{68.9}&{76.6}\\
SkipNet~\cite{bansal2016marr} & {19.8} & {12.0} & {28.2} & {47.9} & {70.0} & {77.8} \\ \midrule
SkipNet~\cite{bansal2016marr} + {\ourmodelshort}&19.6&11.6&28.3&48.9&71.2&78.7\\
Our Baseline & {19.4}&{12.5}&{27.0}&{46.0}&{70.3}&{78.9}\\
Our Baseline + {\ourmodelshortoriginal}~\cite{qi2018geonet} & {{19.0}} & {{11.8}} & {{26.9}} & {{48.4}} & {{71.5}} & {{79.5}}\\
\textcolor{black}{Our Baseline + Loss} & \textcolor{black}{{19.0}} & \textcolor{black}{{11.8}} & \textcolor{black}{{26.9}} & \textcolor{black}{{48.3}} & \textcolor{black}{{71.4}} & \textcolor{black}{{79.5}} \\
Our Baseline + {\ourmodelshort} & {\textbf{18.5}} & {\textbf{11.2}} & {\textbf{26.7}} & {\textbf{50.2}} & {\textbf{73.2}} & {\textbf{80.7}}\\
\bottomrule
\end{tabular}}
\end{table}
\subsection{Datasets}

We evaluate the effectiveness of our method on the NYUD-V2 \cite{silberman2012indoor} and KITTI~\cite{uhrig2017sparsity} datasets.

\vspace{0.1in}
\noindent\textbf{NYUD-V2 dataset.} This dataset  contains $464$ video sequences of indoor scenes, which are further divided into $249$ sequences for training and $215$ for testing. 
We sample $30,816$ frames from the training video sequences as the training data.
For the training set, we use the in-painting method of~\cite{levin2004colorization} to fill in invalid or missing pixels in the ground-truth depth map. We then generate a ground-truth surface normal map following the procedure of \cite{wang2015designing}.

\vspace{0.1in}
\noindent\textbf{KITTI dataset.} This dataset captures various scenes for autonomous driving.
We follow the setting of \cite{eigen2015predicting} and use $22,600$ images from $32$ scenes for training, and 697 images from the other 29 scenes for testing. For the KITTI dataset, we utilize Multi-scale CNN V1~\cite{eigen2014depth} and DORN~\cite{fu2018deep} with author-released models to produce initial depth. We generate the ground-truth normals using the same procedure as in the NYUD-V2 dataset with the provided LiDAR depth.

\subsection{Implementation Details}
Our {\ourmodelshort} is implemented in TensorFlow v1.2~\cite{abadi2016tensorflow}.
For our VGG baseline network, we initialize the two-stream CNNs with networks pre-trained on ImageNet. Other baseline approaches are initialized with their corresponding pre-trained models, which are fixed in the procedure of fine-tuning {\ourmodelshort}. We use Adam~\cite{kingma2014adam} to optimize the network, and the norm of gradients are clipped, so that they are no larger than $5$.
The initial learning rate is $1e^{-4}$. It is adjusted following the polynomial decay strategy with power parameter $0.9$.
Random horizontal flip is utilized for augmentation.
While flipping images, we multiply the x-direction of surface normal maps with $-1$.

The whole system is trained with batch-size $1$ for $40$K iterations on the NYUD-V2 dataset and $80$-k iterations on the KITTI dataset. Hyperparameters $\{\alpha, \beta, \gamma, \lambda, \eta\}$ are set to $\{0.95, 9, 0.05, 0.01, 0.5\}$ according to validation on $5\%$ randomly split training data. $\lambda$ is set to a small value due to numerical instability when computing the matrix inverse in the least square module -- the gradient of Eq.~\eqref{eq:least-square} needs to compute the inverse of matrix $A^TA$, which might be inaccurate if the condition number is large. Setting  $\lambda = 0.01$ mitigates this effect.

\subsection{2D Pixel-wise Metrics}

Following~\cite{eigen2014depth,laina2016deeper,xu2017multi}, we adopt four metrics to evaluate the resulting depth map quantitatively. They are root mean square error (rmse), mean $\log{10}$ error ($\log10$), mean relative error (rel), and pixel accuracy as percentage of pixels with $\max({z_i}/{z_i^{gt}}, z_i^{gt}/{z_i}) < \delta$ for $\delta\in[1.25, 1.25^2,1.25^3]$.
The evaluation metrics for surface normal prediction~\cite{wang2015designing,bansal2016marr,eigen2015predicting} are mean of angle error (mean), median of angle error (median), root mean square error (rmse), and pixel accuracy as percentage of pixels with angle error below threshold $t$ where $t\in[11.25^\circ, 22.5^\circ, 30^\circ]$.

\begin{figure*}
\centering
\begin{tabular}{@{\hspace{0.1mm}}c@{\hspace{0.1mm}}c@{\hspace{0.1 mm}}c@{\hspace{0.1 mm}}c@{\hspace{0.1 mm}}c@{\hspace{0.1 mm}}c@{\hspace{0.1 mm}}c}
\includegraphics[width=0.140\linewidth]{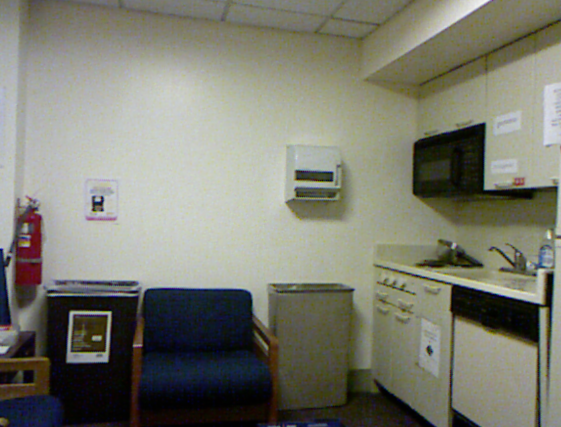} &
\includegraphics[width=0.140\linewidth]{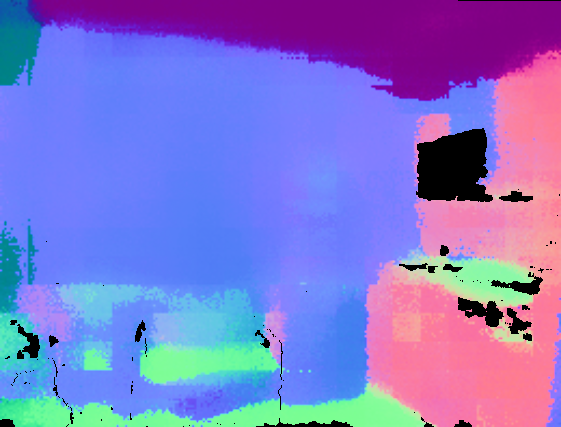}&
\includegraphics[width=0.140\linewidth]{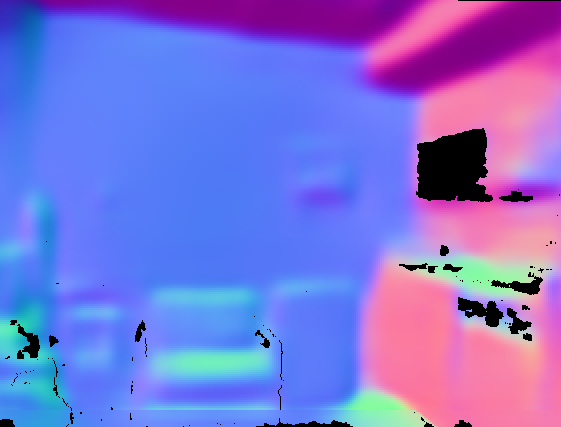}&
\includegraphics[width=0.140\linewidth]{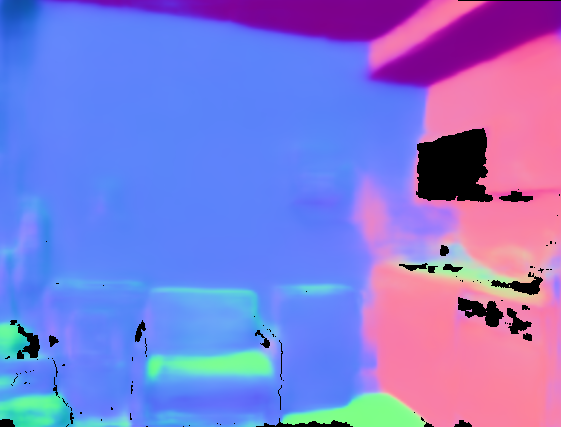}&
\includegraphics[width=0.140\linewidth]{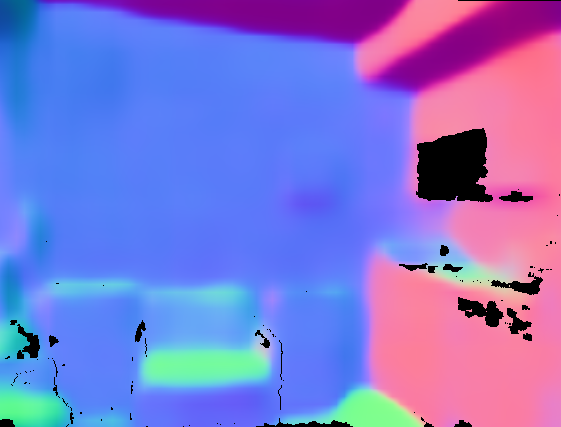}&
\includegraphics[width=0.140\linewidth]{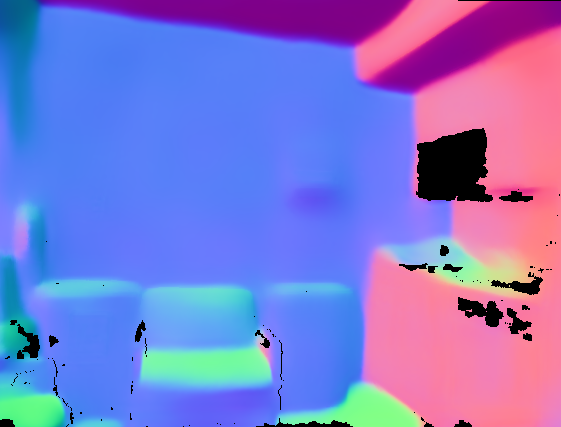}&
\includegraphics[width=0.140\linewidth]{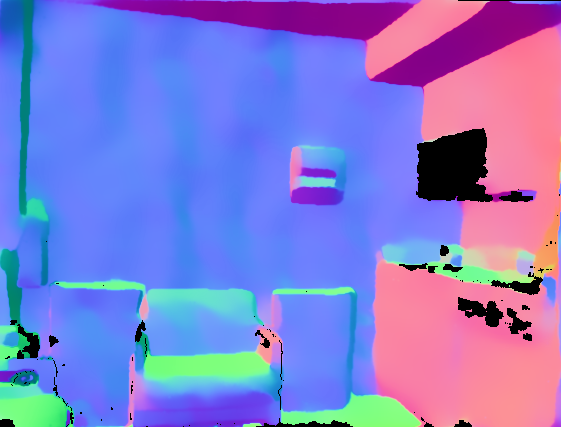}\\
\includegraphics[width=0.140\linewidth]{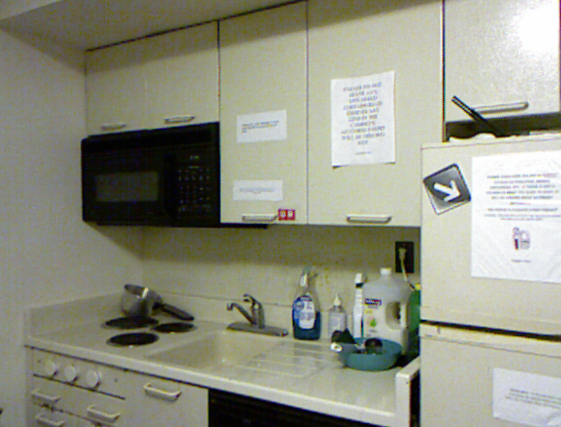} &
\includegraphics[width=0.140\linewidth]{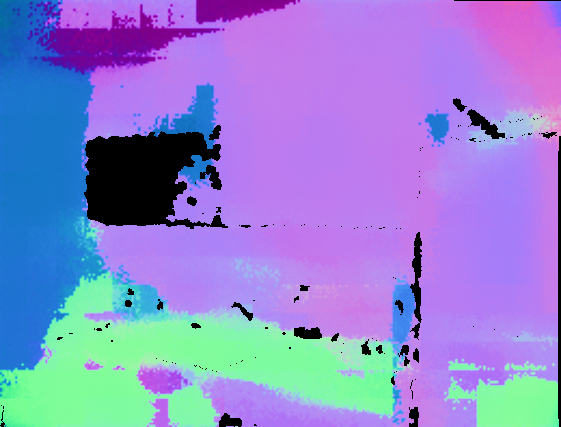}&
\includegraphics[width=0.140\linewidth]{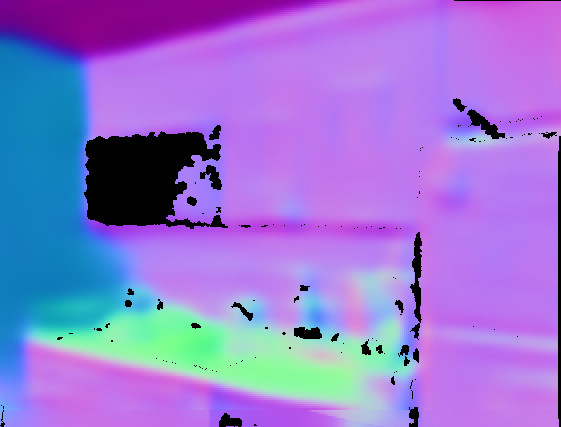}&
\includegraphics[width=0.140\linewidth]{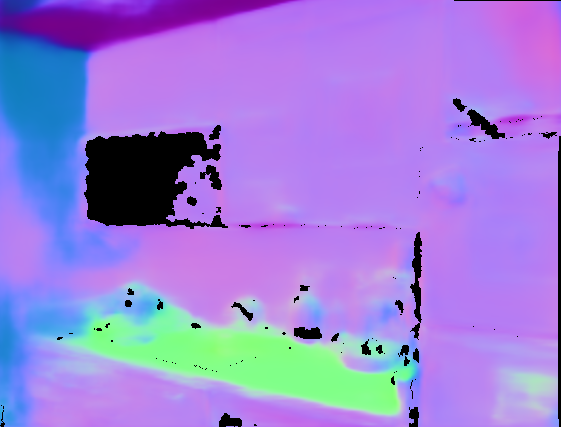}&
\includegraphics[width=0.140\linewidth]{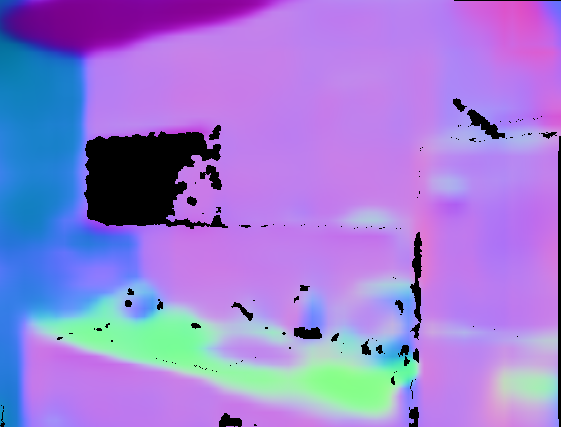}&
\includegraphics[width=0.140\linewidth]{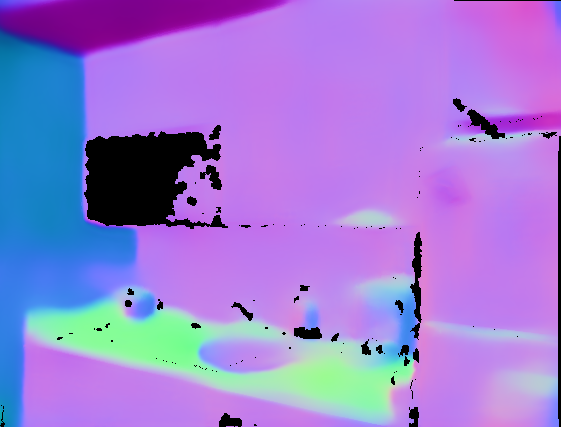}&
\includegraphics[width=0.140\linewidth]{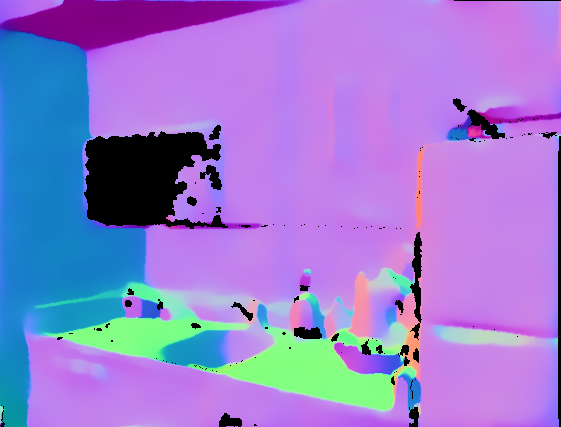}\\
\includegraphics[width=0.140\linewidth]{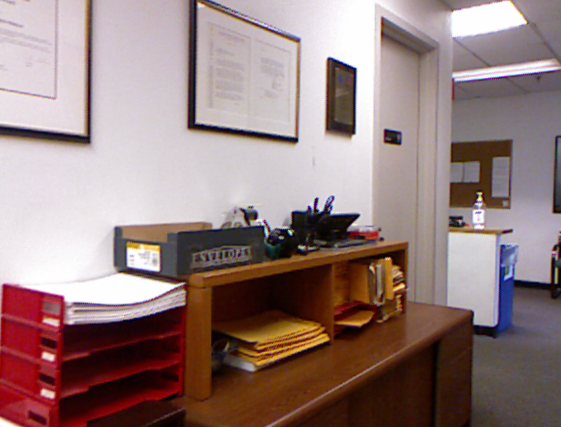} &
\includegraphics[width=0.140\linewidth]{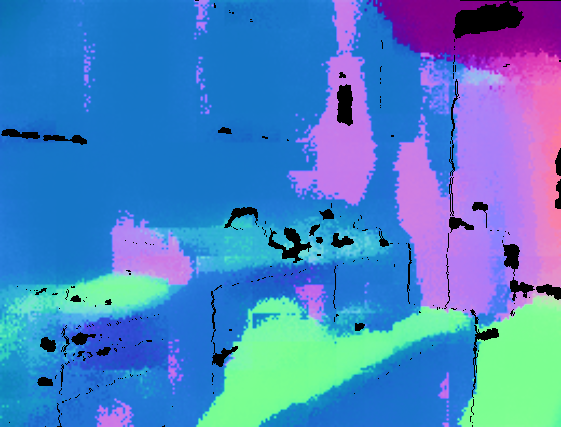}&
\includegraphics[width=0.140\linewidth]{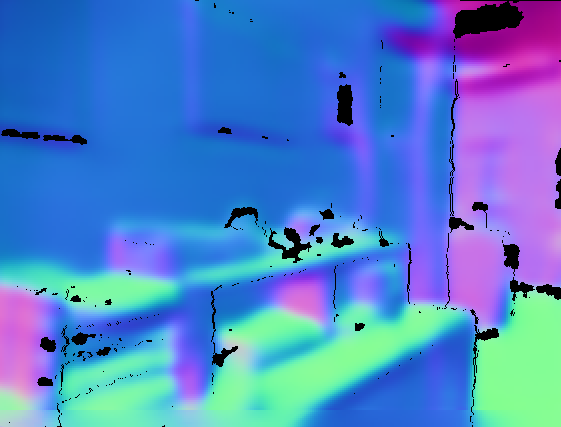}&
\includegraphics[width=0.140\linewidth]{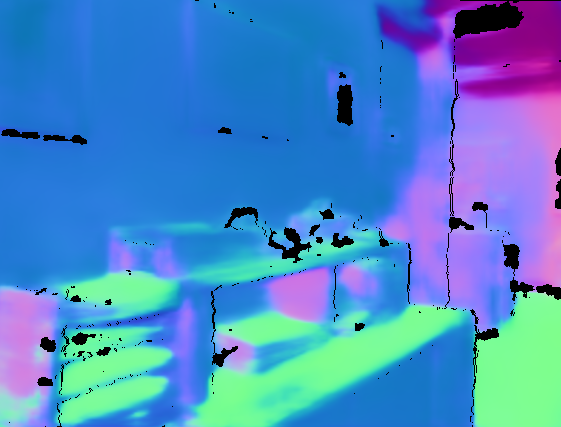}&
\includegraphics[width=0.140\linewidth]{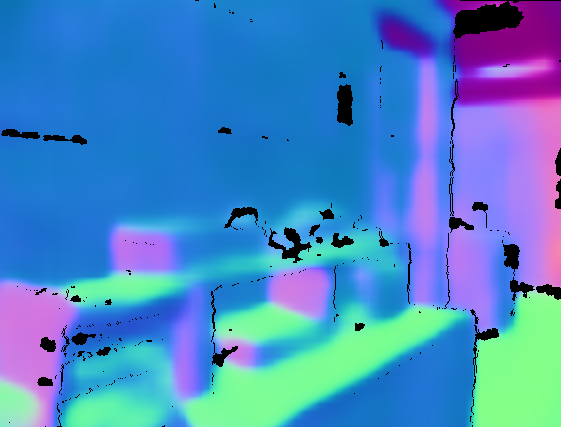}&
\includegraphics[width=0.140\linewidth]{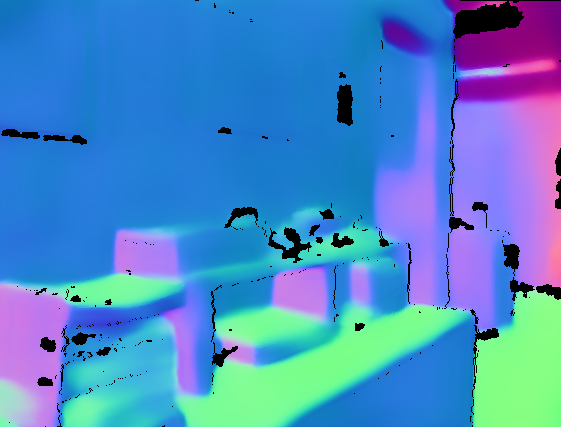}&
\includegraphics[width=0.140\linewidth]{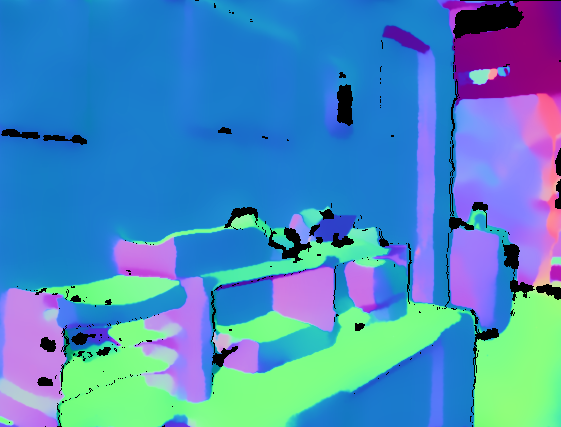}\\
\footnotesize{ (a) Images } &\footnotesize{(b) Deep3D~\cite{wang2015designing}}&\footnotesize{ (c) MS CNN V2~\cite{eigen2015predicting}}&\footnotesize{ (d) SkipNet~\cite{bansal2016marr}}&\footnotesize{ (e) { GeoNet~\cite{qi2018geonet}}}&\footnotesize{ (f) Ours ({\ourmodelshort})}&\footnotesize{(g) Ground truth} \\
\end{tabular}
\caption{Visual comparisons of surface normal predictions using VGG-16 as the backbone architecture.}
\label{fig:vis-normal}
\end{figure*}

\section{Comparison Regarding 2D Metrics}

In this section, we compare our {\ourmodelshort} with existing methods in terms of depth and/or surface normal prediction regarding 2D pixel-wise metrics on {NYUD-V2}~\cite{silberman2012indoor} and {KITTI}~\cite{geiger2013vision} datasets.

\begin{figure}
\centering
\begin{tabular}{@{\hspace{0.1mm}}c@{\hspace{0.1mm}}c@{\hspace{0.1mm}}c}
\includegraphics[width=0.32\linewidth]{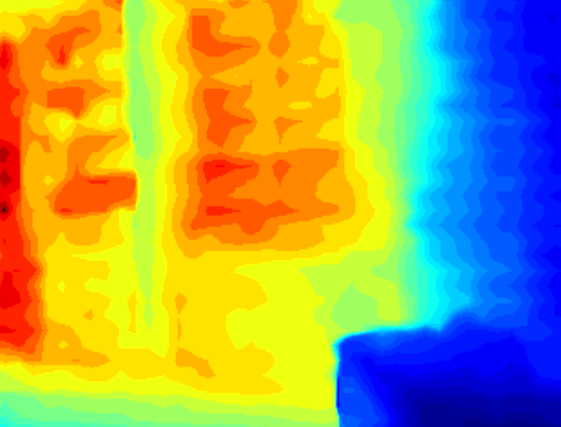}&
\includegraphics[width=0.32\linewidth]{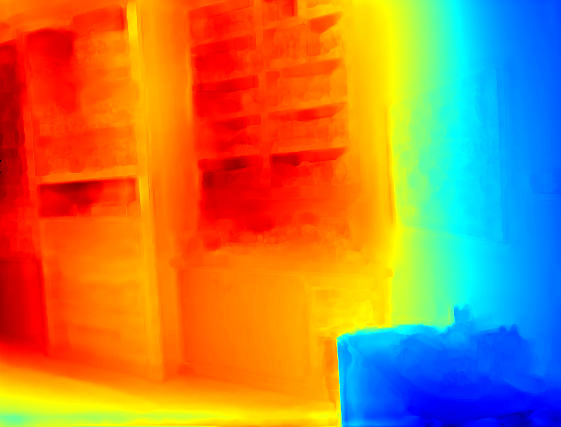}&
\includegraphics[width=0.32\linewidth]{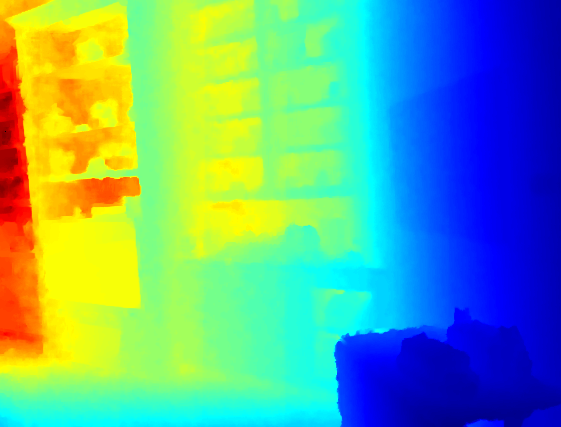}\\
\footnotesize{ RMSE: 0.844}&\footnotesize{ RMSE: 0.860}& \footnotesize{RMSE: 0.000}\\
\includegraphics[width=0.32\linewidth]{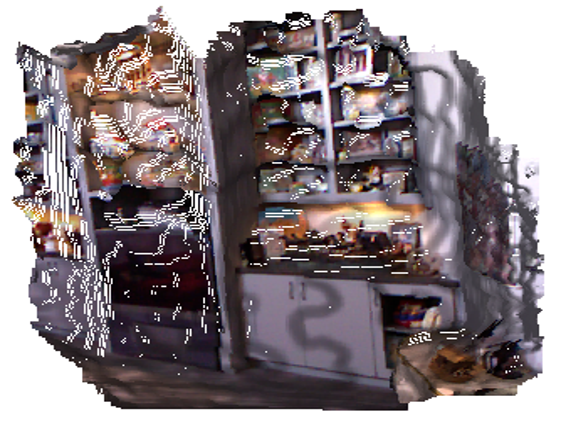}&
\includegraphics[width=0.32\linewidth]{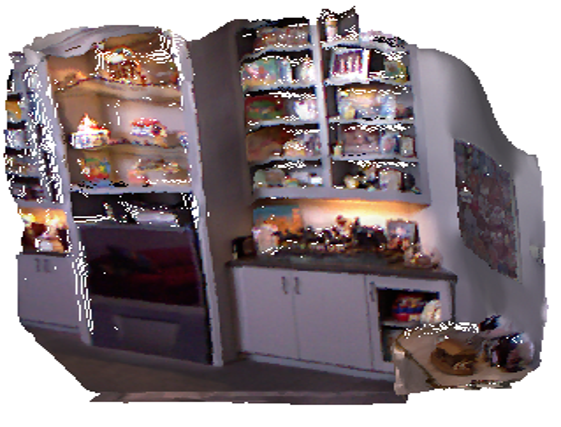}&
\includegraphics[width=0.32\linewidth]{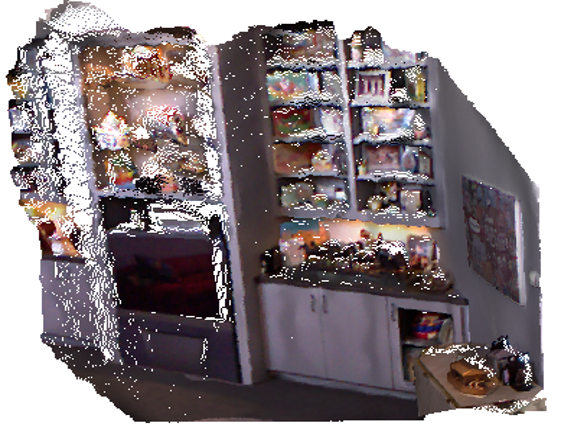}\\
\includegraphics[width=0.32\linewidth]{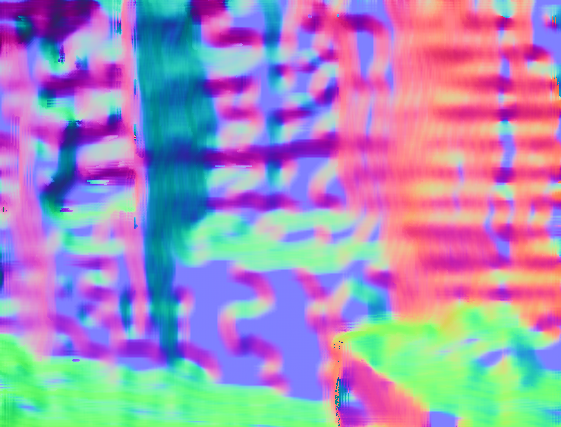}&
\includegraphics[width=0.32\linewidth]{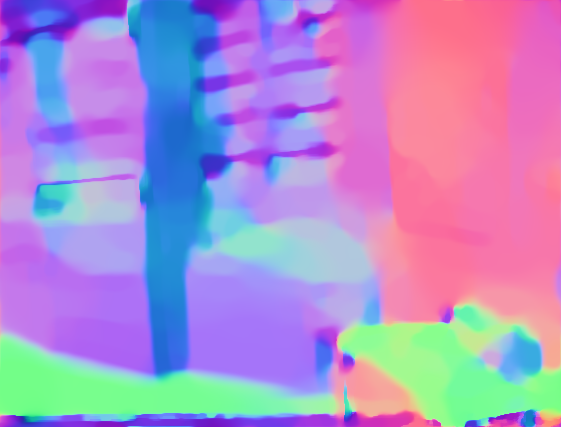}&
\includegraphics[width=0.32\linewidth]{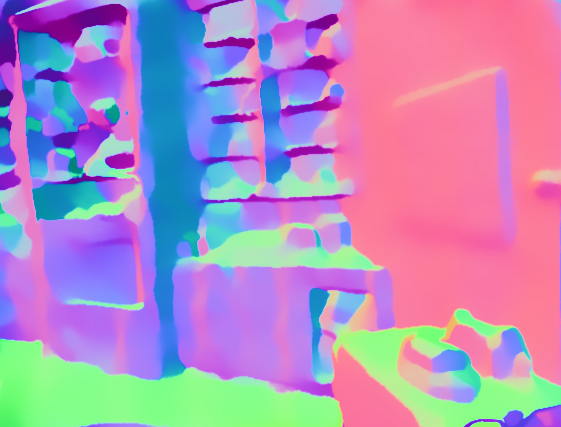}\\

\footnotesize{ (a)  DORN~\cite{fu2018deep} } &\footnotesize{(b) Ours} &\footnotesize{(c) Ground truth} \\
\end{tabular}
\caption{The first row shows the depth prediction results with the corresponding root mean square error listed below.
The second row shows the corresponding 3D point clouds.
The third row shows the estimated surface normal from the corresponding point clouds.}
\label{fig:vis-3d}
\end{figure}
\subsection{Experiments on NYUD-V2 Dataset}
\vspace{0.1in}
\textbf{Surface normal prediction. }
As shown in Tab.~\ref{tab:com-normal},
our {\ourmodelshortoriginal} consistently outperforms previous approaches regarding all metrics. 
{\ourmodelshort} further improves the results by incorporating the edge-aware refinement, ensemble modules, and the iterative inference strategy.
Since we use the same backbone network architecture VGG-16, the improvement stems from our depth-to-normal network. \textcolor{black}{Our explicit formulation is also more effective than implicitly incorporating the constraints via a loss function similar to~\cite{yang2018unsupervised}.}
Moreover, while taking the results from~\cite{bansal2016marr} and our own baseline depth network as the initial normal and depth prediction respectively, our {\ourmodelshort} improves the surface normal produced by~\cite{bansal2016marr}. Especially, the model is more effective in lower threshold regimes of the metric, which are more challenging.

\begin{figure*}
\centering
\begin{tabular}{@{\hspace{0.1mm}}c@{\hspace{0.1mm}}c@{\hspace{0.1 mm}}c@{\hspace{0.1 mm}}c@{\hspace{0.1 mm}}c@{\hspace{0.1 mm}}c@{\hspace{0.1 mm}}c}
\includegraphics[width=0.142\linewidth]{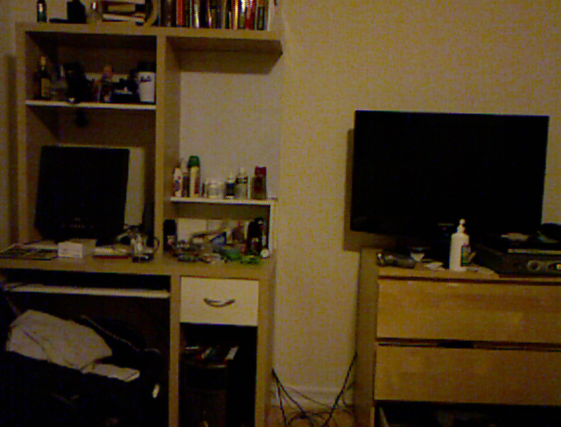} &
\includegraphics[width=0.142\linewidth]{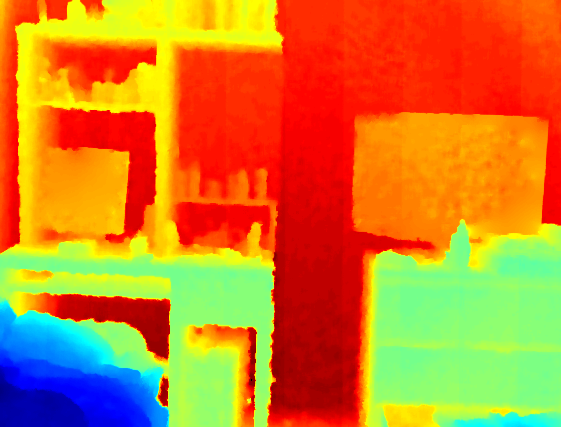}&
\includegraphics[width=0.142\linewidth]{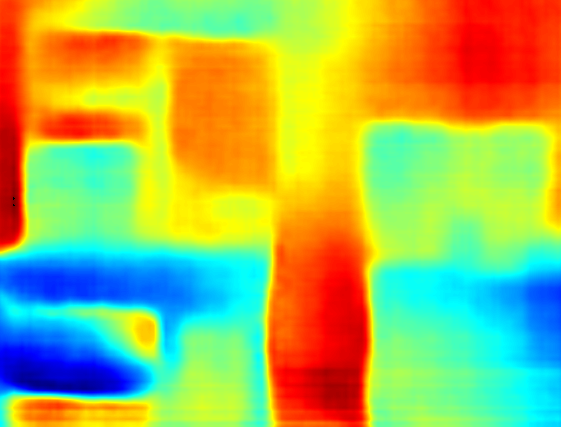}&
\includegraphics[width=0.142\linewidth]{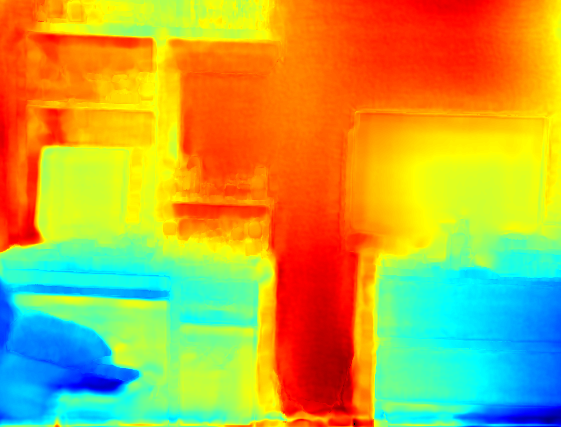}&
\includegraphics[width=0.142\linewidth]{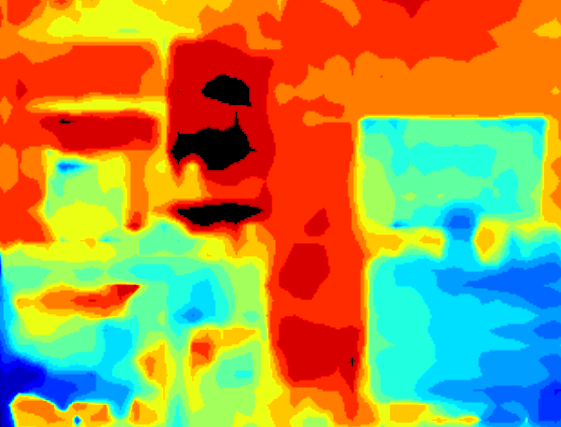}&
\includegraphics[width=0.142\linewidth]{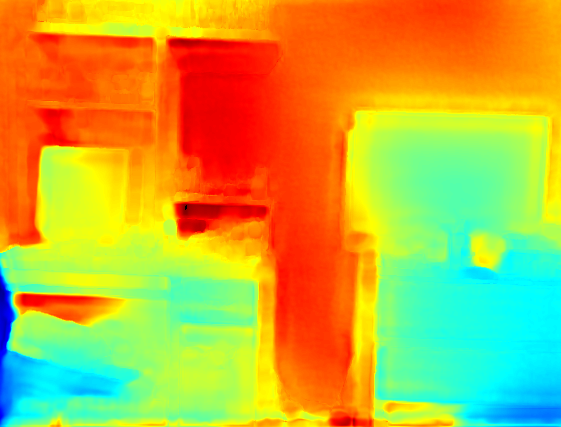}&
\includegraphics[width=0.142\linewidth]{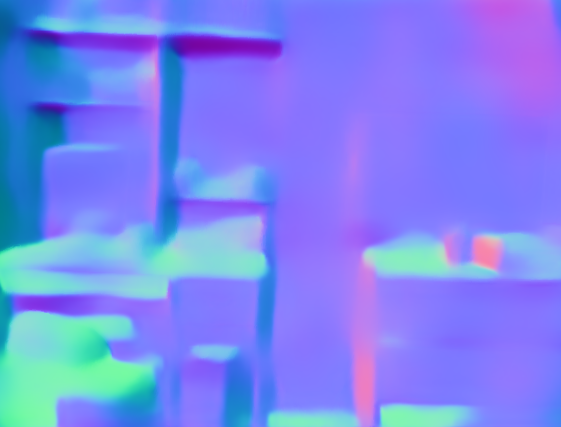}\\
\includegraphics[width=0.142\linewidth]{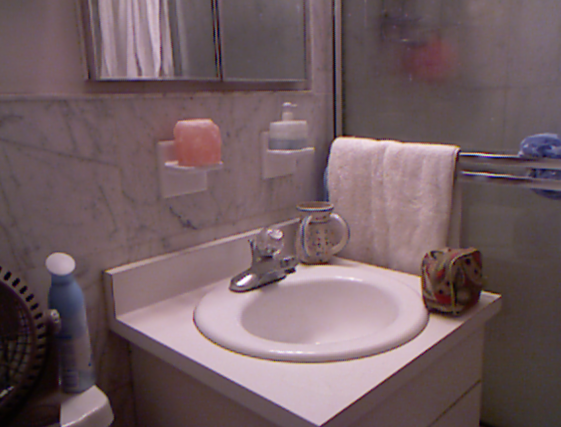} &
\includegraphics[width=0.142\linewidth]{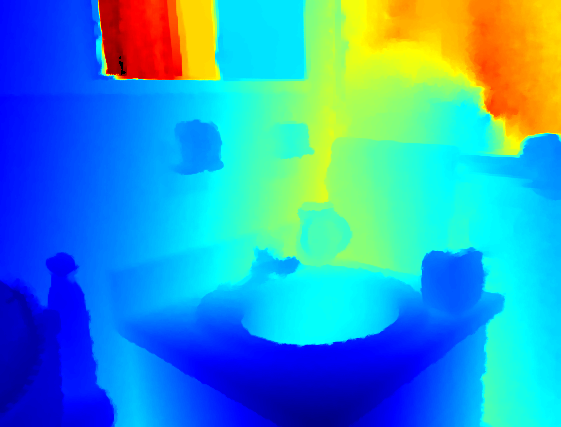}&
\includegraphics[width=0.142\linewidth]{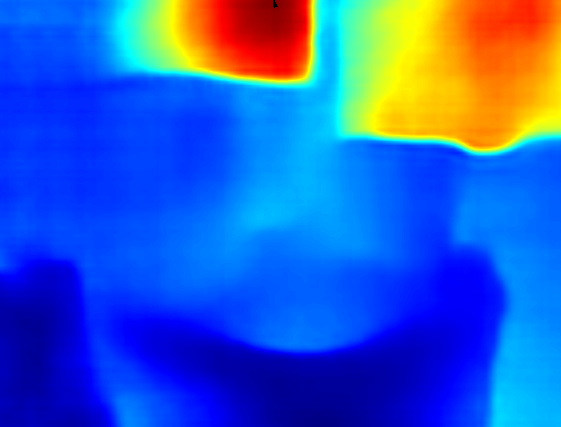}&
\includegraphics[width=0.142\linewidth]{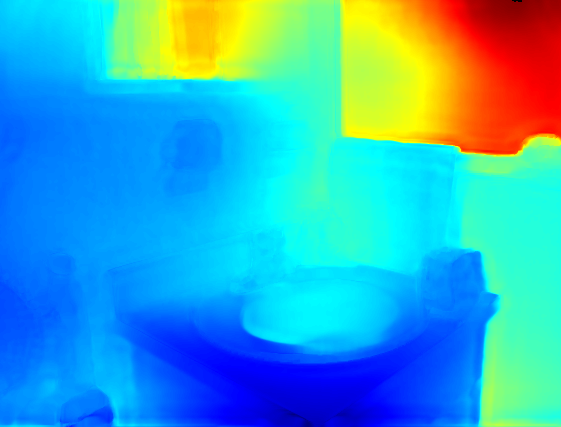}&
\includegraphics[width=0.142\linewidth]{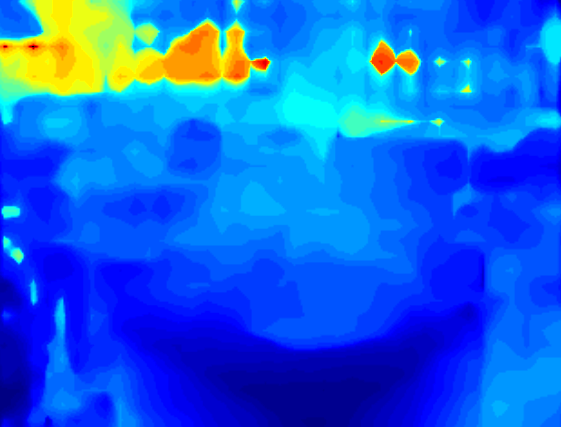}&
\includegraphics[width=0.142\linewidth]{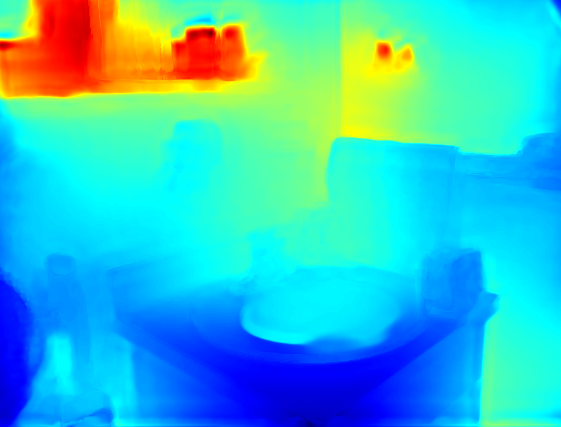}&
\includegraphics[width=0.142\linewidth]{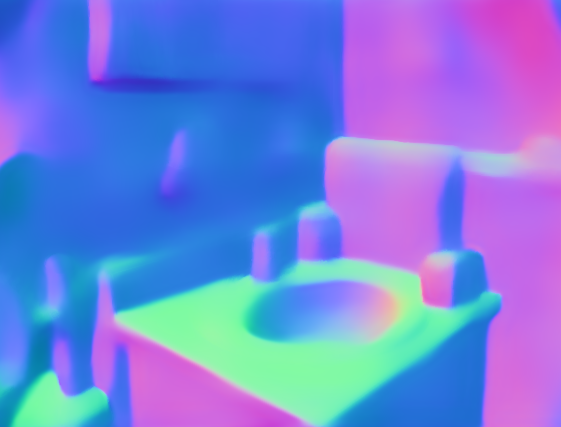}\\
\includegraphics[width=0.142\linewidth]{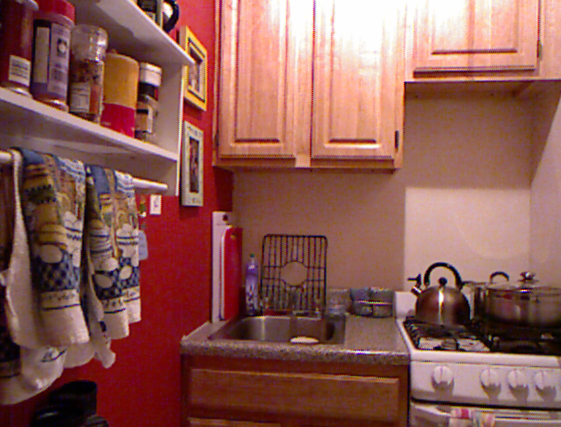} &
\includegraphics[width=0.142\linewidth]{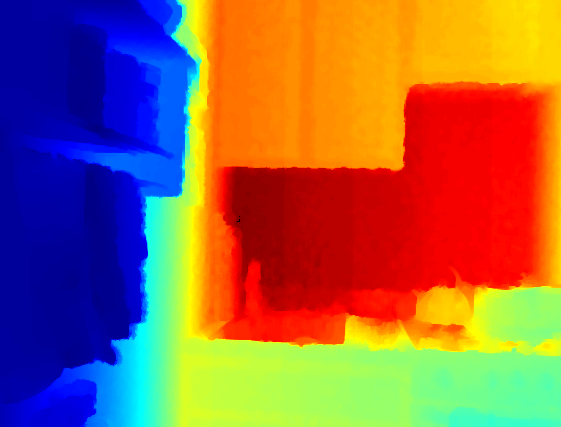}&
\includegraphics[width=0.142\linewidth]{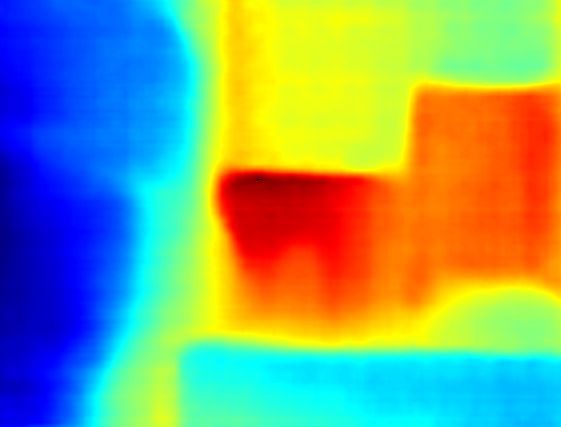}&
\includegraphics[width=0.142\linewidth]{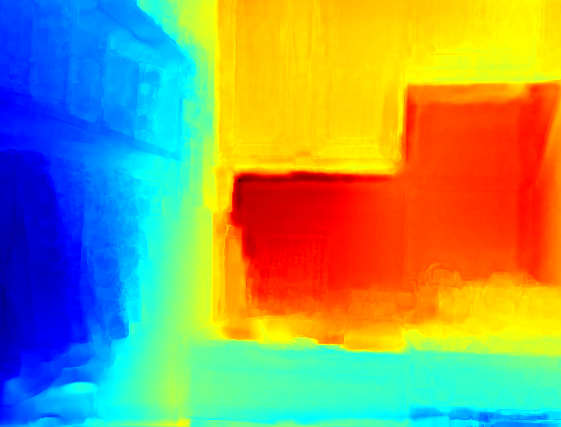}&
\includegraphics[width=0.142\linewidth]{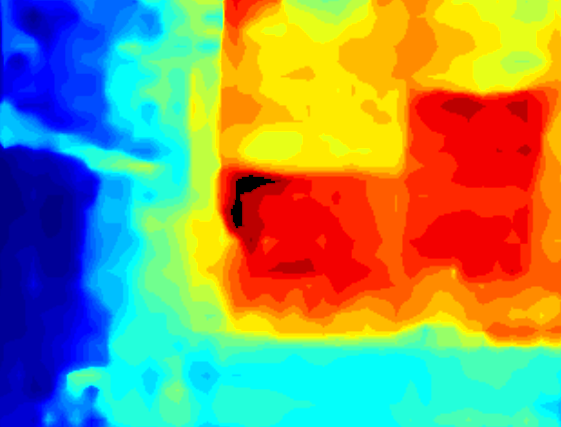}&
\includegraphics[width=0.142\linewidth]{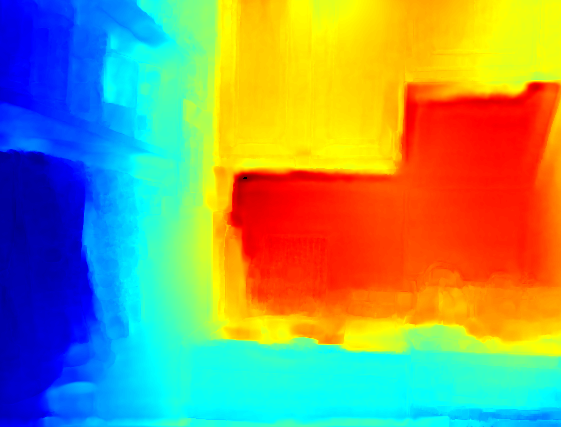}&
\includegraphics[width=0.142\linewidth]{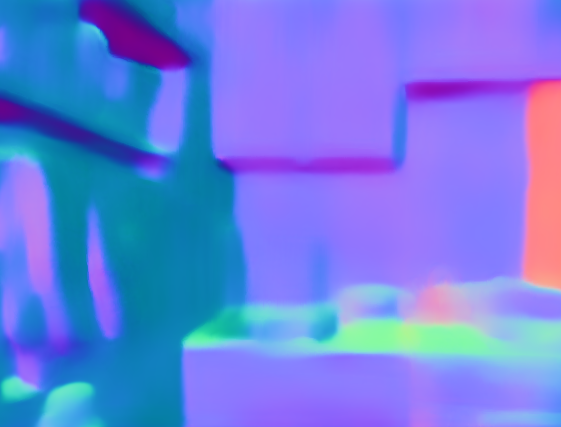}\\
\includegraphics[width=0.142\linewidth]{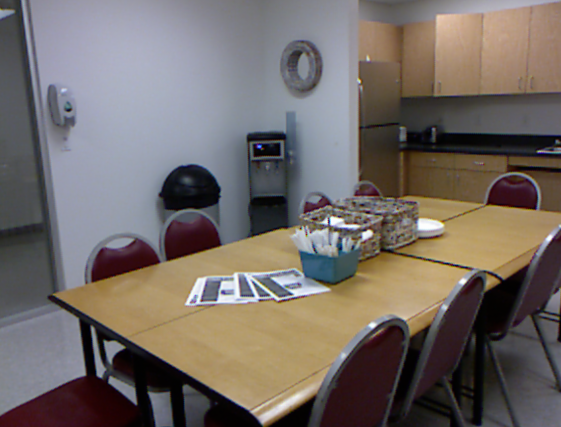} &
\includegraphics[width=0.142\linewidth]{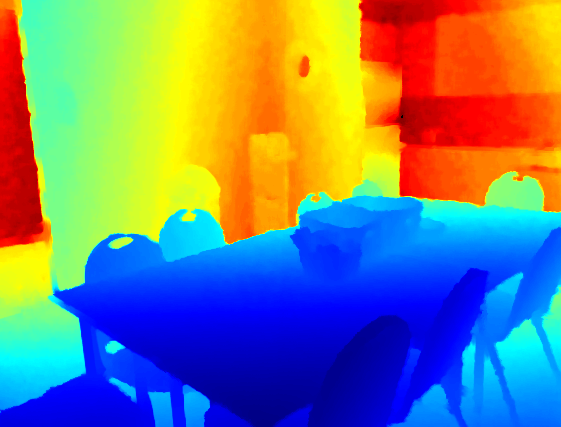}&
\includegraphics[width=0.142\linewidth]{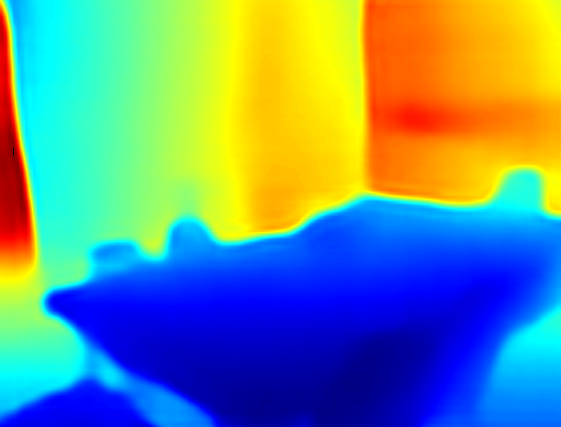}&
\includegraphics[width=0.142\linewidth]{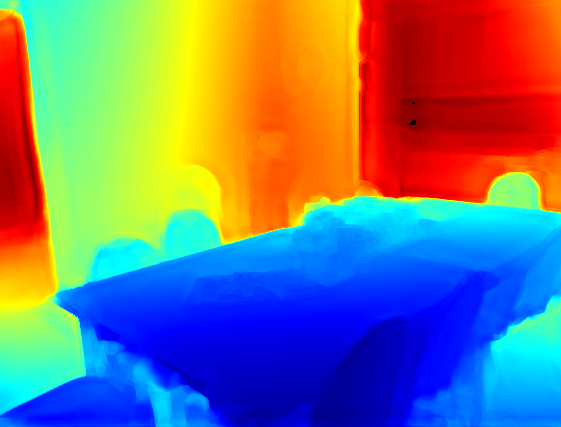}&
\includegraphics[width=0.142\linewidth]{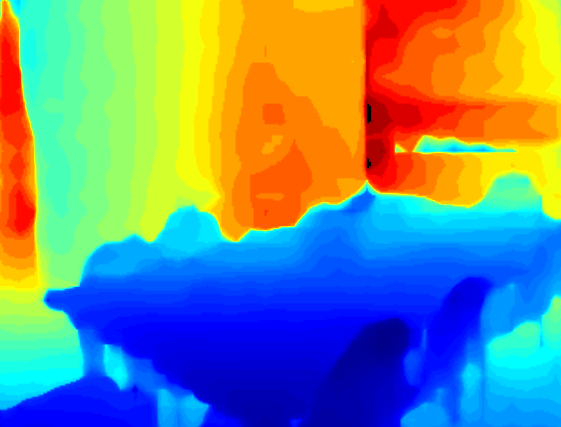}&
\includegraphics[width=0.142\linewidth]{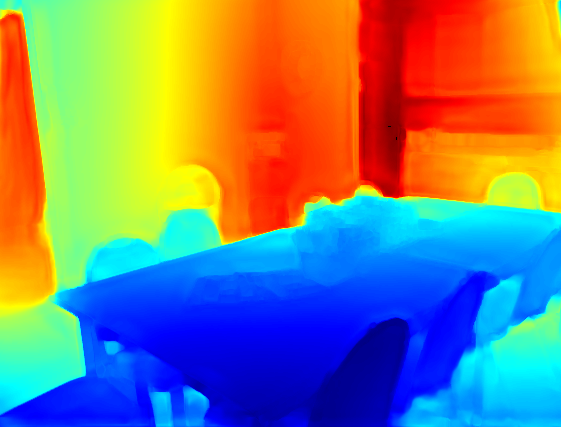}&
\includegraphics[width=0.142\linewidth]{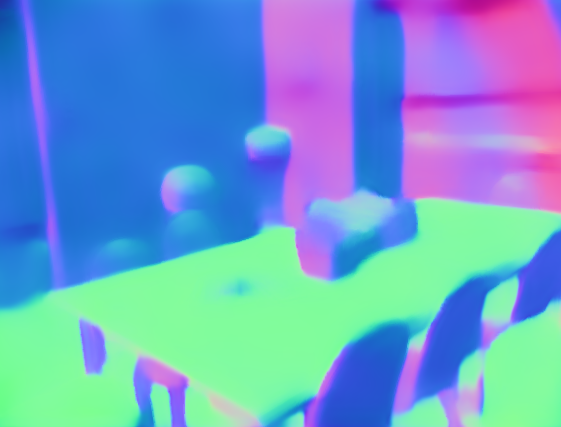}\\
\footnotesize{ (a) Images } &\footnotesize{(b) Ground truth} &\footnotesize{(c) FCRN~\cite{laina2016deeper}}&\footnotesize{ (d) FCRN~\cite{laina2016deeper} + Ours}&\footnotesize{ (e) DORN }&\footnotesize{(f) DORN + Ours} &\footnotesize{(g) Ours (normal)} \\
\end{tabular}
\caption{Visual illustrations on depth prediction. ``FCRN~\cite{laina2016deeper} + Ours'' indicates that we utilize FCRN~\cite{laina2016deeper} as the depth backbone in our system. ``DORN~\cite{laina2016deeper} + Ours'' indicates that DORN~\cite{laina2016deeper} is utilized as the depth backbone in our system. Please zoom in to see more details.}
\label{fig:vis-depth}
\end{figure*}

\vspace{0.1in}
\noindent\textbf{Depth prediction. }
In the task of depth prediction, we adopt VGG-16 which is most commonly adopted by state-of-the-art methods. As shown in Tab.~\ref{tab:com-depth}, our {\ourmodelshort} performs better than state-of-the-art approaches regarding all evaluation metrics with the VGG-16 architecture. Furthermore, {\ourmodelshort} outperforms {\ourmodelshortoriginal} due to its edge-aware refinement, ensemble modules, and iterative inference.
Among all these methods, SURGE~\cite{wang2016surge} is the only one that shares the same objective, {\ie}, jointly predicting depth and surface normal. It builds a CRF on top of a VGG-16 network.
As shown in Tab.~\ref{tab:com-depth}, 
using the same backbone network,  our {\ourmodelshort} significantly outperforms SURGE. We argue that this is due to the fact that our model does not impose  assumptions on the surface shape and the underlying geometry.
\textcolor{black}{Our model also performs favorably compared to modeling geometric constraint via loss function similar to ~\cite{yang2018unsupervised}.}
We also test the generalization ability of {\ourmodelshort} by directly taking the depth maps produced by {Multi-scale CRF~\cite{xu2017multi}}, Multi-scale CNN V2~\cite{eigen2015predicting}, {Local Network~\cite{chakrabarti2016depth}}, and DORN~\cite{fu2018deep} as the initial depth, and surface normals produced by our own baseline networks as the initial normal. Even without further fine-tuning, {\ourmodelshort} consistently improves all baseline results as shown in Tab.~\ref{tab:com-depth}. Note that when FCRN~\cite{chakrabarti2016depth} is end-to-end fine-tuned with {\ourmodelshort}, the results are further improved.

\begin{table*}
\centering
\caption{Performance of depth prediction on NYUD-V2 test set. ``Baseline'' means using VGG-16 to directly predict depth from raw images.
The backbone architecture for FCRN~\cite{laina2016deeper} and DORN~\cite{fu2018deep} are ResNet-50 and ResNet-101 respectively. The results for DORN~\cite{fu2018deep} are derived by evaluating their model.
* denotes that {\ourmodelshortoriginal} or {\ourmodelshort} with the backbone network are end-to-end finetuned. We do not finetune {\ourmodelshort} with DORN~\cite{fu2018deep} since the released Caffe code is not compatible with ours. \textcolor{black}{``Baseline + Loss'' indicates that we only use a geometry-aware loss function as~\cite{yang2018unsupervised}.}} 

\label{tab:com-depth}
\setlength{\tabcolsep}{10pt}
\scalebox{1.0}{
\begin{tabular}{l|c|ccc|ccc}
\toprule
\multirow{1}{*}{Backbone} &
\multirow{1}{*}{Method}&
\multicolumn{3}{c|}{{Error}}&
\multicolumn{3}{c}{{Accuracy}} \\
&{}& {{rmse}} & {{$\log10$}} & {{rel}} & {{$\delta<1.25$}} & {{$\delta<1.25^2$}} & {{$\delta<1.25^3$}} \\
\midrule
\midrule
&{DepthTransfer~\cite{karsch2012depth}}&1.214&{-}&0.349&0.447&0.745&0.897\\
&{SemanticDepth~\cite{ladicky2014pulling}}&-&-&-&0.542&0.829&0.941\\
&{DC-depth~\cite{liu2014discrete}}&1.06&0.127&0.335&-&-&-\\
AlexNet or&{Global-Depth~\cite{zhuo2015indoor}}&1.04&0.122&0.305&0.525&0.829&0.941\\
None-CNN&{NRF~\cite{roy2016monocular}}&{0.744} & {0.078} & {0.187} & {0.801} &{0.950} & {0.986} \\
&{GCL/RCL~\cite{baig2016coupled}} &{0.802}&{-}&{-}&{0.605}&{0.890}&{0.970}\\
&{CNN + HCRF~\cite{wang2015towards}}& {0.907} & {-} & {0.215} & {0.605} & {0.890} & {0.970} \\ \hline \hline
&{SURGE~\cite{wang2016surge}} & {0.643}&{-}&{0.156}&{0.768}&{0.951}&{0.989}\\
&{FCRN~\cite{laina2016deeper}} &{0.790}&{0.083}&{0.194}&{0.629}&{0.889}&{0.971}\\
&{Multi-scale CRF~\cite{xu2017multi}} &{0.688}&{0.073}&{0.175}&{0.741}&{0.934}&{0.982}\\
&{Multi-scale CNN V2~\cite{eigen2015predicting}}& {0.641} & {-} & {0.158} & {0.769} & {0.950} & {0.988} \\
&{Local Network~\cite{chakrabarti2016depth}}&{0.620}&{-}&{0.149}&{{0.806}}&{{0.958}}&{0.987}\\\cmidrule{2-8}
VGG&Multi-scale CNN V2~\cite{eigen2015predicting} + {\ourmodelshort}&0.637&0.067&0.157&0.772&0.951&0.988\\
&Multi-scale CRF~\cite{xu2017multi} + {\ourmodelshort}&{0.683}&{0.072}&{0.173}&{0.746}&{0.935}&{0.983}\\
&{Local Network~\cite{chakrabarti2016depth} + {\ourmodelshort}} &0.615&0.061&0.147&0.810&0.959&0.987\\
&Our Baseline  &{0.626}&{0.068}&{0.155}&{0.768}&{0.951}&{0.988}\\
&Our Baseline + {\ourmodelshortoriginal}*\cite{qi2018geonet}&{{0.608}} & {{0.065}}&{{0.149}} & {0.786} &{0.956} &{{0.990}}\\
& \textcolor{black}{Our Baseline + Loss \cite{yang2018unsupervised}} &{\textcolor{black}{0.615}} & {\textcolor{black}{0.065}}&{\textcolor{black}{0.150}} & {\textcolor{black}{0.782}} &{\textcolor{black}{0.954}} &{\textcolor{black}{0.989}} \\
&Our Baseline + {\ourmodelshort}*&\textbf{0.600}&\textbf{0.063}&\textbf{0.144}&\textbf{0.791}&\textbf{0.960}&\textbf{0.991}\\ \hline \hline
&{Multi-scale CRF~\cite{xu2017multi}}&{0.586}& {{0.052}}& {{0.121}}&{0.811}&{0.954}&{0.988}\\
&{FCRN~\cite{laina2016deeper}} &{0.584}&{0.059}&{0.136}&{0.822}&{0.955}&{0.971}\\
ResNet&{DORN~\cite{fu2018deep} }&{0.552}& {{0.051}}& {{0.115}}&{0.826}&{0.960}&{0.985}\\\cmidrule{2-8}
&FCRN~\cite{laina2016deeper} + {\ourmodelshort} & {0.575}&0.058&0.134&{0.828}&{0.957}&\textbf{0.989}\\
&FCRN~\cite{laina2016deeper} + {\ourmodelshort}* & {0.558}&0.055&0.129&{0.839}&{0.960}&\textbf{0.990}\\

&DORN~\cite{fu2018deep} + {\ourmodelshort} & \textbf{0.527} &\textbf{0.049}&\textbf{0.113}&\textbf{0.862}&\textbf{0.965}&{0.989}\\
\bottomrule
\end{tabular}}
\end{table*}

\begin{figure*}
\centering
{\footnotesize
\begin{tabular}{@{\hspace{0.1mm}}c@{\hspace{0.1mm}}c@{\hspace{0.1 mm}}c@{\hspace{0.1 mm}}c@{\hspace{0.1 mm}}c@{\hspace{0.1 mm}}c@{\hspace{0.1 mm}}c@{\hspace{0.1 mm}}c@{\hspace{0.1 mm}}c}
\includegraphics[width=0.110\linewidth]{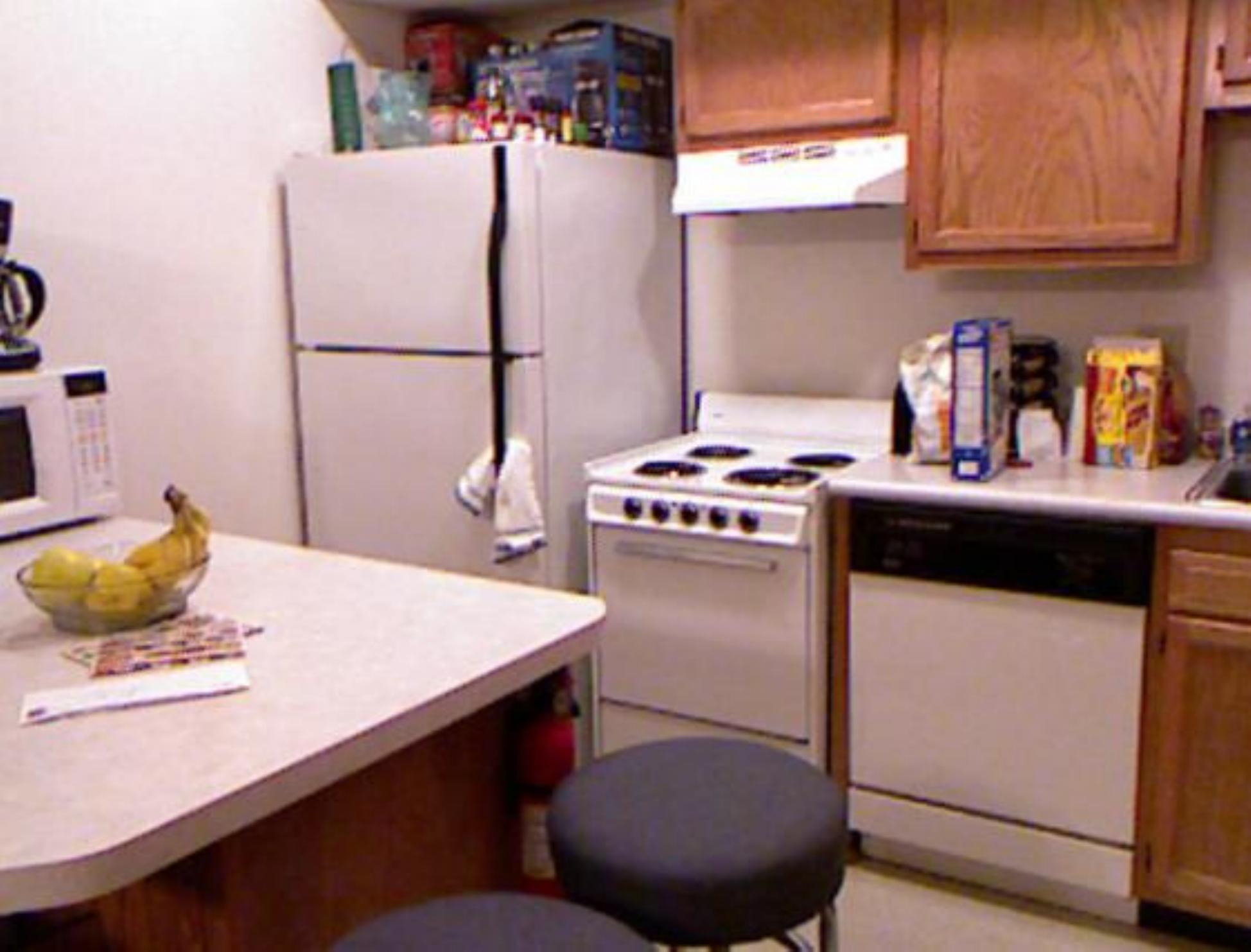} &
\includegraphics[width=0.110\linewidth]{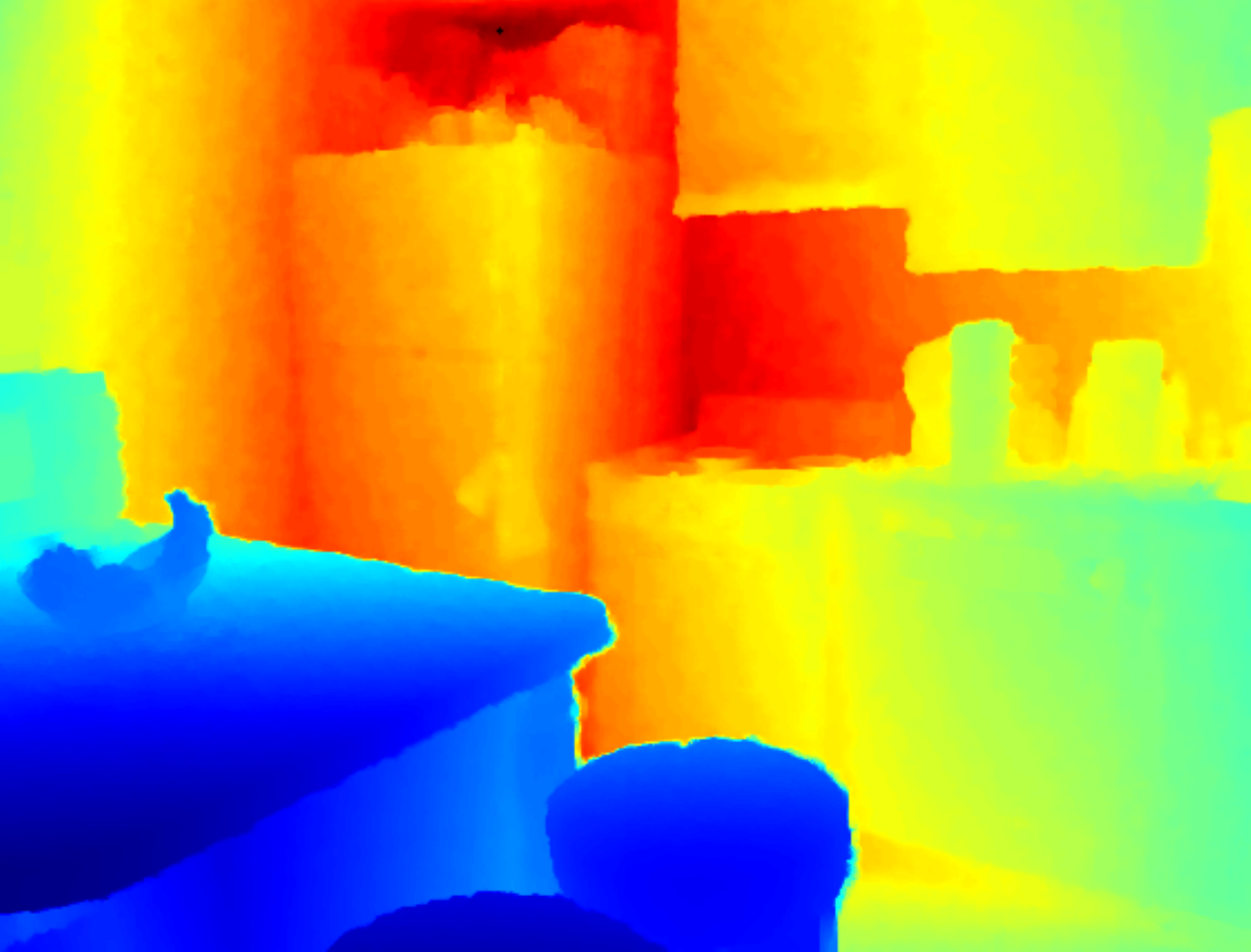}&
\includegraphics[width=0.110\linewidth]{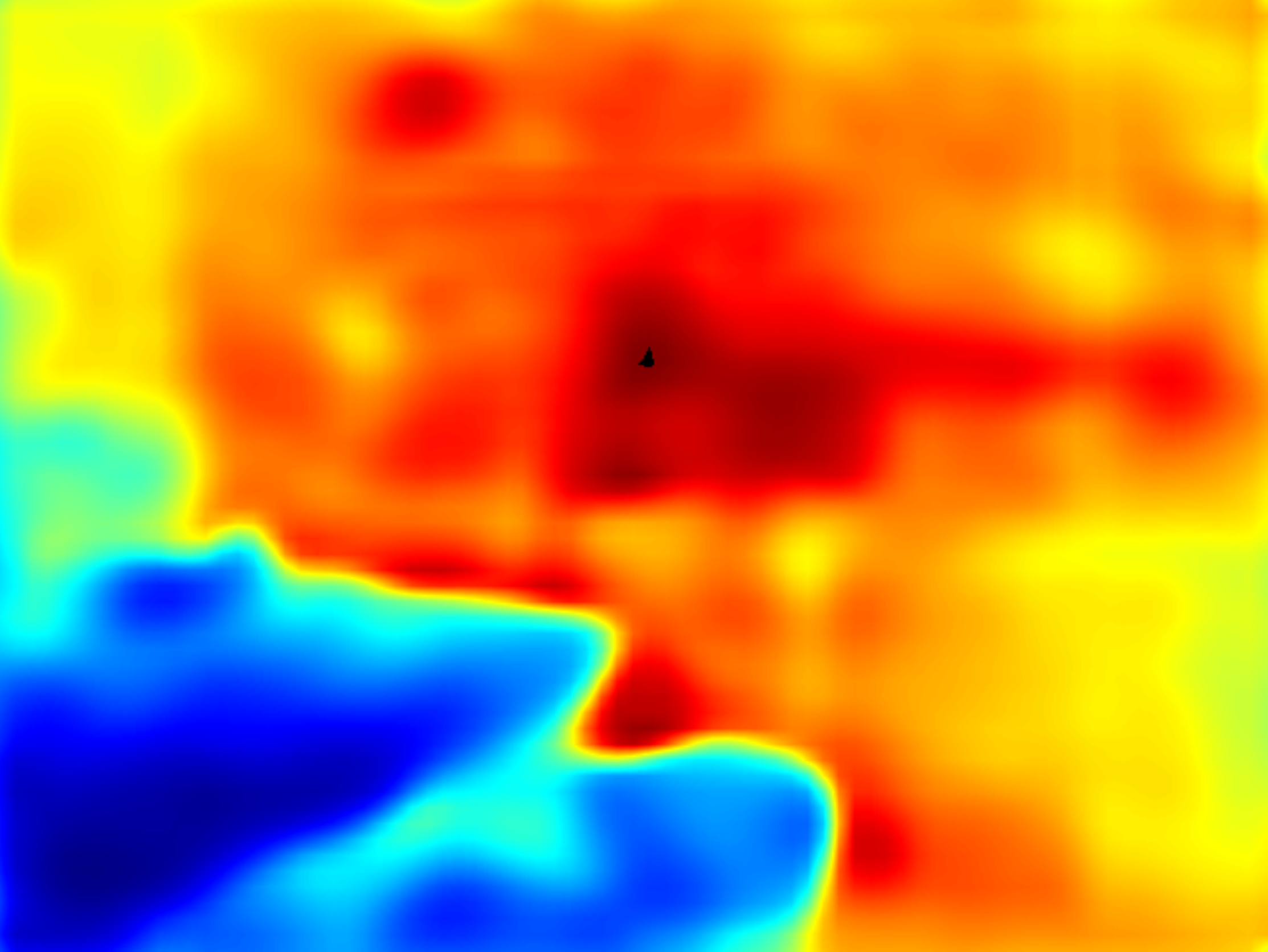}&
\includegraphics[width=0.110\linewidth]{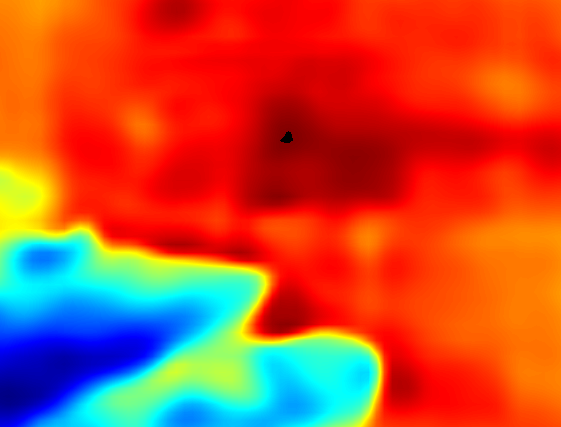}&
\includegraphics[width=0.110\linewidth]{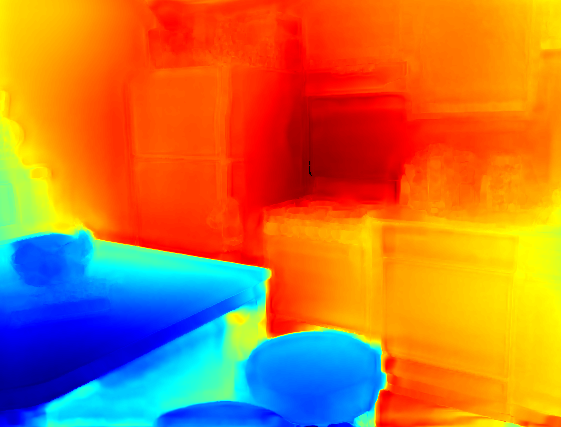}&
\includegraphics[width=0.110\linewidth]{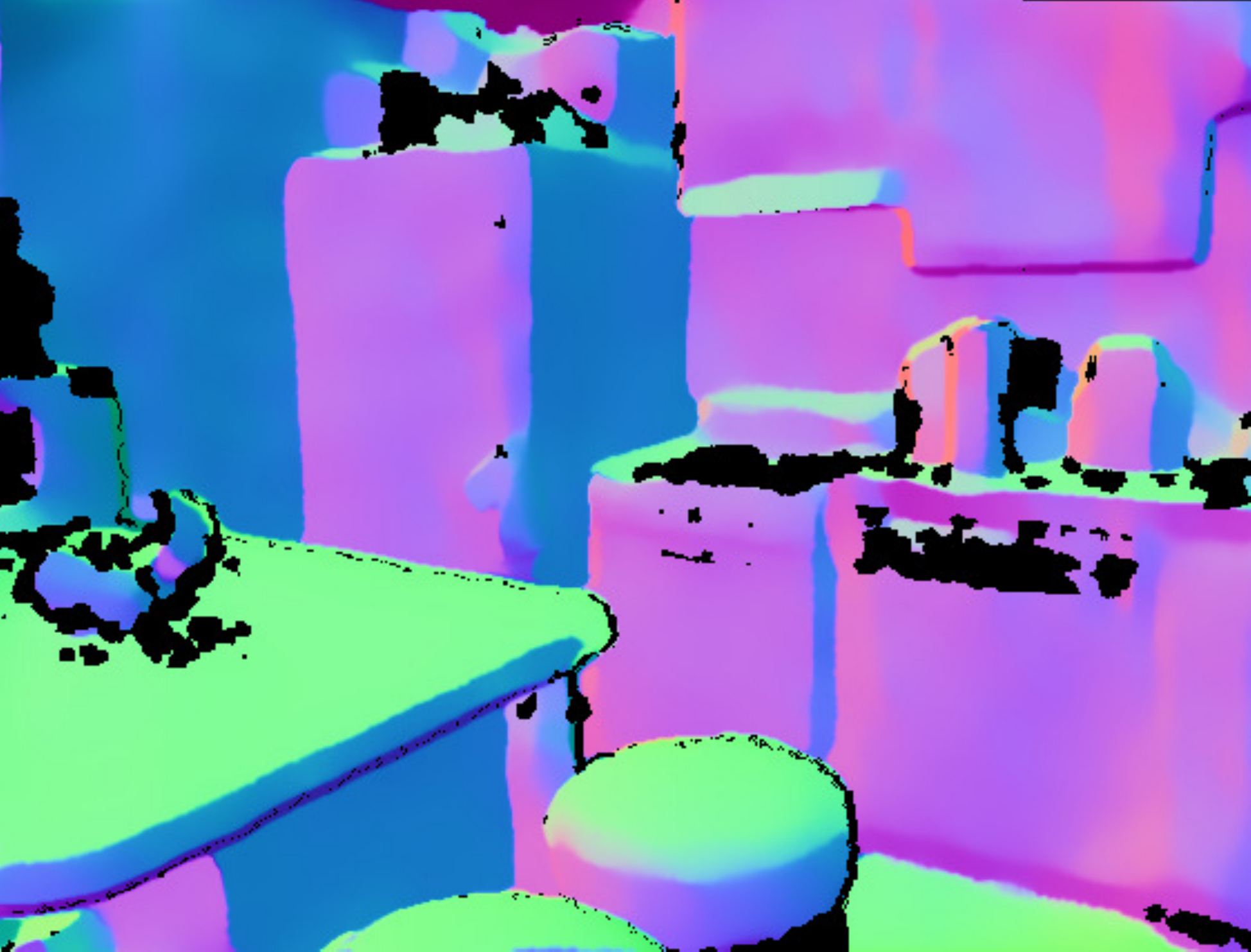}&
\includegraphics[width=0.110\linewidth]{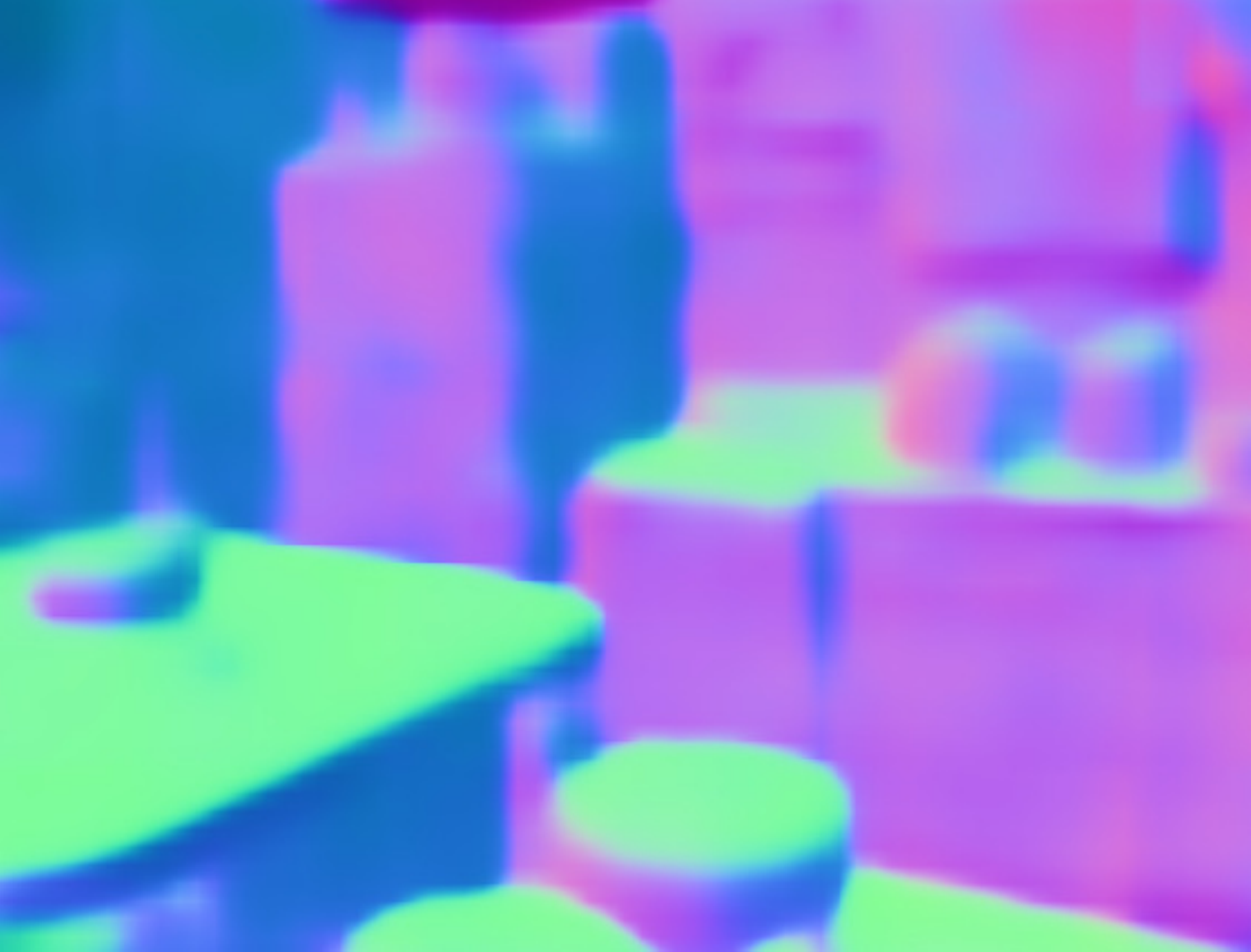} &
\includegraphics[width=0.110\linewidth]{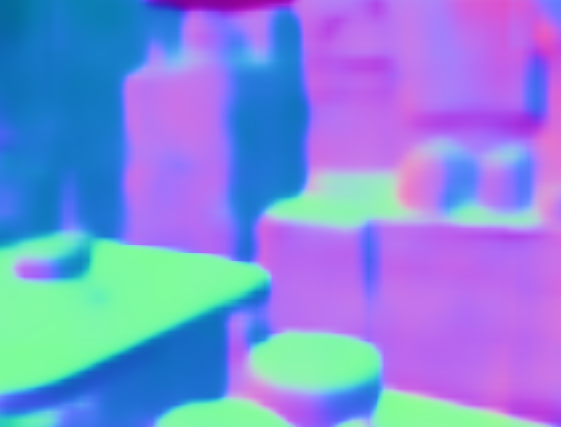} &
\includegraphics[width=0.110\linewidth]{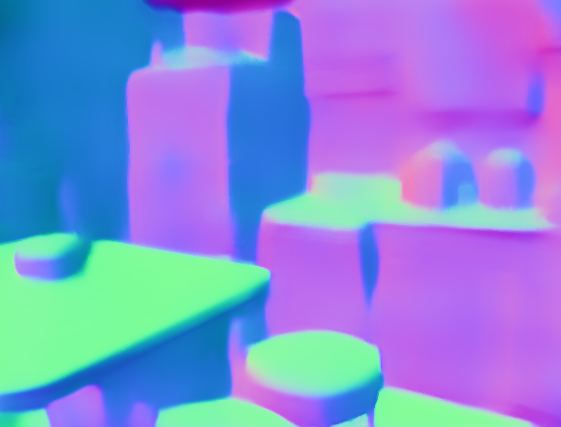}\\
\includegraphics[width=0.110\linewidth]{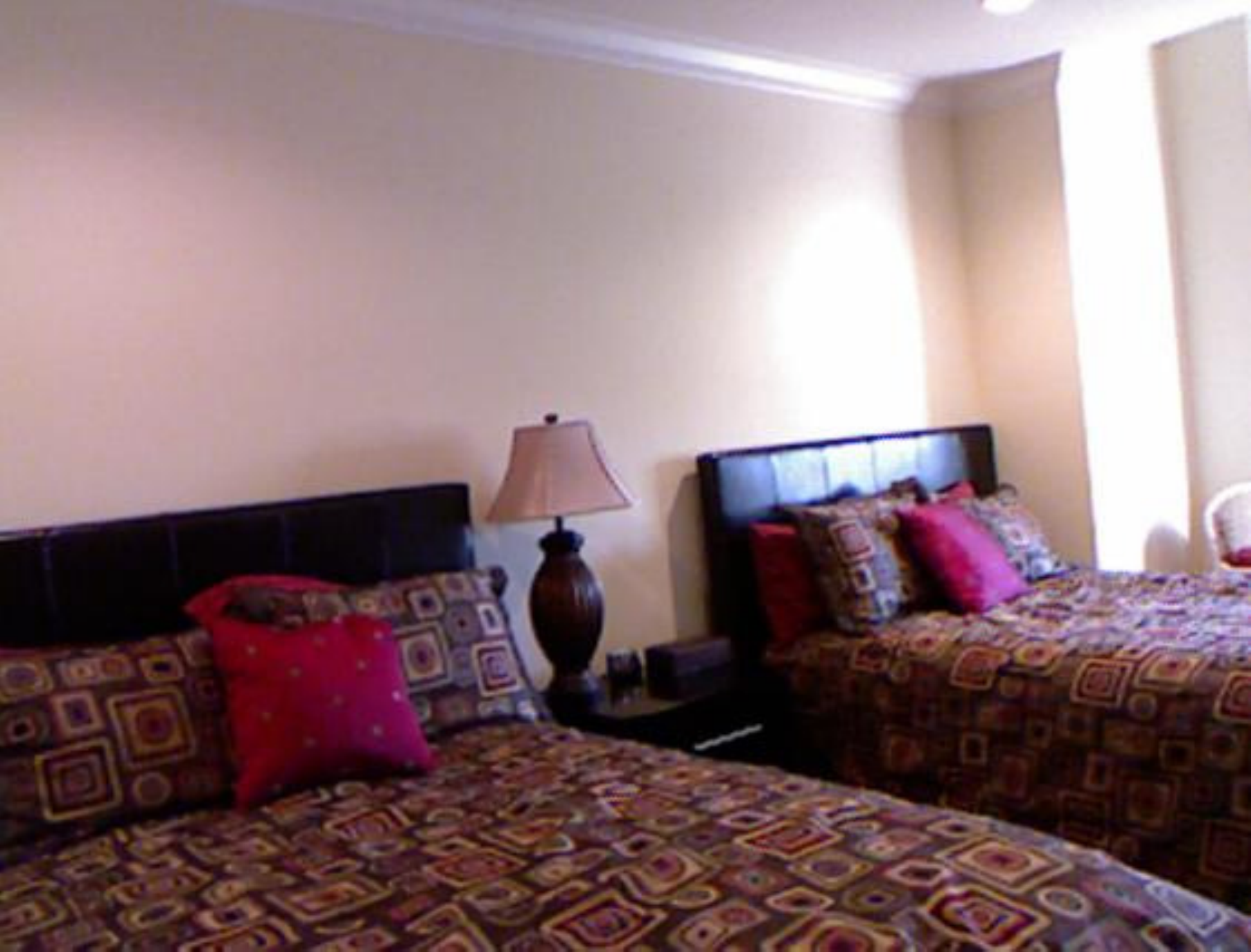} &
\includegraphics[width=0.110\linewidth]{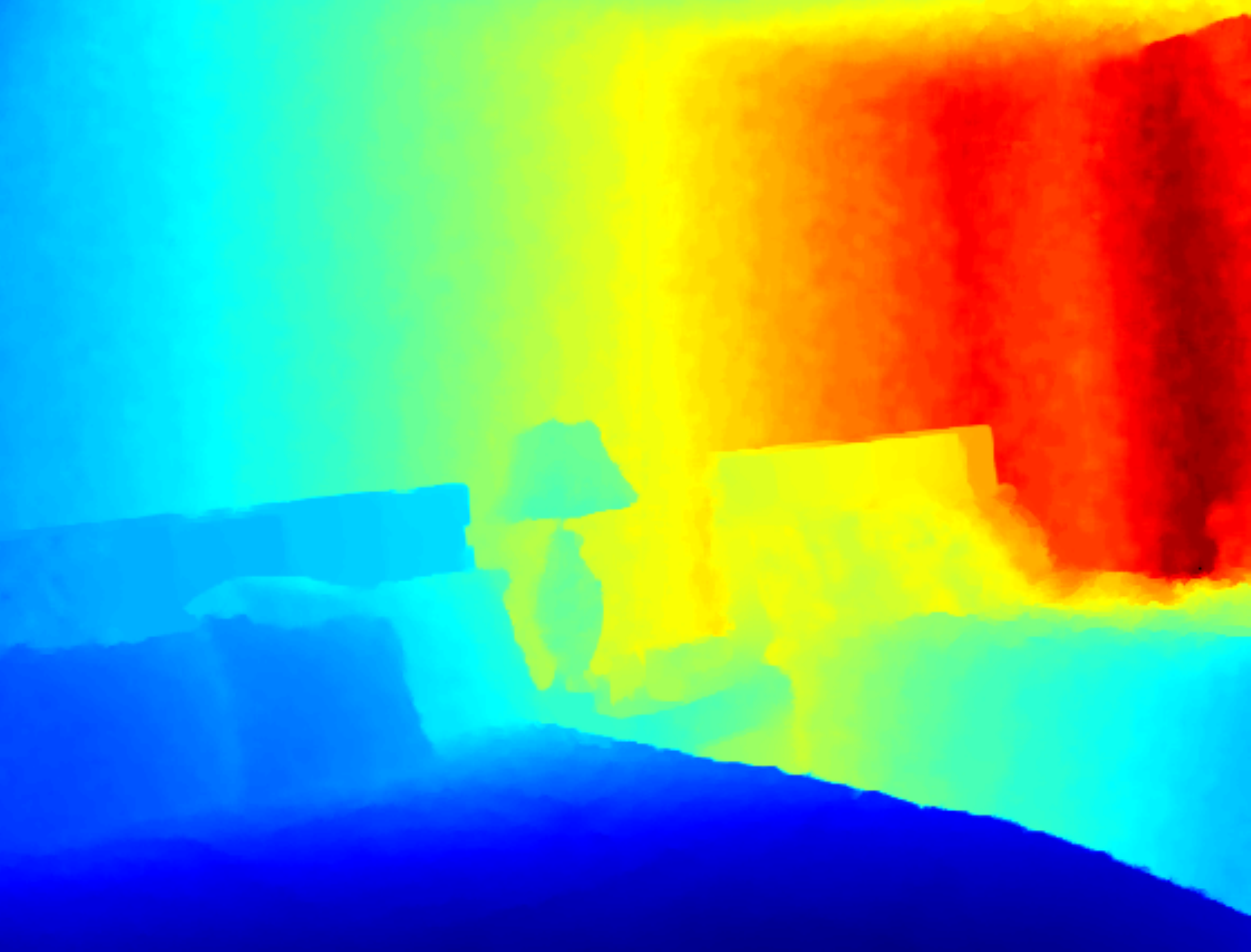}&
\includegraphics[width=0.110\linewidth]{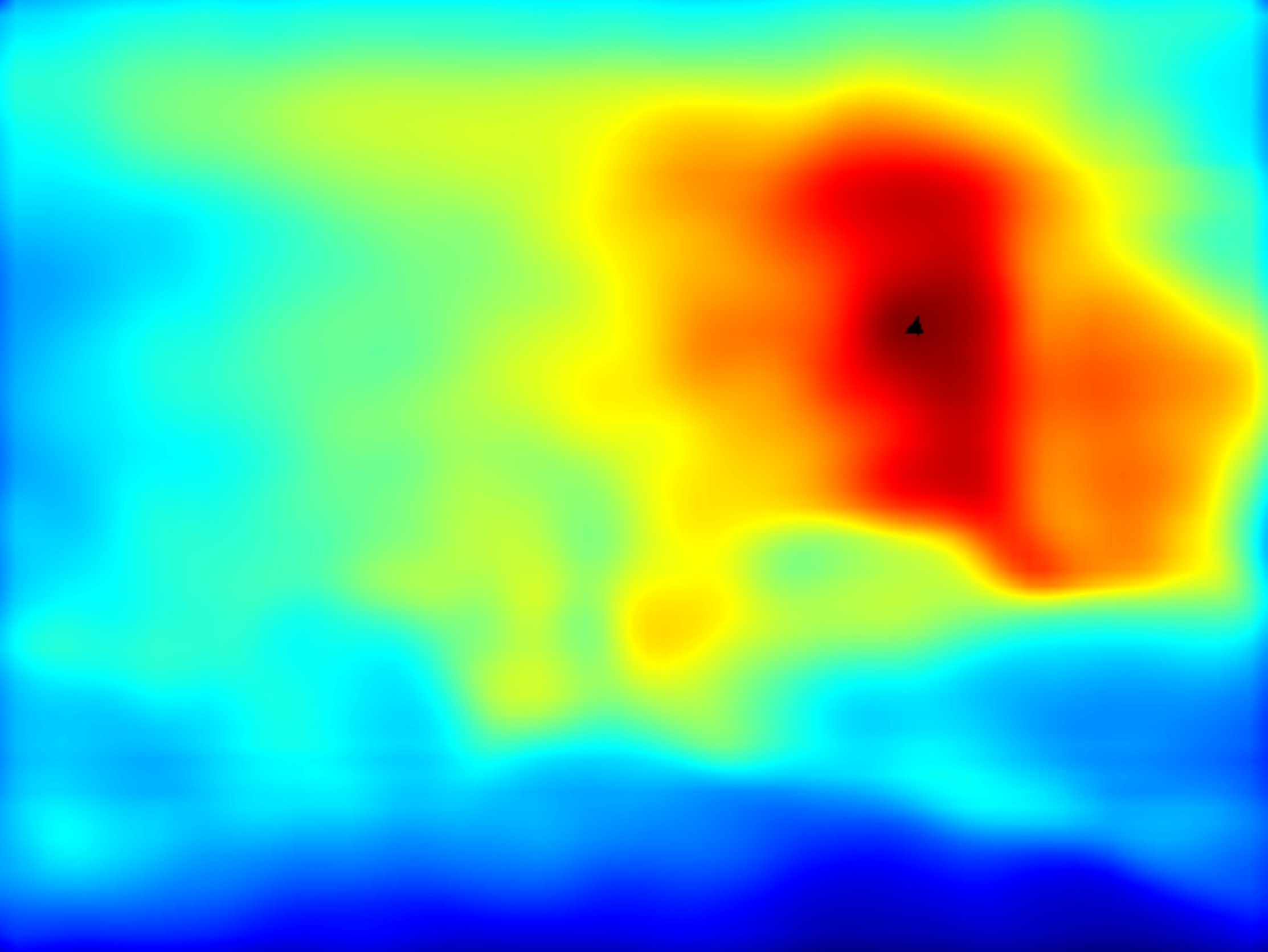}&
\includegraphics[width=0.110\linewidth]{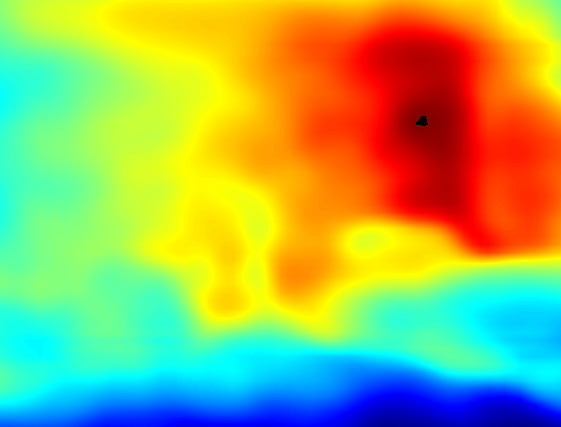}&
\includegraphics[width=0.110\linewidth]{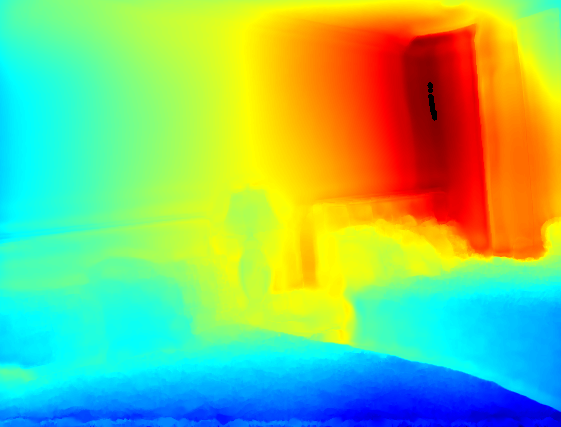}&
\includegraphics[width=0.110\linewidth]{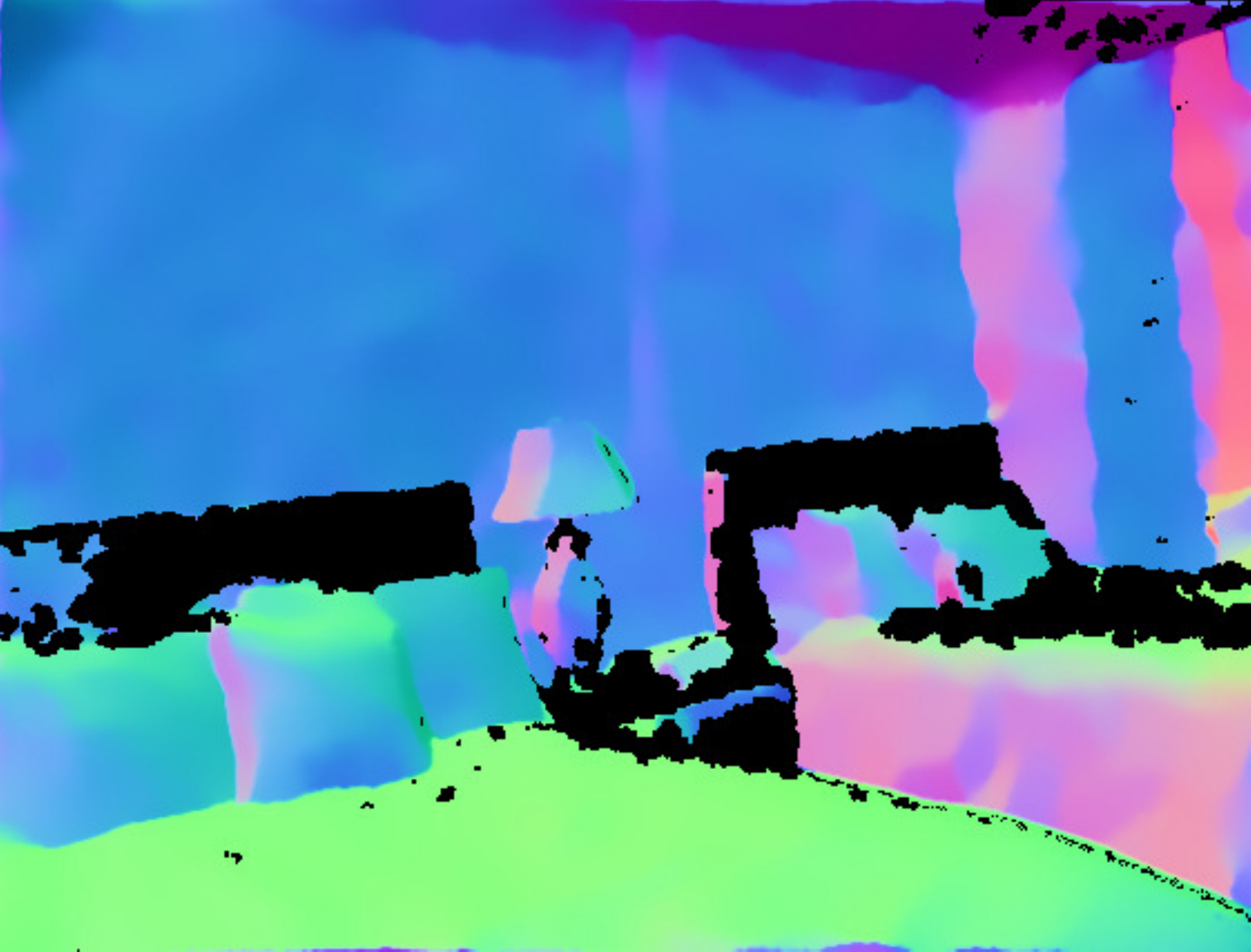}&
\includegraphics[width=0.110\linewidth]{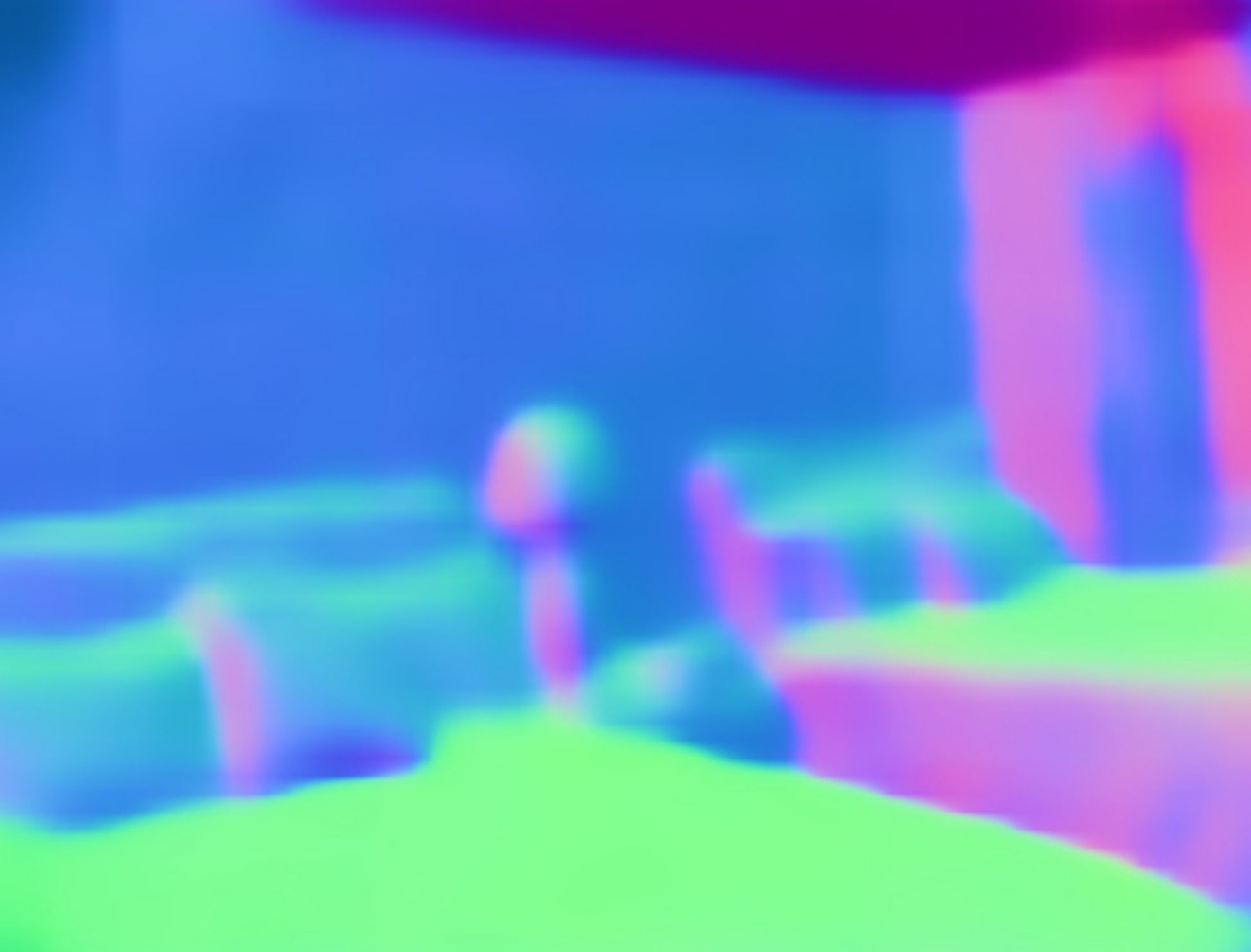} &
\includegraphics[width=0.110\linewidth]{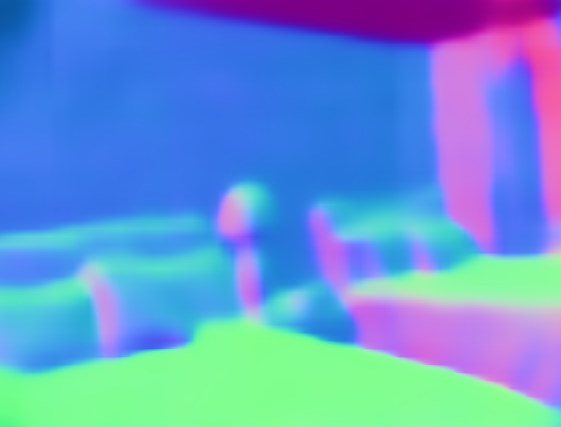} &
\includegraphics[width=0.110\linewidth]{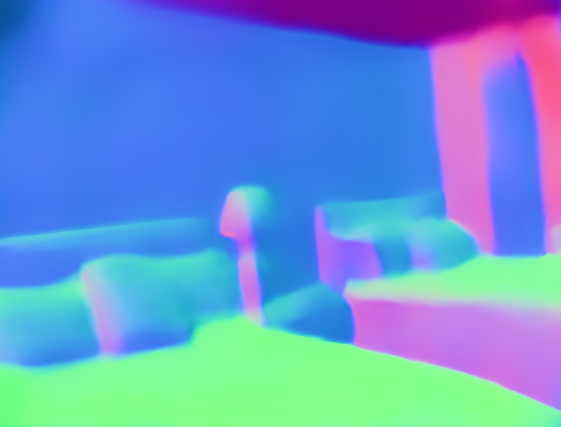}\\
\includegraphics[width=0.110\linewidth]{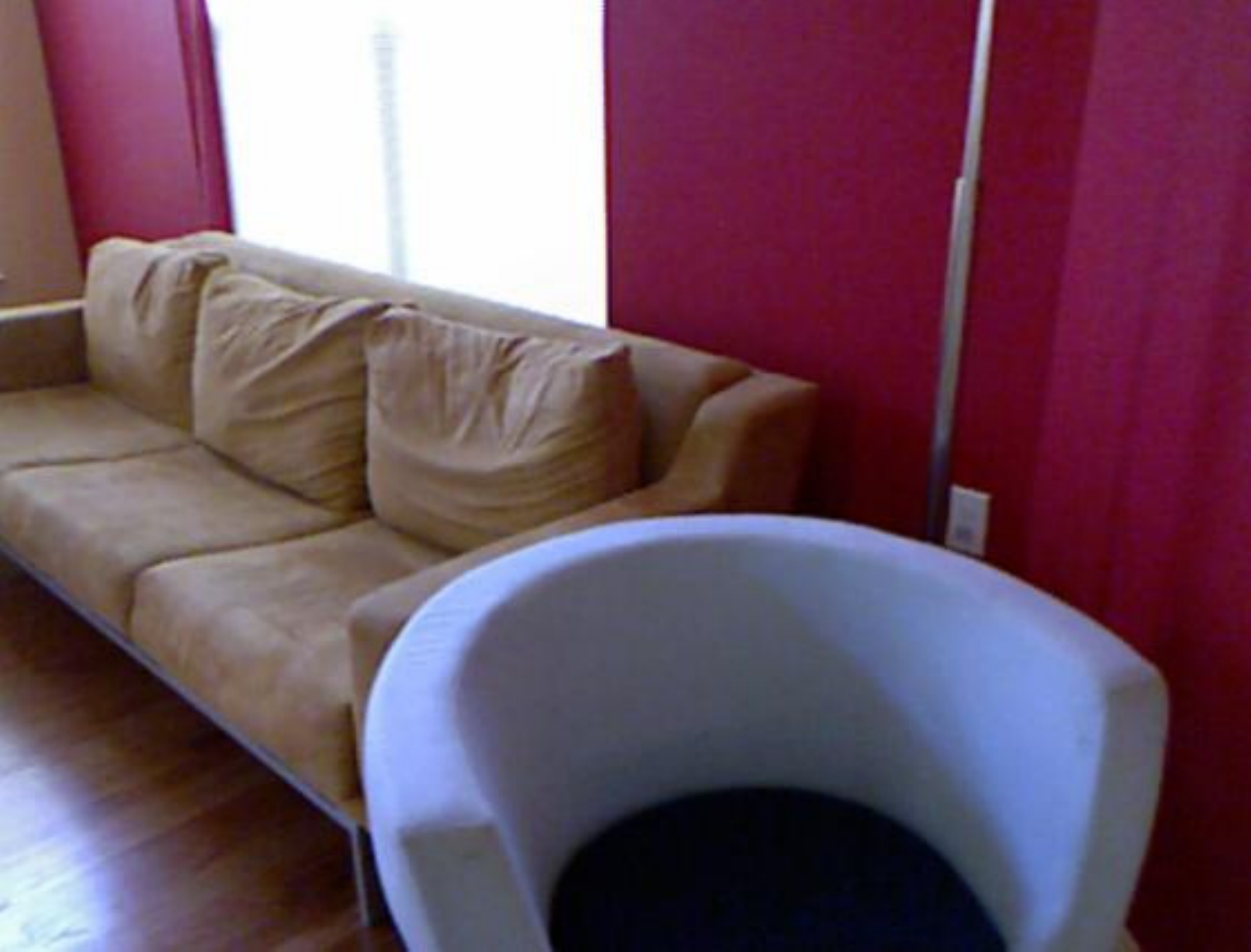} &
\includegraphics[width=0.110\linewidth]{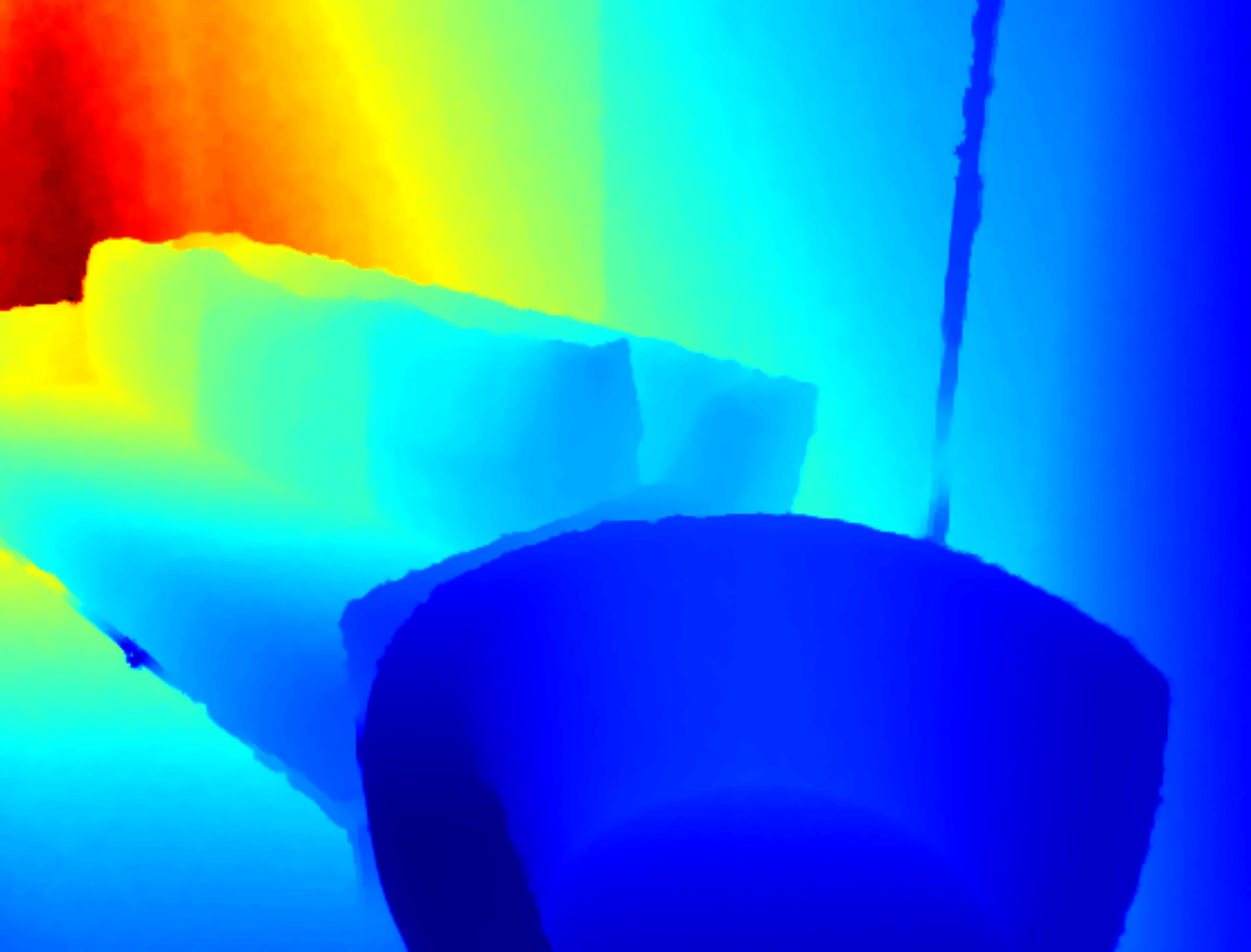}&
\includegraphics[width=0.110\linewidth]{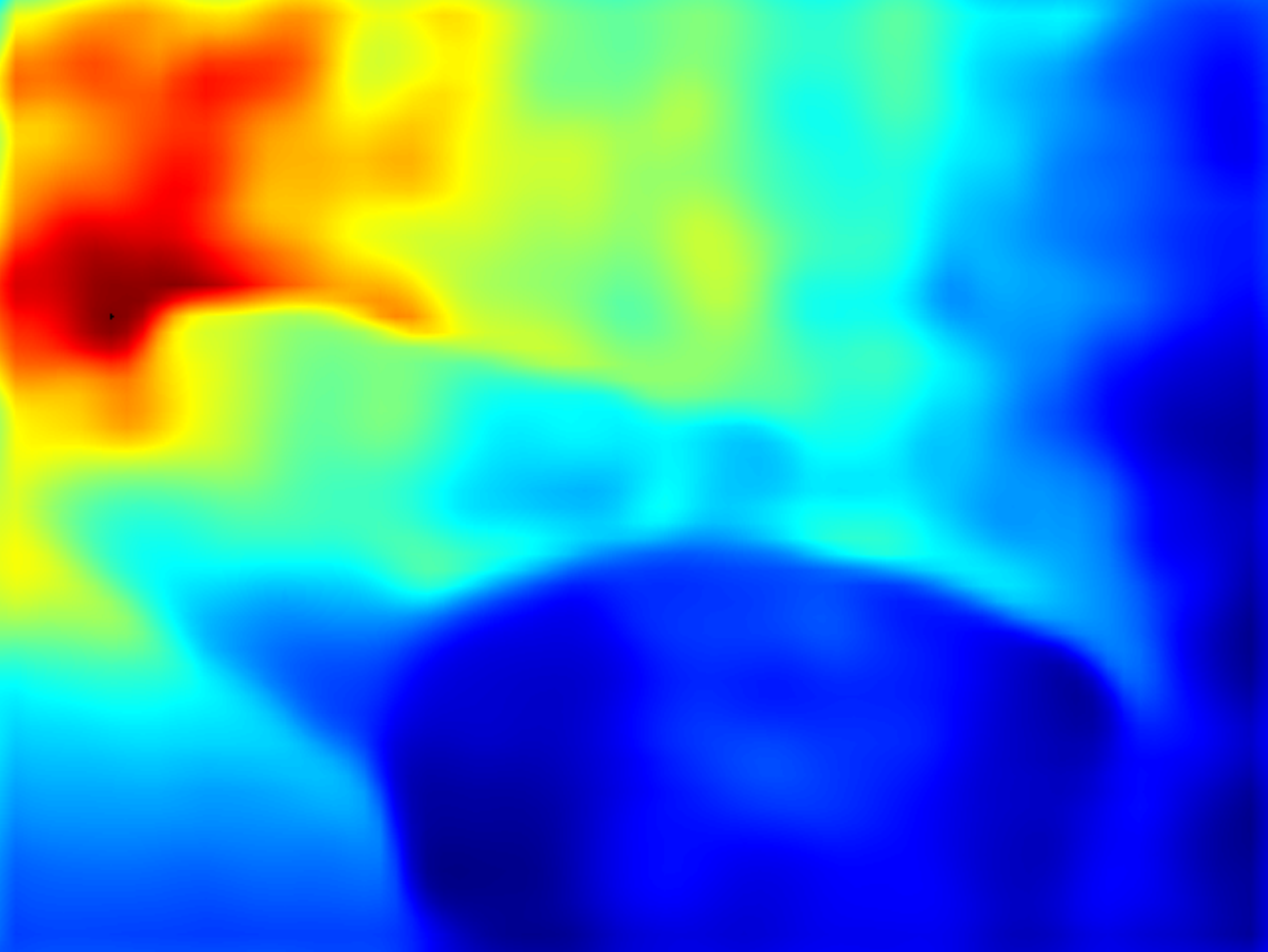}&
\includegraphics[width=0.110\linewidth]{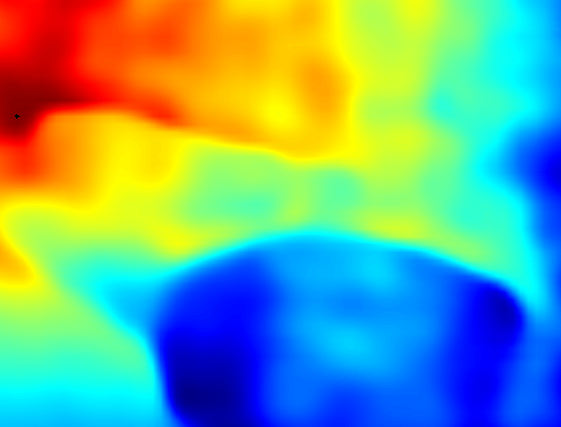}&
\includegraphics[width=0.110\linewidth]{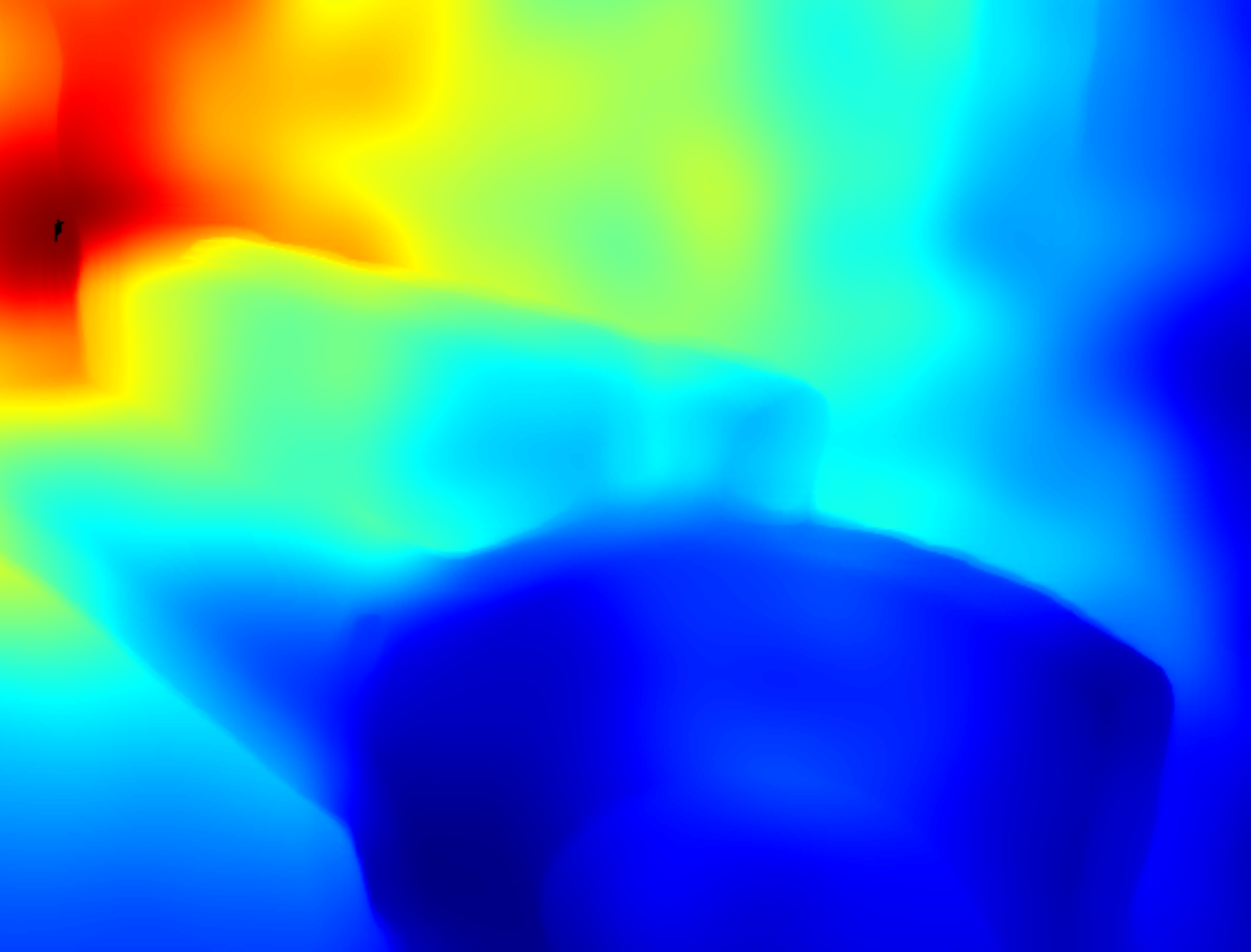}&
\includegraphics[width=0.110\linewidth]{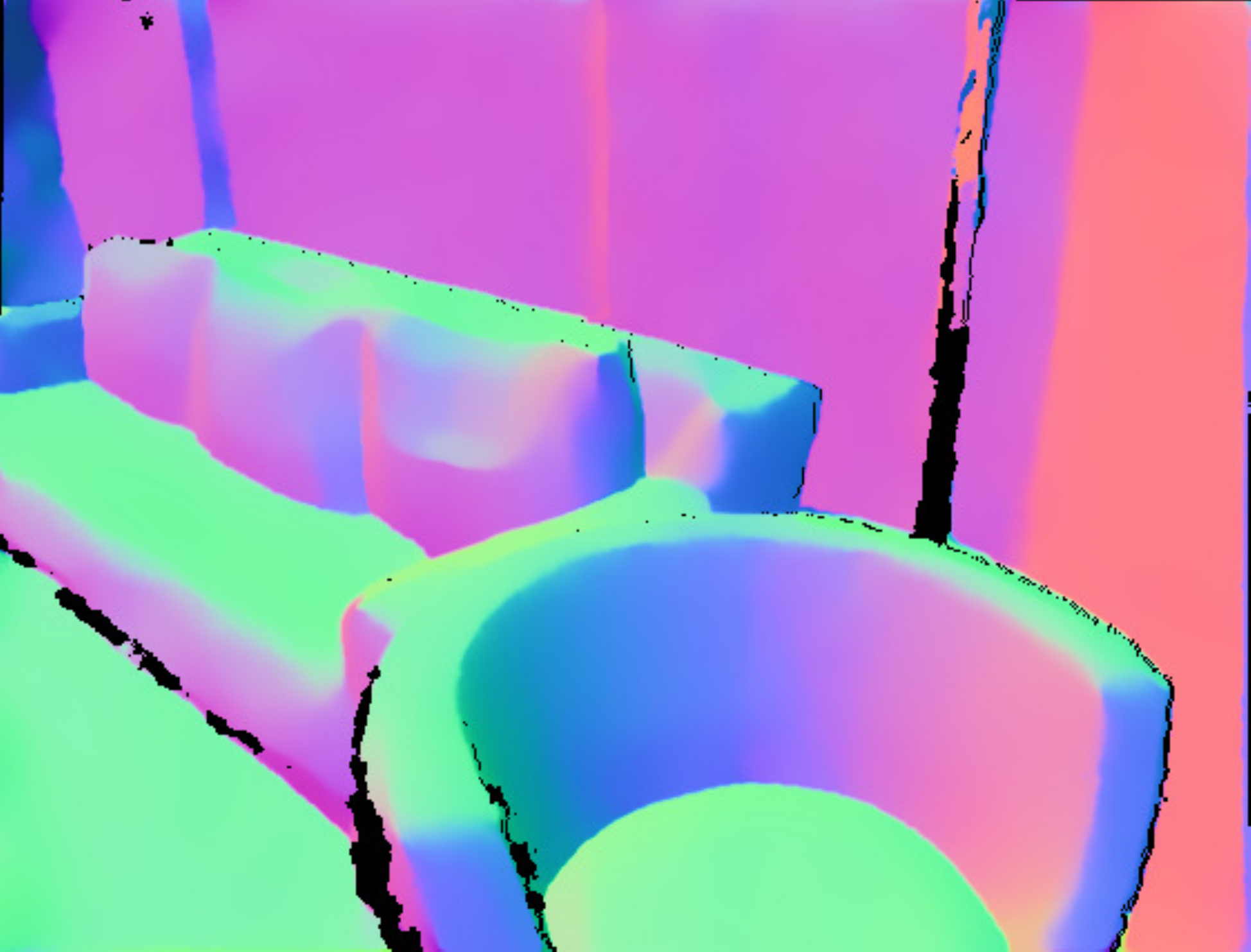}&
\includegraphics[width=0.110\linewidth]{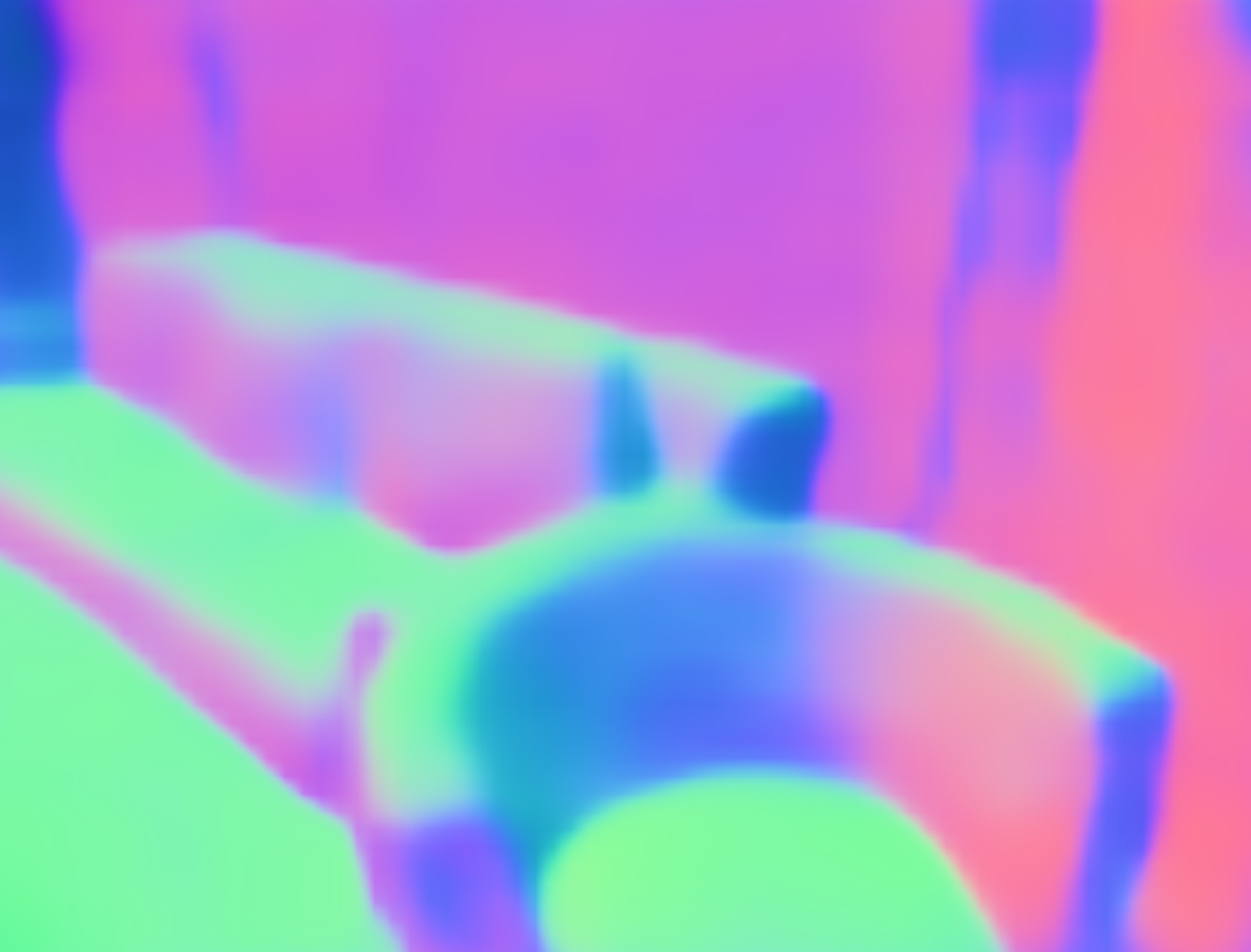} &
\includegraphics[width=0.110\linewidth]{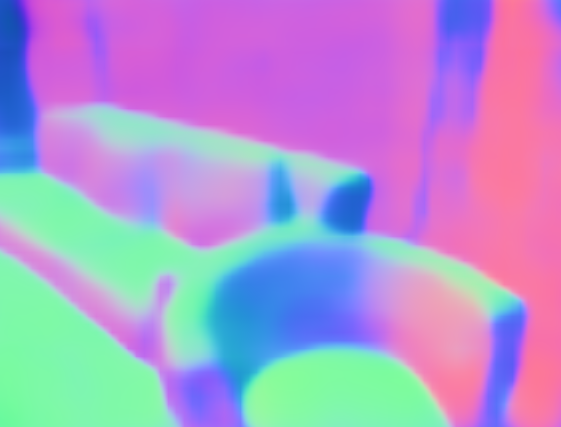} &
\includegraphics[width=0.110\linewidth]{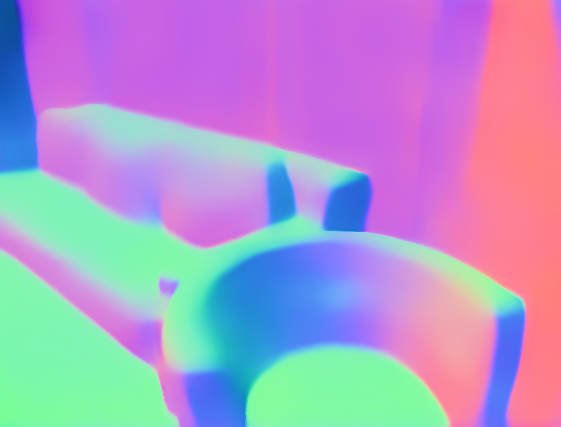}\\
{ (a) Image } &{(b) GT (D)} &{(c) VGG (D)}& \textcolor{black}{(d) Loss}&{ (e) Ours (D)} &{(f) GT (N) } &{(g) VGG (N) } &\textcolor{black}{(h) Loss (N)}&{ (i) Ours (N)} \\
\end{tabular}}
\caption{Visual comparisons on joint prediction of depth and surface normal with VGG-16 as the backbone architecture. GT stands for ``ground truth". 
\textcolor{black}{``(D)" and ``(N)" represent depth and surface normal respectively. Loss indicates that the geometric constraint is only adopted in constructing the loss.
.}
}
\label{fig:vis-joint}
\end{figure*}

\vspace{0.1in}
\noindent\textbf{Visual comparison.~~}
Fig.~\ref{fig:vis-depth} depicts a visual comparison with FCRN~\cite{laina2016deeper} and DORN~\cite{fu2018deep} on depth prediction.
Our {\ourmodelshort} generates more accurate depth maps with regard to the washbasin and small objects on the table in the $2$-nd and $4$-th rows respectively.
We also show the corresponding surface normal predictions to verify that our {\ourmodelshort} takes advantage of them to improve the depth prediction. 
We refer the reader to look closely at the wall in the $1$-st row of the figure.
DORN~\cite{fu2018deep} achieves decent depth prediction performance on the NYUD-V2 dataset. However, the visual quality of results still has much room for improvement since they are not piecewise smooth in planar regions. 
When the produced depth is refined by {\ourmodelshort}, the visual quality is significantly improved as illustrated in Fig.~\ref{fig:vis-depth} (columns e and f).
We further compare our normal prediction results with those of other methods, including Deep3D~\cite{wang2015designing}, Multi-scale CNN V2~\cite{eigen2015predicting}, and SkipNet~\cite{bansal2016marr} in Fig.~\ref{fig:vis-normal}.
{\ourmodelshort} produces results with nice details on, {\eg}, the chair, washbasin, and wall from the $1$-st, $2$-nd, and $3$-rd rows respectively. More results of joint prediction are shown in Fig.~\ref{fig:vis-joint}.
From these figures, it is clear that our model does a better job than previous approaches in terms of geometry estimation.

\begin{table*}
\centering
\caption{Quantitative comparisons of depth predictions on the KITTI dataset. * denotes that we directly evaluate results or model released by respective authors. ``Multi-scale CNN V1*~\cite{eigen2014depth} + {\ourmodelshort}'' indicates that we utilize Multi-scale CNN V1*~\cite{eigen2014depth} to produce the initial depth. ``DORN~\cite{fu2018deep} + {\ourmodelshort}'' represents that we utilize DORN~\cite{fu2018deep} to produce the initial depth.}
\label{tab:com-depth-kitti}
\setlength{\tabcolsep}{10pt}
\scalebox{1.0}{
\begin{tabular}{c|ccc|ccc}
\toprule
\multirow{1}{*}{Method}&
\multicolumn{3}{c|}{{Error}}&
\multicolumn{3}{c}{{Accuracy}} \\
& {{rmse}} & {{$\log10$}} & {{rel}} & {{$\delta<1.25$}} & {{$\delta<1.25^2$}} & {{$\delta<1.25^3$}} \\
\midrule
\midrule
Make3D~\cite{saxena2009make3d}&8.73&0.361&0.280&0.601&0.820&0.926\\
Dis-cont Depth~\cite{liu2016learning}&6.99&0.289&0.217&0.647&0.882&0.961\\
LRC~\cite{godard2017unsupervised}&4.94&0.206&0.114&0.861&0.949&0.976\\
Semi-depth~\cite{kuznietsov2017semi}&4.62&0.189&0.113&0.862&0.960&0.986\\
Multi-scale CRF~\cite{xu2017multi}&4.69&-&0.125&0.816&0.951&0.983\\
Multi-scale CNN V1*~\cite{eigen2014depth}&8.03&0.337&0.350&0.474&0.827&0.945\\
DORN~\cite{fu2018deep}*&\textbf{4.06}&0.175&0.101&0.891&0.965&\textbf{0.986}\\\midrule
Multi-scale CNN V1*~\cite{eigen2014depth} + {\ourmodelshort}&7.96&0.321&0.341&0.500&0.838&0.948\\
DORN~\cite{fu2018deep} + {\ourmodelshort} & {4.10} &\textbf{0.172}&\textbf{0.094}&\textbf{0.897}&\textbf{0.968}&\textbf{0.986}\\
\bottomrule
\end{tabular}}
\end{table*}

\subsection{Experiments on the KITTI Dataset}
\begin{figure*}[t]
\centering
\begin{tabular}{@{\hspace{0.1mm}}c@{\hspace{0.1mm}}c@{\hspace{0.1mm}}c}
\includegraphics[width=0.33\linewidth]{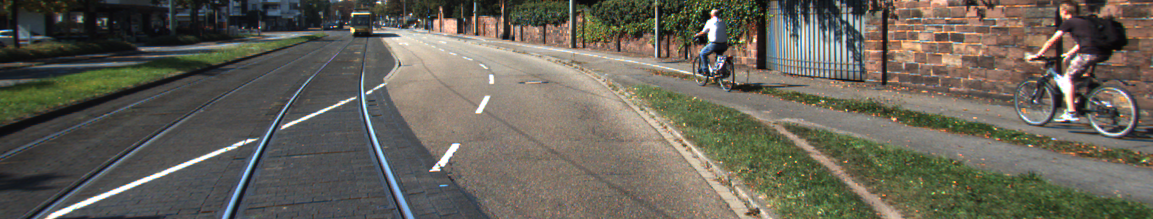}&
\includegraphics[width=0.33\linewidth]{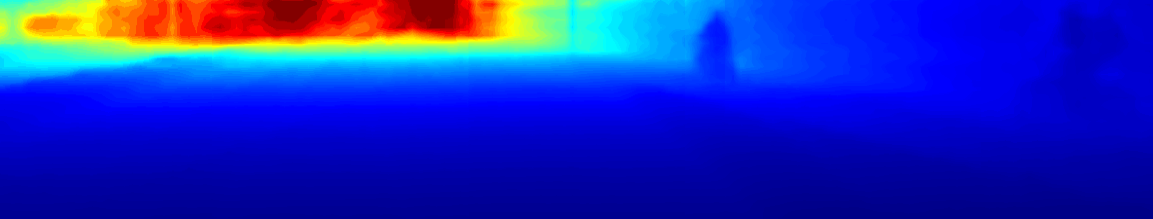}&
\includegraphics[width=0.33\linewidth]{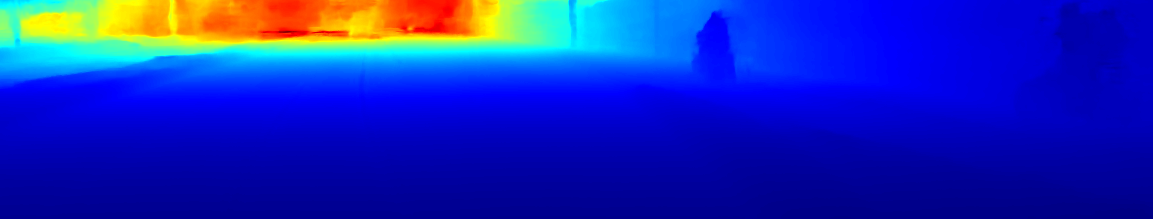}\\
\footnotesize{ Input}&\footnotesize{DORN~\cite{fu2018deep} (Depth)}&\footnotesize{DORN~\cite{fu2018deep} + {\ourmodelshort} (Depth)}\\
\includegraphics[width=0.33\linewidth]{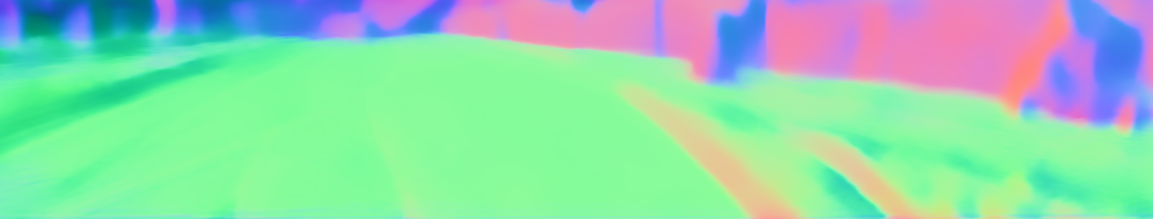}&
\includegraphics[width=0.33\linewidth]{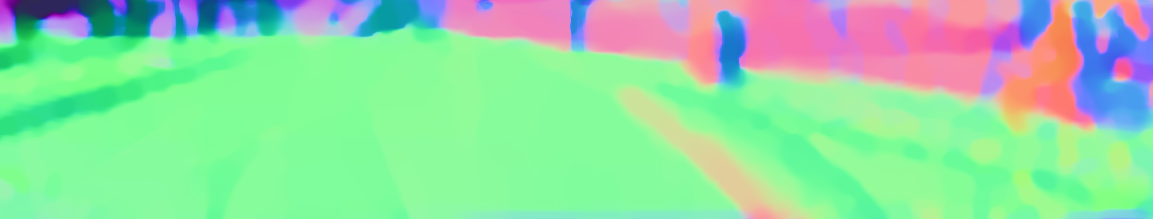}&
\includegraphics[width=0.33\linewidth]{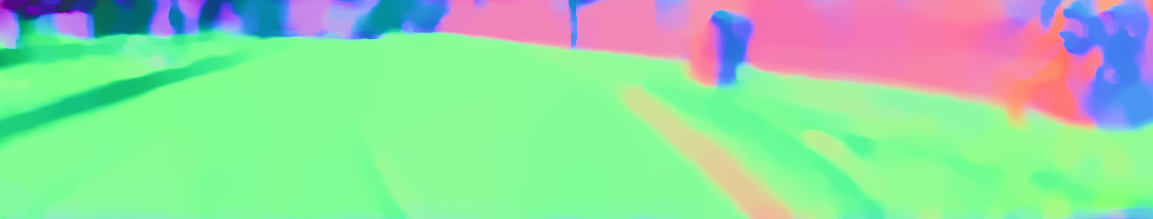}\\
\footnotesize{ Ours (Normal)}&\footnotesize{DORN~\cite{fu2018deep} (Normal)}& \footnotesize{DORN~\cite{fu2018deep} + {\ourmodelshort}  (Normal)}\\
\includegraphics[width=0.33\linewidth]{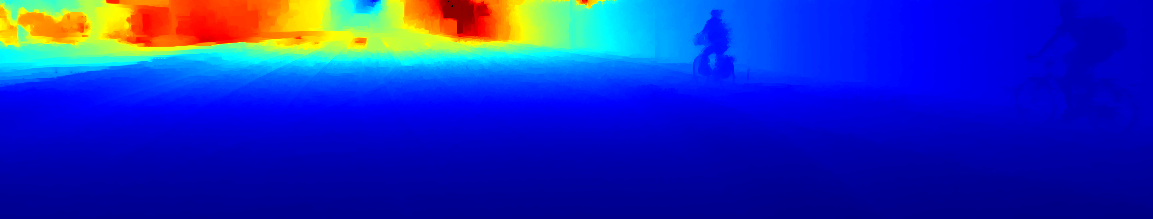}&
\includegraphics[width=0.33\linewidth]{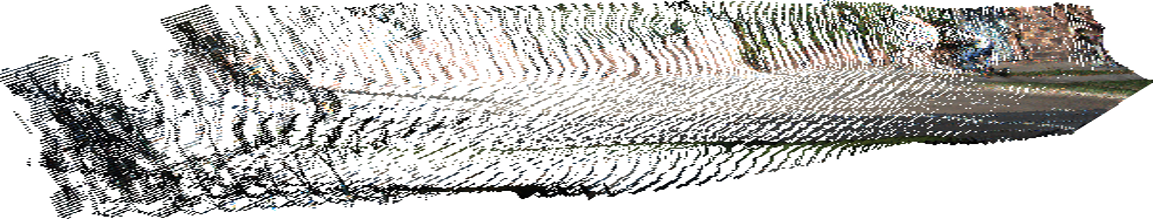}&
\includegraphics[width=0.33\linewidth]{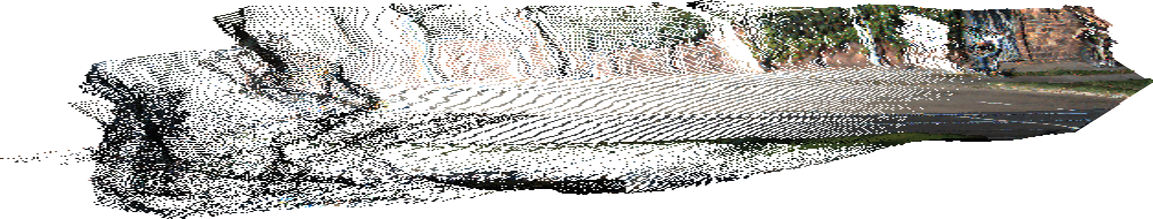}\\
\footnotesize{ GT (Depth)}&\footnotesize{DORN~\cite{fu2018deep} (Point Clouds)}& \footnotesize{DORN~\cite{fu2018deep} + {\ourmodelshort} (Point Clouds)}\\
\end{tabular}
\caption{``DORN~\cite{fu2018deep} + {\ourmodelshort}'' represents that we utilize DORN~\cite{fu2018deep} to produce the initial depth.
The first row shows the depth prediction results.
The second row (column 1) shows normal map directly predicted by our approach for reference.
The second row (columns 2-3) shows the corresponding normal directly estimated from the generated depth.
The third row (column 1) shows the depth ground truth from LIDAR (invalid fields are filled with method~\cite{levin2004colorization} for visualization).
The third row (columns 2-3) shows the corresponding point clouds.}
\label{fig:vis-3d-kitti}
\end{figure*}

We further conduct experiments on {KITTI} dataset to verify the effectiveness of our model  in outdoor scenes. The {KITTI} dataset only provides sparse depth annotations. 
We finetune our {\ourmodelshort} on the training set of KITTI  for $80$-k iterations with batch size $1$. The initial learning rate is $1e{-4}$ and is adjusted following a polynomial decay strategy with a power parameter of $0.9$.
Quantitative comparisons for depth and surface normals are shown in Tables~\ref{tab:com-depth-kitti} and \ref{tab:com-normal_kitti} respectively.
Our method outperforms the state-of-the-art. Visual comparisons are shown in Fig.~\ref{fig:vis-3d-kitti}.
Our {\ourmodelshort} improves boundary prediction results in non-planar regions (see Fig.~\ref{fig:vis-3d-kitti} first row: person and pole regions), and produces smooth results in planar regions (wall and road regions in the 1-st row of Fig.~\ref{fig:vis-3d-kitti}).
The reconstructed normals from {\ourmodelshort} have less noise (2-nd row of Fig.~\ref{fig:vis-3d-kitti}).
Corresponding point cloud visualization (wall regions and persons in the 3-rd row of Fig.~\ref{fig:vis-3d-kitti}) further validates that {\ourmodelshort} improves 3D reconstruction.

\begin{table}
\centering
\caption{Performance of surface normal prediction on the KITTI dataset.}
\label{tab:com-normal_kitti}
\setlength{\tabcolsep}{2pt}
\scalebox{0.95}{
\begin{tabular}{c|ccc|ccc}
\toprule
\multirow{1}{*}{} &
\multicolumn{3}{c|}{{Error}}  &
\multicolumn{3}{c}{{Accuracy}} \\
& {{mean}} & {{median}} & {{rmse}} & {{$11.25^\circ$}} & {{$22.5^\circ$}} & {{$30^\circ$}} \\
\midrule
\midrule
Baseline & 15.21& 8.17& 23.76 & 60.08 & 78.95 & 85.23\\
Ours& \textbf{14.87} & \textbf{7.79} & \textbf{23.46} & \textbf{61.24} & \textbf{79.52} & \textbf{85.60}\\
\bottomrule
\end{tabular}}

\end{table}

\subsection{Running Time Analysis}
We test our {\ourmodelshort} on a PC with an Intel i7-6950 CPU and a single TitanX GPU.
When using VGG-16 as the backbone network, it takes  $0.87$ seconds to produce the final depth and normal estimates for an image with size $480 \times 640$. In comparison, \textit{Local Network}~\cite{chakrabarti2016depth} takes around $24$s to predict the depth map of the same-sized image; Multi-scale CRF~\cite{xu2018monocular} takes around $2.25$s to process an image; SURGE~\cite{wang2016surge}\footnote{We do not have the exact running time as there is no released code.} also takes longer time since \textcolor{black} {it has to go through four VGG-16 networks and requires multiple  mean-field inference steps.}

\section{Benchmark Depth Prediction in 3D}
Previous metrics for depth prediction from a single image only focus on 2D pixel-wise metrics without considering its usefulness in reconstructing real 3D surface, which is very crucial in real-world applications.
The reconstruction of the 3D surface heavily depends on the corresponding 3D point cloud -- 2D pixel-wise metrics cannot directly measure the reconstruction quality in 3D. 
As shown in Fig.~\ref{fig:vis-3d}, depth from DORN~\cite{fu2018deep} ($0.844$) has slightly lower RMSE error than ours ($0.860$), while the point cloud generated from our prediction is obviously more structured than the one of DORN (Fig.~\ref{fig:vis-3d} second row). 
Our normal estimation is also better (Fig.~\ref{fig:vis-3d} third row).
The above examples indicate the necessity of a metric regarding the 3D reconstruction quality.

Here we propose a complementary 3D geometric metric (3DGM) to evaluate ``how much the predicted depth helps high-quality 3D surface reconstruction''. 
We also evaluate state-of-the-art approaches using this 3D geometric metric.
We utilize our {\ourmodelshort} to further refine these results to verify that {\ourmodelshort} can generally improve the 3D reconstruction quality. 
In the following, we elaborate on our proposed 3D geometric metric and show more 3D visual examples.

\vspace{0.1in}
\textbf{3D geometric metric.~~} We now introduce the 3D geometric metric (3DGM) for evaluating depth results by measuring its usefulness in reconstructing 3D surfaces.
Specifically, to compute the metric, we first cast the predicted depth map into its 3D position. 
Then, the corresponding surface normals are derived with the provided development kit and further denoised with TV-denoising following the method of~\cite{ldiscriminatively}. 
We finally compare it with the surface normal estimated from the ground truth depth produced by Kinect or LIDAR.
We evaluate methods~\cite{eigen2015predicting,laina2016deeper,fu2018deep,xu2017multi} regarding the new 3D metrics on both NYUD-V2 and KITTI datasets.
Quantitative results on the NYUD-V2 dataset are shown in Tab.~\ref{tab:com-normal_all}.
{\ourmodelshort} consistently improves the 3D geometric metric by a large margin regardless of the choice of backbone architectures.
Similar results on the KITTI dataset are obtained as shown in Tab.~\ref{tab:com-depth-to-norm-kitti}.

\begin{table}
\centering
\caption{Quantitative comparisons on the NYUD-V2 dataset in terms of 3D geometric metric (3DGM).}
\label{tab:com-normal_all}
\setlength{\tabcolsep}{2pt}
\scalebox{0.95}{
\begin{tabular}{c|ccc|ccc}
\toprule
\multirow{1}{*}{} &
\multicolumn{3}{c|}{{Error}}  &
\multicolumn{3}{c}{{Accuracy}} \\
& {{mean}} & {{median}} & {{rmse}} & {{$11.25^\circ$}} & {{$22.5^\circ$}} & {{$30^\circ$}} \\
\midrule
\midrule
Our Baseline  & 42.39 & 37.61 & 50.81 & 12.09 & 28.97 & 39.68 \\
Our Baseline + {\ourmodelshort}  & \textbf{35.78} & \textbf{30.10} &\textbf{43.86} & \textbf{16.07} & \textbf{37.28} & \textbf{49.84} \\\midrule
Multi-scale CNN~\cite{eigen2015predicting} & 34.52 & 26.20 & 44.32 & 20.35& 43.84 & 55.58 \\
Multi-scale CNN~\cite{eigen2015predicting} + {\ourmodelshort} & \textbf{29.61} & \textbf{21.54} & \textbf{39.14} & \textbf{26.61}& \textbf{51.71} & \textbf{63.11} \\ \midrule
Multi-scale CRF~\cite{xu2017multi}& 35.47 & 28.63 & 44.47 & 19.17 & 40.47 & 51.93 \\
Multi-scale CRF~\cite{xu2017multi} + {\ourmodelshort}& \textbf{31.79} & \textbf{25.30} & \textbf{40.12} & \textbf{21.65} & \textbf{45.14} & \textbf{57.28} \\\midrule
FCRN~\cite{laina2016deeper} &30.32 & 21.74 & 40.28 & 28.46 & 51.26 & 61.61 \\
FCRN~\cite{laina2016deeper} + {\ourmodelshort} &\textbf{28.27} & \textbf{19.88} & \textbf{38.19} & \textbf{30.06} & \textbf{54.67} & \textbf{65.46} \\\midrule
DORN~\cite{fu2018deep} & 32.94 & 25.49 & 42.94 & 25.61 & 45.55 & 56.10\\
DORN~\cite{fu2018deep} + {\ourmodelshort} & \textbf{29.39} & \textbf{20.74} & \textbf{39.88} & \textbf{30.08} & \textbf{52.95} & \textbf{63.45}\\
\bottomrule
\end{tabular}}
\end{table}

\begin{figure*}
\centering
\begin{tabular}{@{\hspace{0.1mm}}c@{\hspace{0.1mm}}c@{\hspace{0.1 mm}}c@{\hspace{0.1 mm}}c@{\hspace{0.1 mm}}c@{\hspace{0.1 mm}}c}
\includegraphics[width=0.163\linewidth]{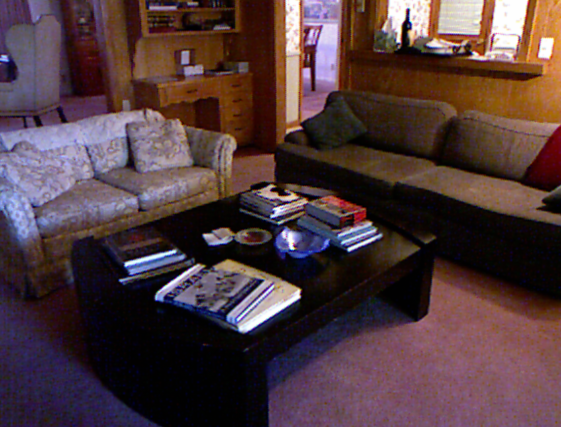} &
\includegraphics[width=0.163\linewidth]{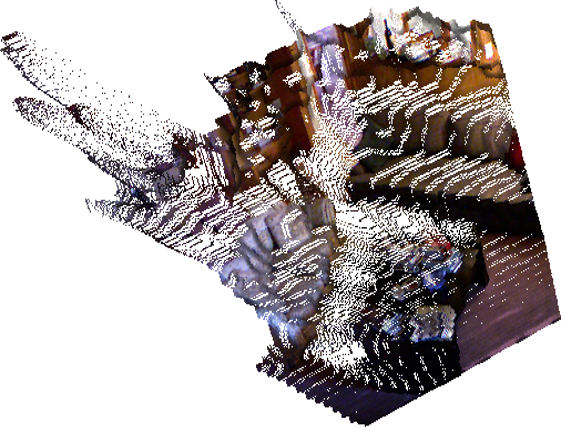}&
\includegraphics[width=0.163\linewidth]{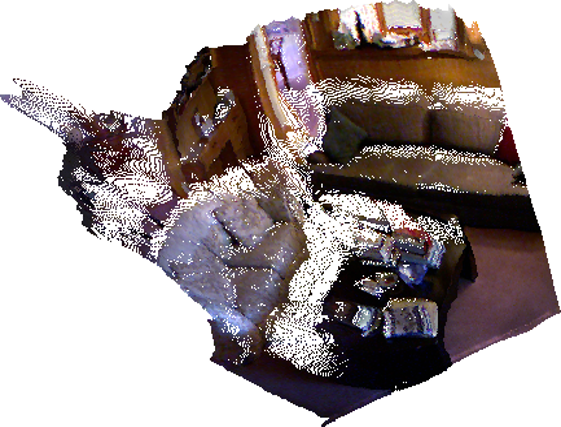}&
\includegraphics[width=0.163\linewidth]{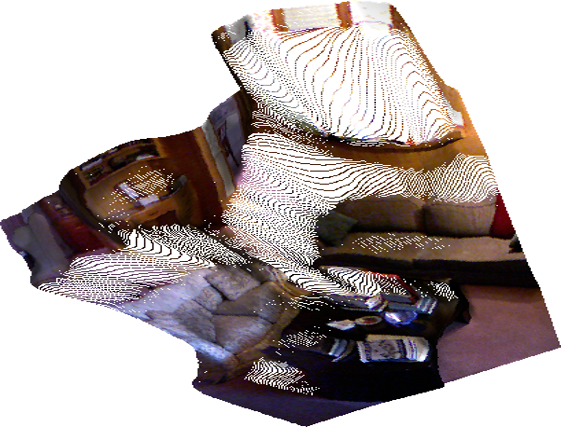}&
\includegraphics[width=0.163\linewidth]{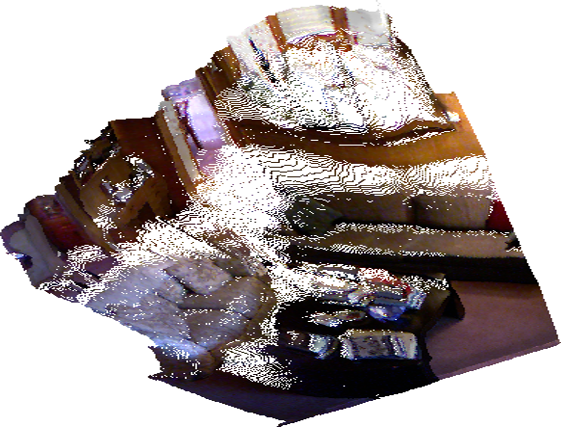}&
\includegraphics[width=0.163\linewidth]{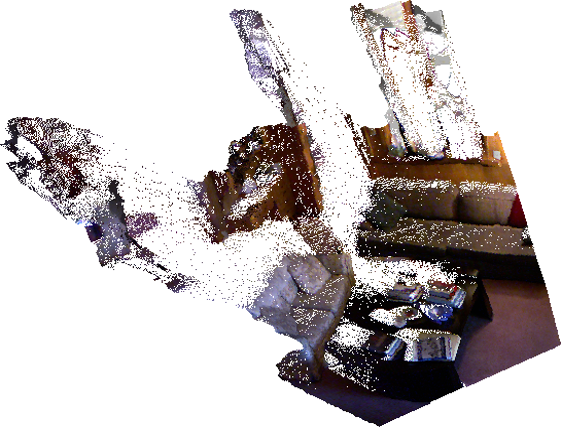}\\
\includegraphics[width=0.163\linewidth]{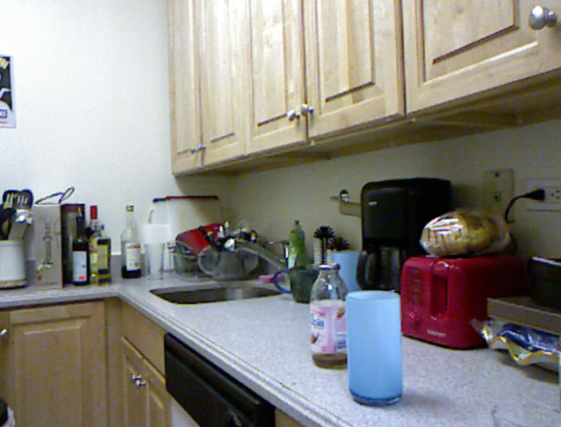} &
\includegraphics[width=0.163\linewidth]{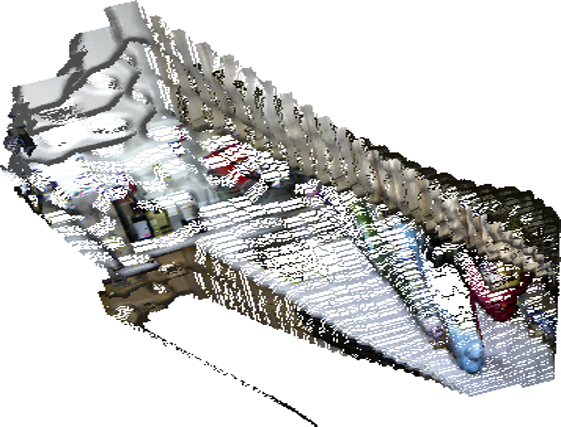}&
\includegraphics[width=0.163\linewidth]{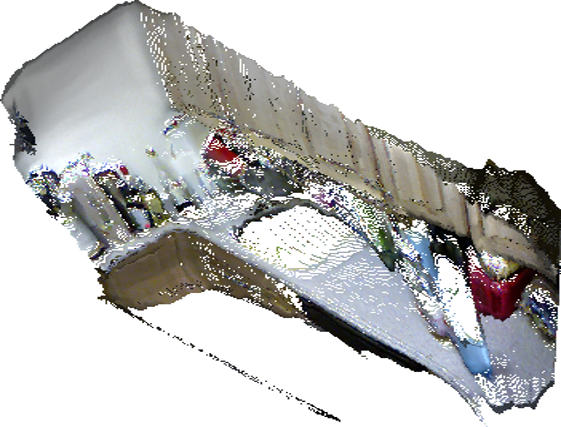}&
\includegraphics[width=0.163\linewidth]{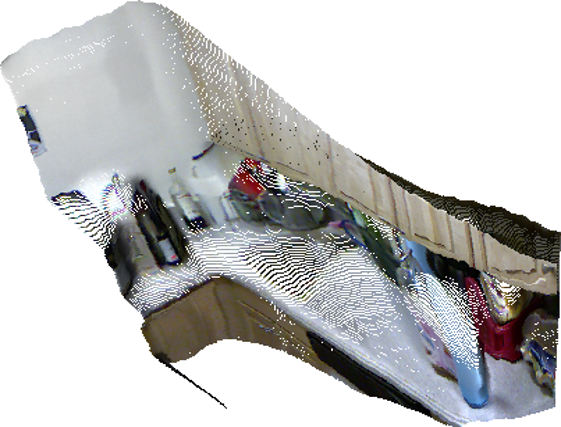}&
\includegraphics[width=0.163\linewidth]{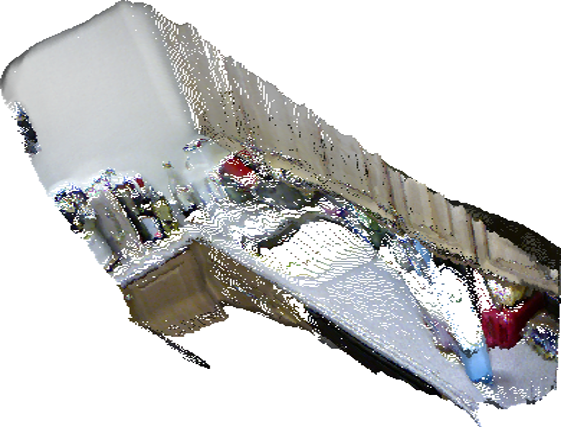}&
\includegraphics[width=0.163\linewidth]{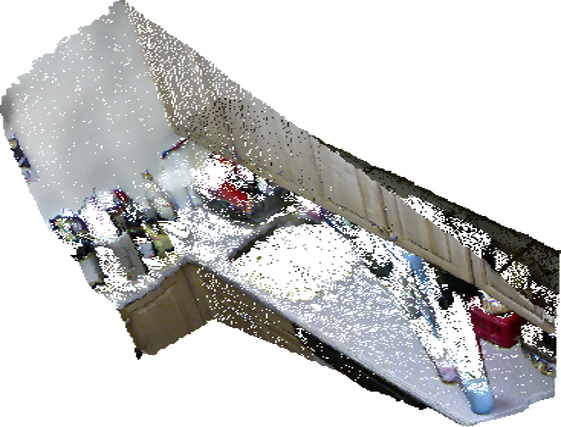}\\
\includegraphics[width=0.163\linewidth]{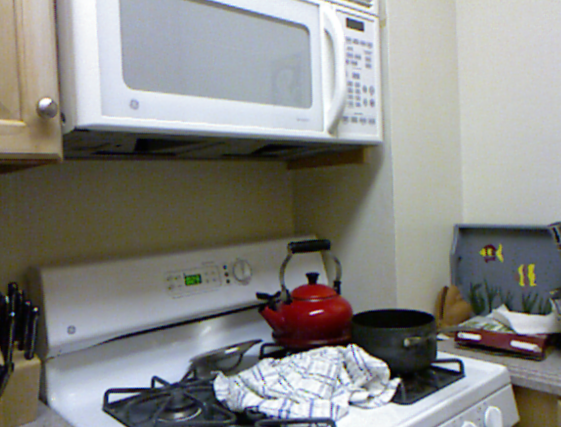} &
\includegraphics[width=0.163\linewidth]{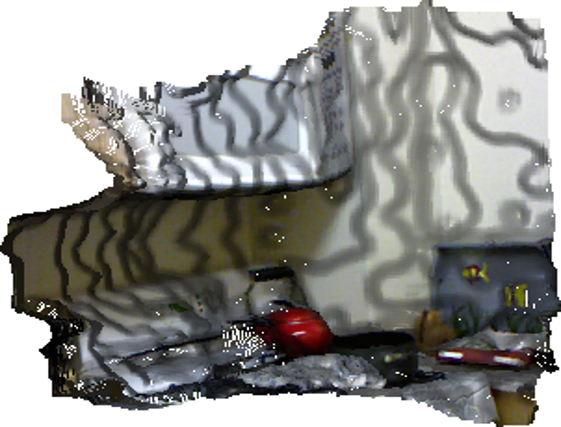}&
\includegraphics[width=0.163\linewidth]{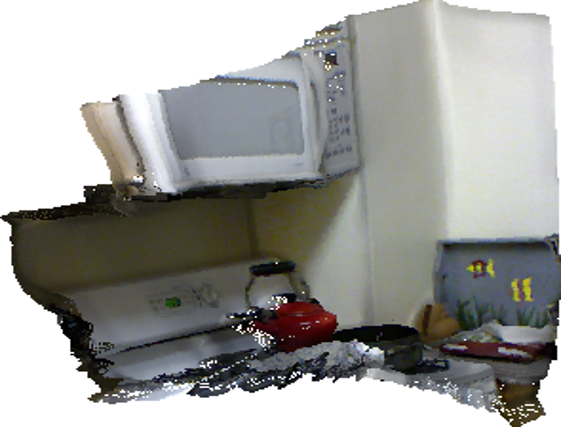}&
\includegraphics[width=0.163\linewidth]{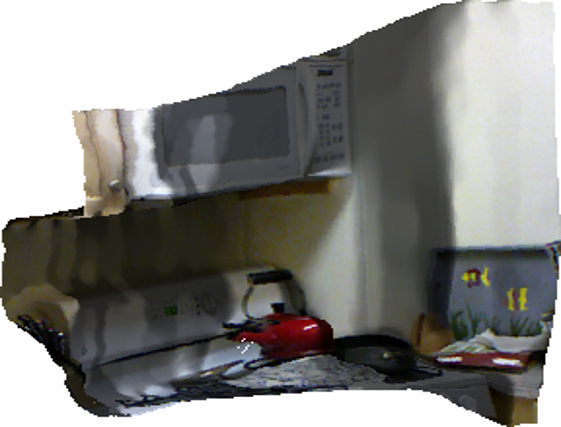}&
\includegraphics[width=0.163\linewidth]{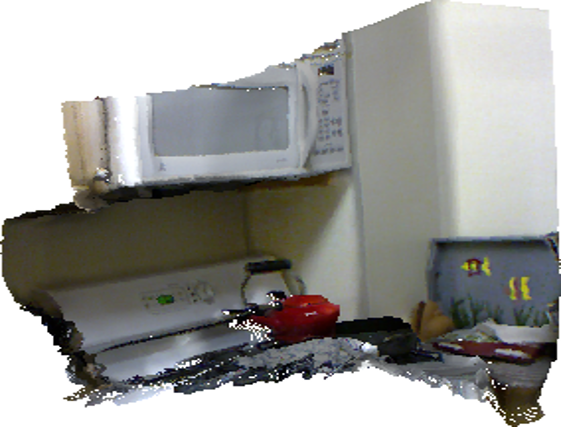}&
\includegraphics[width=0.163\linewidth]{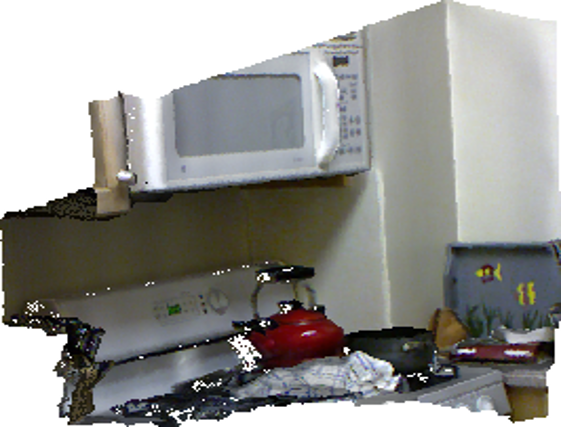}\\
\footnotesize{ (a) Input } &\footnotesize{(b) DORN~\cite{fu2018deep}}&\footnotesize{(c) DORN~\cite{fu2018deep} + {\ourmodelshort}}&\footnotesize{ (d) FCRN~\cite{laina2016deeper}}&\footnotesize{ (e) FCRN~\cite{laina2016deeper} + {\ourmodelshort}}&\footnotesize{(f) GT} \\
\end{tabular}
\caption{\textcolor{black}{Visual comparison of 3D point clouds on the NYUD-V2 dataset. ``FCRN~\cite{laina2016deeper} + {\ourmodelshort}'' indicates that we utilize FCRN~\cite{laina2016deeper} as the depth backbone in our system.
``DORN~\cite{fu2018deep} + {\ourmodelshort}'' indicates that DORN~\cite{fu2018deep} serves as the depth backbone for our system.}}
\label{fig:vis-3dpred}
\end{figure*}

\begin{table}
\centering
\caption{Quantitative comparisons on the KITTI dataset in terms of 3DGM.}
\label{tab:com-depth-to-norm-kitti}
\setlength{\tabcolsep}{2pt}
\scalebox{0.95}{
\begin{tabular}{c|ccc|ccc}
\toprule
\multirow{1}{*}{} &
\multicolumn{3}{c|}{{Error}}  &
\multicolumn{3}{c}{{Accuracy}} \\
& {{mean}} & {{median}} & {{rmse}} & {{$11.25^\circ$}} & {{$22.5^\circ$}} & {{$30^\circ$}} \\
\midrule
\midrule

Multi-scale CNN~\cite{eigen2015predicting} & 38.00 &29.44& 47.00& 11.42& 39.53 & 50.74 \\
Multi-scale CNN~\cite{eigen2015predicting} + {\ourmodelshort} & \textbf{36.63} & \textbf{28.25} & \textbf{45.29} & \textbf{11.18}& \textbf{40.97} &\textbf{52.39} \\ \midrule
DORN~\cite{fu2018deep} & 22.49& 13.00& 33.25 & 45.53 & 66.65 & 74.71\\
DORN~\cite{fu2018deep} + {\ourmodelshort} & \textbf{21.89} & \textbf{12.31} & \textbf{32.59} & \textbf{46.84} & \textbf{68.99} & \textbf{76.73}\\
\bottomrule
\end{tabular}}

\end{table}
\vspace{0.1in}
\textbf{Qualitative visualization in 3D.~~}\label{sec:3D-result}
The 3D qualitative comparisons of predicted depth maps on the NYUD-V2 dataset are shown in Fig.~\ref{fig:vis-3dpred}. {\ourmodelshort} consistently improves the 3D point cloud reconstruction quality. Planar regions are well preserved and details of small objects are clearer. DORN~\cite{fu2018deep} is a top-performing approach for depth prediction in terms of 2D metrics. However, under the 3D metric, the geometric constraint is not well satisfied, resulting in problematic 3D visualization as well. After refined by {\ourmodelshort}, the 3D quality has been significantly improved. This further validates that 2D metrics are insufficient to fully measure the depth quality, and {\ourmodelshort} smooths prediction in planar regions considering geometric constraints, and refines boundaries with the weighted propagation.

\begin{figure*}
\centering
\begin{tabular}{@{\hspace{0.1mm}}c@{\hspace{0.1mm}}c@{\hspace{0.1 mm}}c@{\hspace{0.1 mm}}c@{\hspace{0.1 mm}}c@{\hspace{0.1 mm}}c@{\hspace{0.1 mm}}c}
\includegraphics[width=0.140\linewidth]{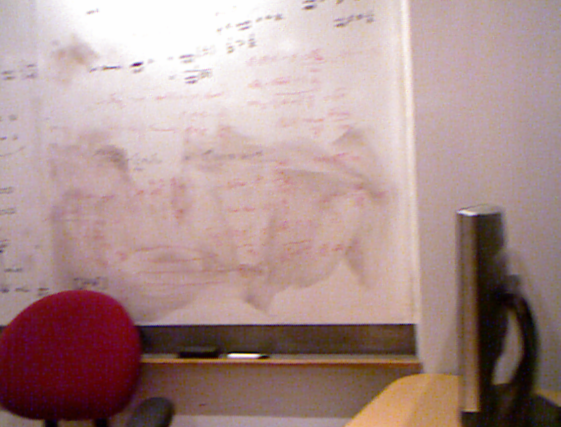} &
\includegraphics[width=0.140\linewidth]{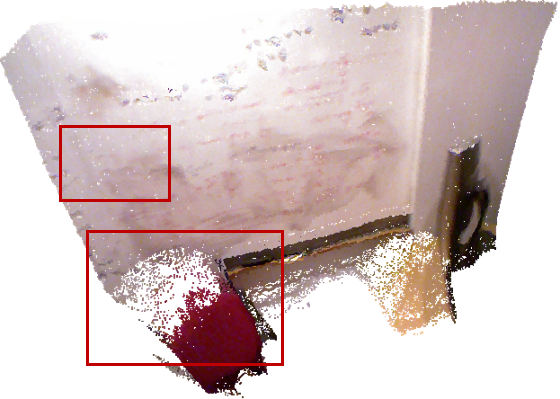}&
\includegraphics[width=0.140\linewidth]{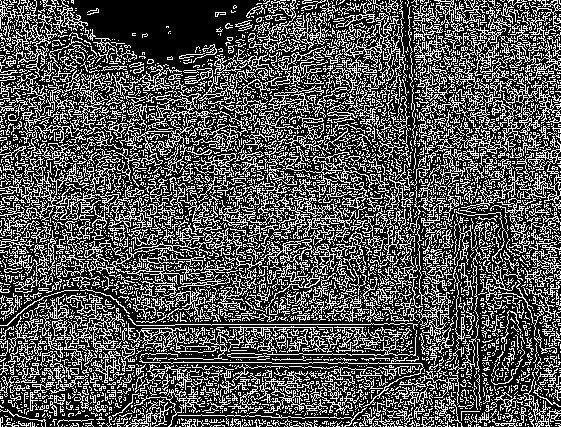}&
\includegraphics[width=0.140\linewidth]{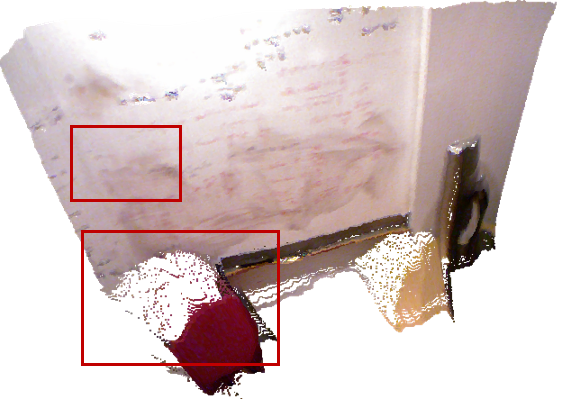}&
\includegraphics[width=0.140\linewidth]{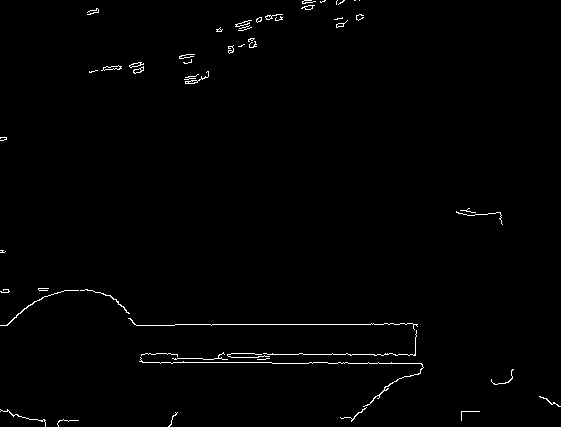}&
\includegraphics[width=0.140\linewidth]{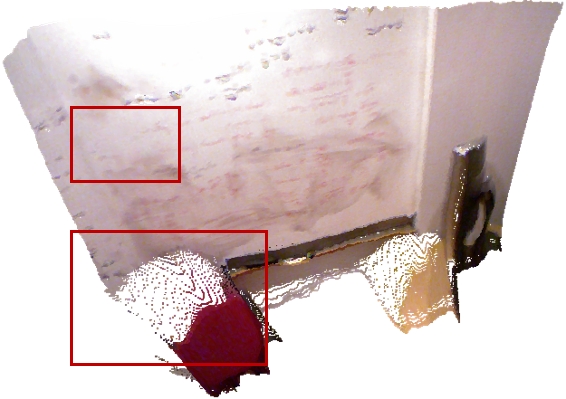}&
\includegraphics[width=0.140\linewidth]{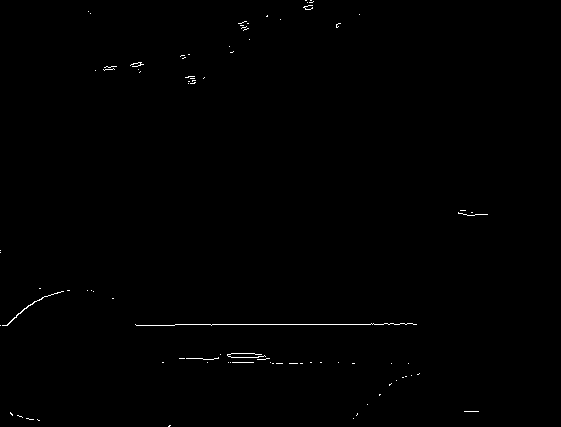}\\
\footnotesize{ (a) Image } &\footnotesize{(b) Point Cloud (Low)}&\footnotesize{ (c) Edge (Low)}&\footnotesize{ (d) Point Cloud (Mid)}&\footnotesize{ (e) Edge (Mid)}&\footnotesize{ (f) Point Cloud (High)}&\footnotesize{(g) Edge (High)} \\
\end{tabular}
\caption{\textcolor{black}{Comparisons of different canny parameters for surface reconstruction. Zoom in to see clearer.}}
\label{fig:vis-canny}
\end{figure*}
\begin{table}
\centering
\caption{Ablation studies on surface normal estimation on the NYUD-V2 dataset. {\ourmodelshortoriginal}: depth-to-normal and normal-to-depth modules. Full Model: all the modules with iterative inference.}
\label{tab:ablation-norm}
\setlength{\tabcolsep}{2pt}
\scalebox{0.95}{
\begin{tabular}{c|ccc|ccc}
\toprule
\multirow{1}{*}{} &
\multicolumn{3}{c|}{{Error}}  &
\multicolumn{3}{c}{{Accuracy}} \\
& {{mean}} & {{median}} & {{rmse}} & {{$11.25^\circ$}} & {{$22.5^\circ$}} & {{$30^\circ$}} \\
\midrule
\midrule
Baseline & {19.4}&{12.5}&{27.0}&{46.0}&{70.3}&{78.9}\\
With {\ourmodelshortoriginal}~\cite{qi2018geonet}& {{19.0}} & {{11.8}} & {{26.9}} & {{48.4}} & {{71.5}} & {{79.5}}\\
With Ensemble& 18.9 & 11.8 & 26.9 & 48.3& 71.9 & 79.8\\
With Edge-aware& 18.6 &11.3& 26.7 & 50.0 & 72.7 & 80.4\\
Full Model&\textbf{18.5}&\textbf{11.2}&\textbf{25.7}&\textbf{50.2}&\textbf{73.2}&\textbf{80.7}\\
\midrule \midrule
\textcolor{black}{Canny (Low)} & \textcolor{black}{18.8} &\textcolor{black}{11.4}& \textcolor{black}{26.9} & \textcolor{black}{49.4} & \textcolor{black}{72.2} &\textcolor{black} {80.0}\\
\textcolor{black}{Canny (Mid)} & \textcolor{black}{18.6} &\textcolor{black}{11.3}& \textcolor{black}{26.7} & \textcolor{black}{50.0} & \textcolor{black}{72.7} & \textcolor{black}{80.4}\\
\textcolor{black}{Canny (High)} & \textcolor{black}{18.7} &\textcolor{black}{11.4}& \textcolor{black}{26.8} & \textcolor{black}{50.0} & \textcolor{black}{72.7} & \textcolor{black}{80.4}\\ \bottomrule
\end{tabular}}
\end{table}

\section{Ablation Studies}
\begin{figure*}
\centering
\begin{tabular}{@{\hspace{0.1mm}}c@{\hspace{0.1mm}}c@{\hspace{0.1 mm}}c@{\hspace{0.1 mm}}c@{\hspace{0.1 mm}}c@{\hspace{0.1 mm}}c}
\includegraphics[width=0.163\linewidth]{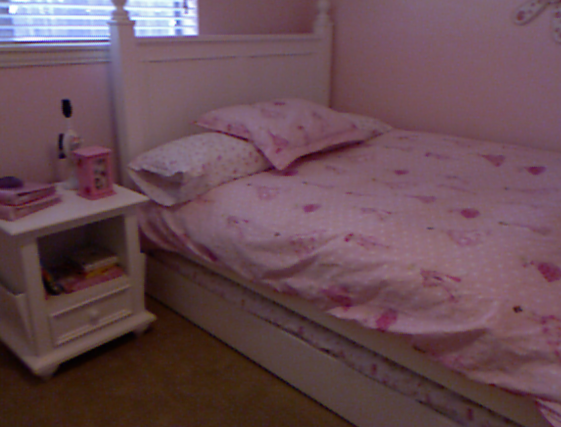} &
\includegraphics[width=0.163\linewidth]{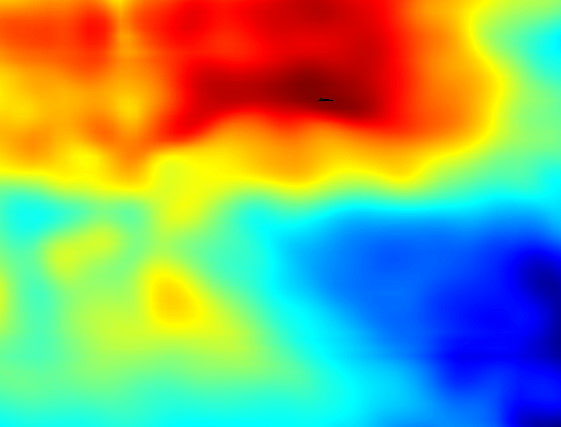}&
\includegraphics[width=0.163\linewidth]{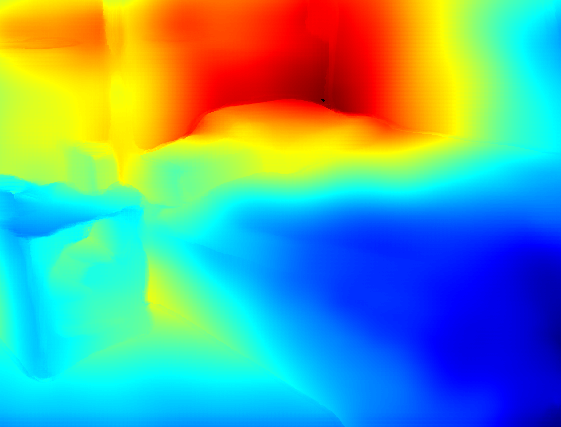}&
\includegraphics[width=0.163\linewidth]{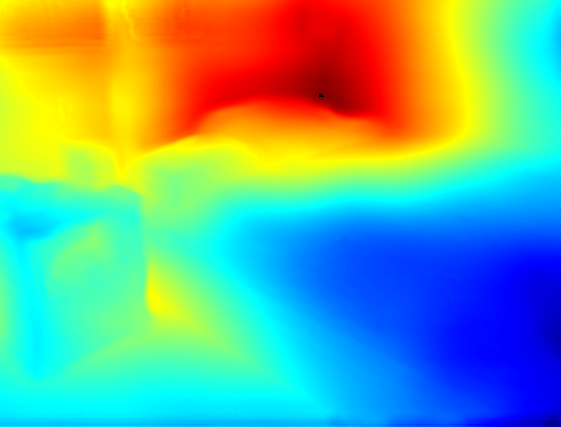}&
\includegraphics[width=0.163\linewidth]{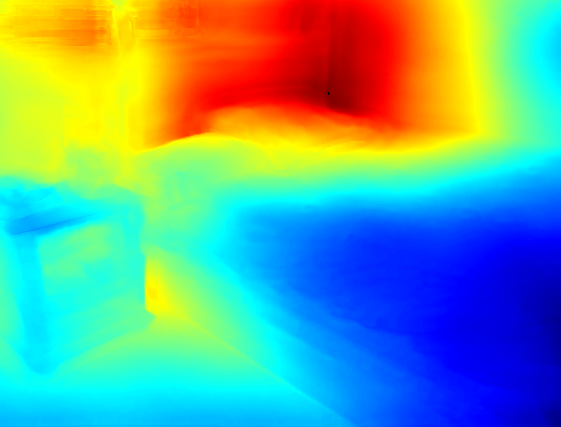}&
\includegraphics[width=0.163\linewidth]{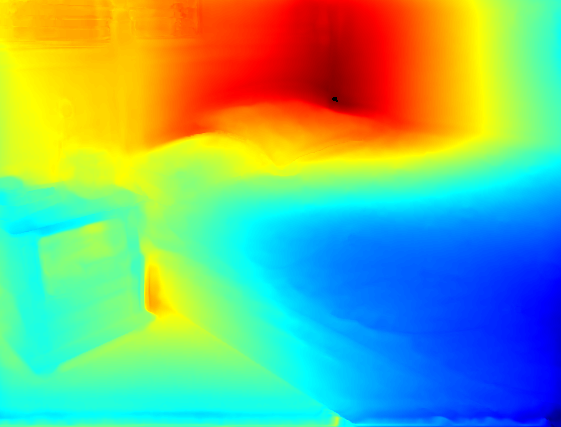}\\
\includegraphics[width=0.163\linewidth]{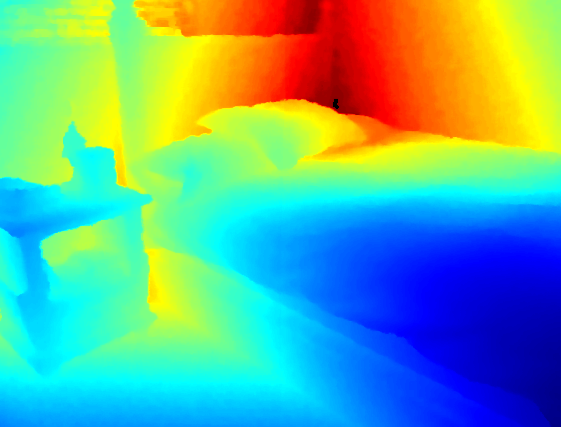} &
\includegraphics[width=0.163\linewidth]{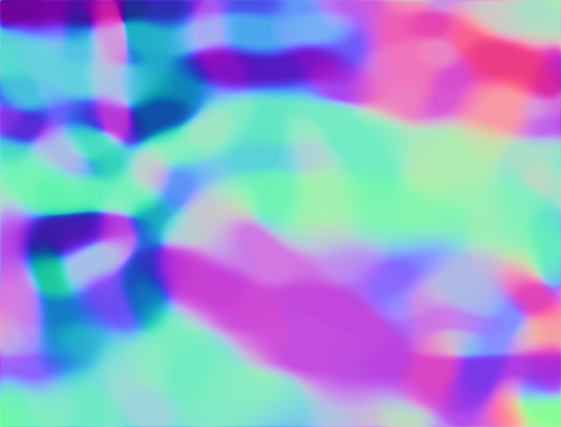}&
\includegraphics[width=0.163\linewidth]{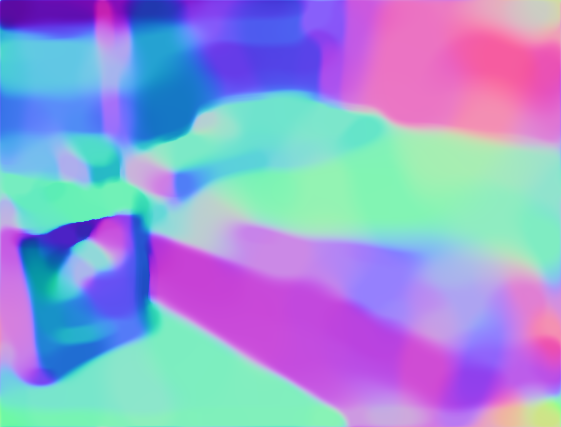}&
\includegraphics[width=0.163\linewidth]{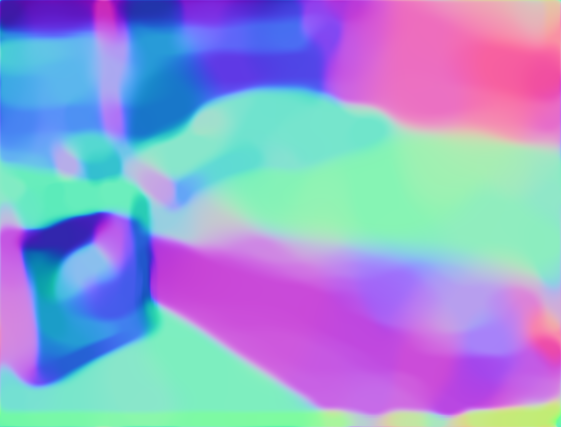}&
\includegraphics[width=0.163\linewidth]{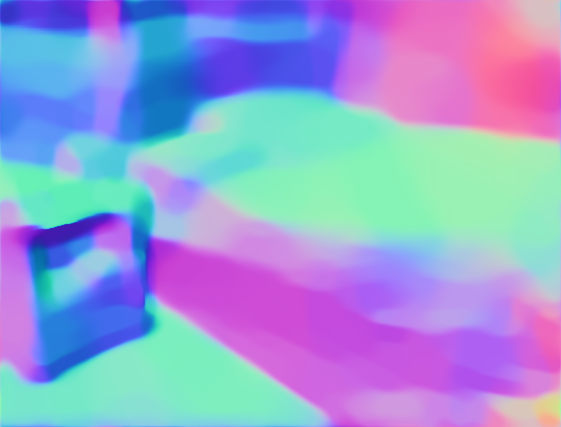}&
\includegraphics[width=0.163\linewidth]{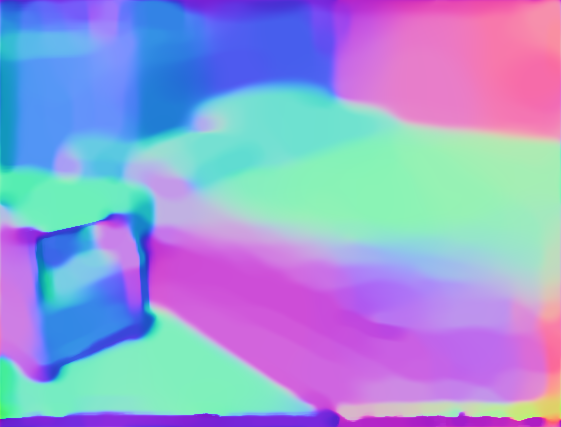}\\
\includegraphics[width=0.163\linewidth]{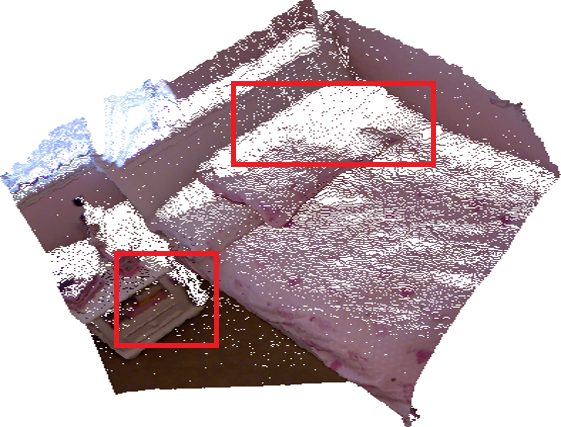} &
\includegraphics[width=0.163\linewidth]{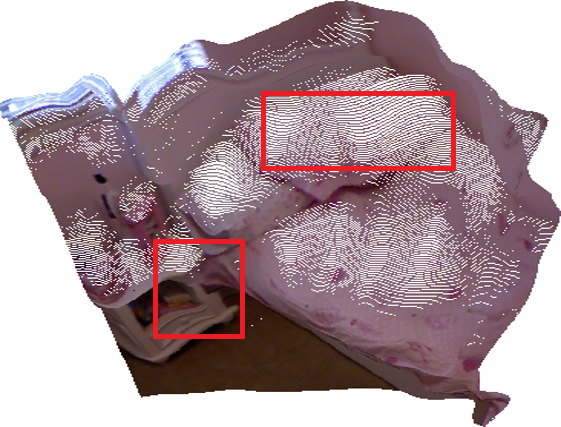}&
\includegraphics[width=0.163\linewidth]{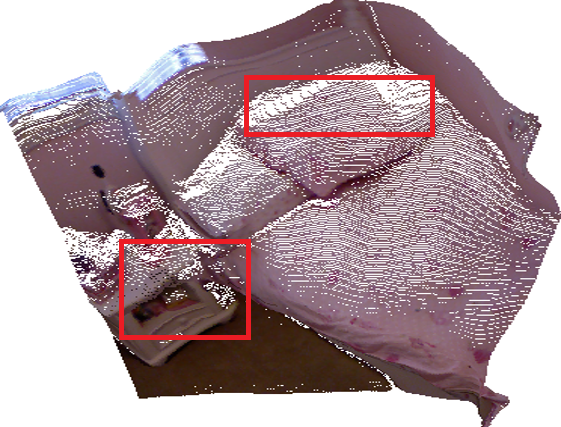}&
\includegraphics[width=0.163\linewidth]{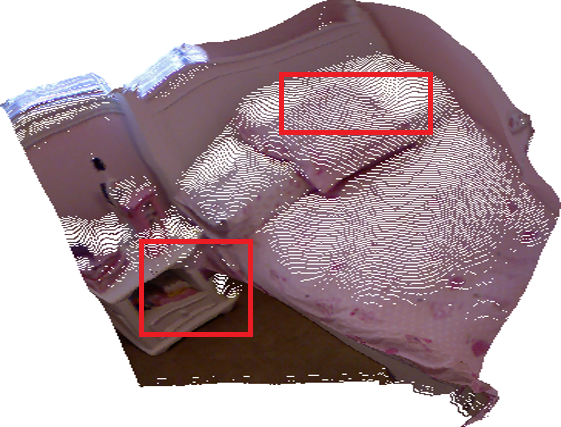}&
\includegraphics[width=0.163\linewidth]{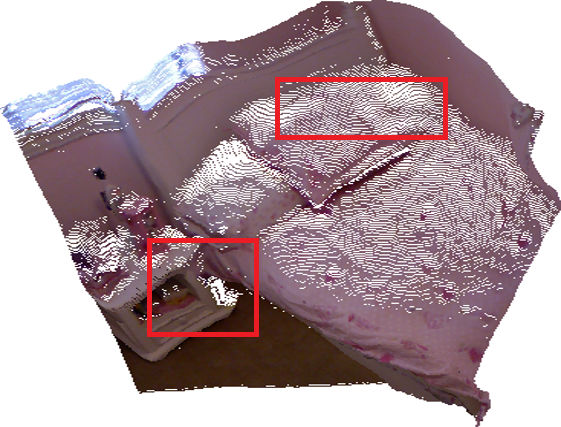}&
\includegraphics[width=0.163\linewidth]{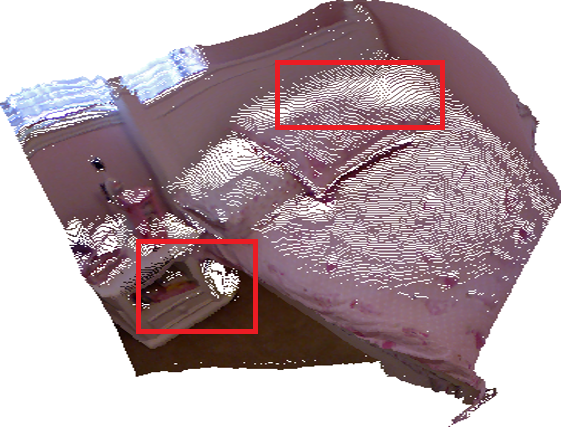}\\
\includegraphics[width=0.163\linewidth]{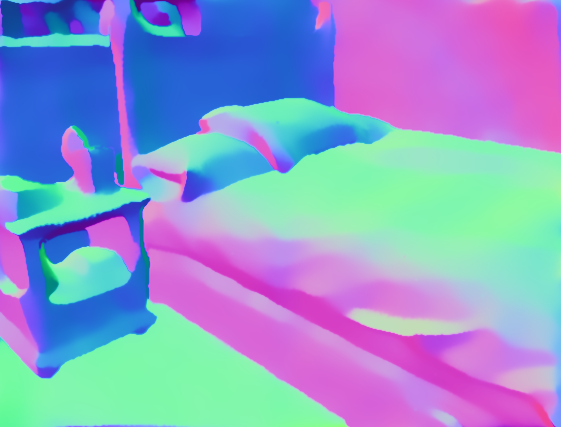} &
\includegraphics[width=0.163\linewidth]{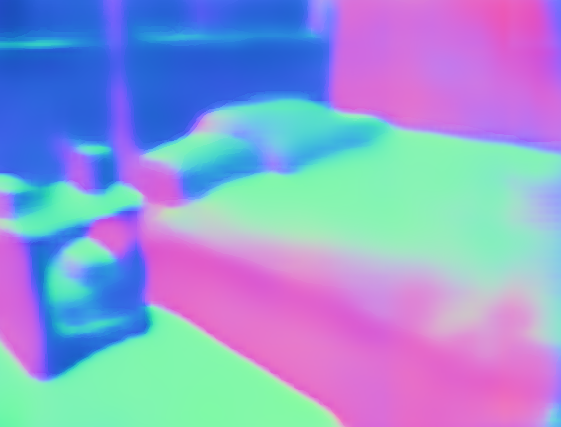}&
\includegraphics[width=0.163\linewidth]{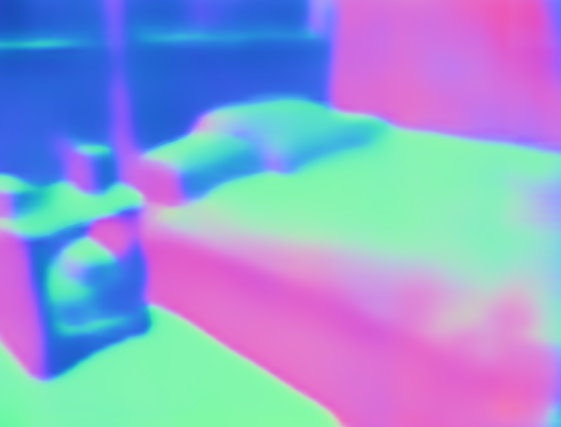}&
\includegraphics[width=0.163\linewidth]{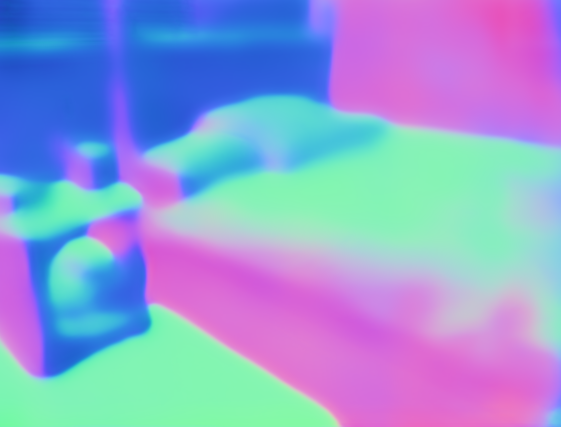}&
\includegraphics[width=0.163\linewidth]{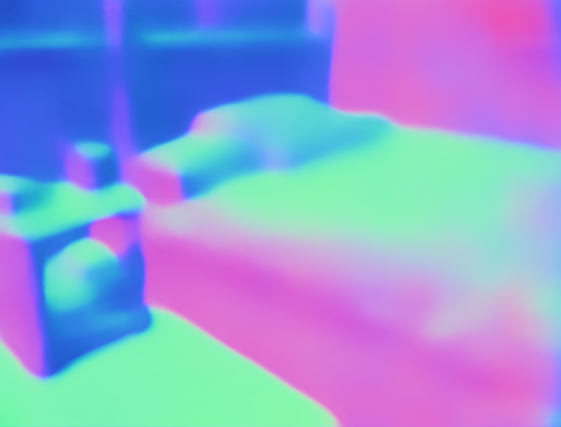}&
\includegraphics[width=0.163\linewidth]{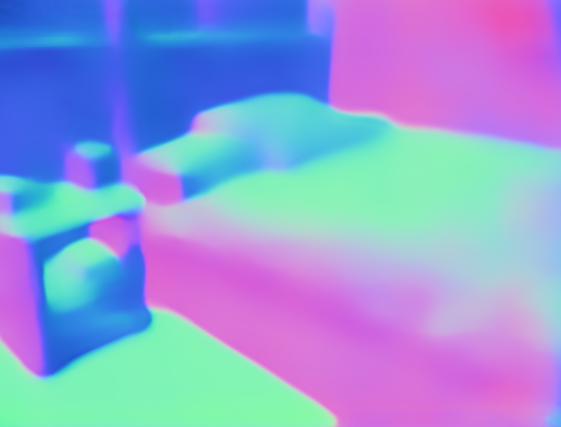}\\
\footnotesize{ (a) Input \& GT } &\footnotesize{(b) Baseline}&\footnotesize{(c) With {\ourmodelshortoriginal}}&\footnotesize{ (d) With Ensemble}&\footnotesize{ (e) With Edge-aware}&\footnotesize{(f) Full Model} \\
\end{tabular}
\caption{Qualitative comparisons for ablation studies.
{\ourmodelshortoriginal}: depth-to-normal and normal-to-depth modules. Full Model: all modules with iterative inference.
First row: depth prediction results;
Second row: surface normal generated from predicted depth in the first row;
Third row: point cloud visualization;
Fourth row: surface normal directly predicted from the image.}
\label{fig:vis-ablation}
\end{figure*}

We evaluate the effectiveness of each component of {\ourmodelshort} both quantitatively and qualitatively on the NYUD-V2 dataset. Tab.~\ref{tab:ablation-depth} shows the effectiveness of different components, including depth-to-normal, normal-to-depth, ensemble module, and edge-aware refinement module in terms of 2D pixel-wise metrics.
Tab.~\ref{tab:ablation-norm} shows the influence of different components for surface normal prediction.
The ensemble module improves the performance via fusing predictions from the geometric module and the backbone network.
The edge-aware refinement module improves the output by reducing the noise and making the boundary predictions more accurate.
Quantitative results in terms of 3DGM are shown in Tab.~\ref{tab:com-depth-to-norm}. 
Visual comparisons are given in Fig.~\ref{fig:vis-ablation}. 
As can be seen, {\ourmodelshortoriginal}, including depth-to-normal and normal-to-depth modules, smooths the prediction in planar regions (the wall region in Fig.~\ref{fig:vis-ablation} (c)) and meanwhile preserves the details of small objects (the pillow and counter in Fig.~\ref{fig:vis-ablation} (c)). 
The ensemble module further enhances the result by combing the initial and the geometric predictions, making it closer to ground truth (Fig.~\ref{fig:vis-ablation} (d)). 
The edge-aware module refines the boundary (bed and counter in Fig.~\ref{fig:vis-ablation} (e)). \textcolor{black} {For the Canny edge detection,  we use the function in OpenCV  to compute the edge maps. The ``minValue'' threshold is set to be the mean of pixel intensities of the dataset (around $100$) and the ``maxValue''  is two times the mean value (around $200$). Here, we validate the robustness of our pipeline to the  parameters of the Canny edge detector by changing the thresholds to be extremely low values (i.e., ``minValue'': 0 and ``maxValue'': 0  ) and extremely high values (i.e., ``minValue'': 255 and ``maxValue'': 255). ``Mid'' corresponds to the values used in our experiments (i.e., ``minValue'': 100 and ``maxValue'': 200).
Experimental results in Tab.~\ref{tab:ablation-depth} and Tab. ~\ref{tab:ablation-norm} show that the performance is robust to parameters of the Canny edge detector benefited from the learn-able residual module. We also show visual comparisons in Fig.~\ref{fig:vis-canny}. Extremely low edge thresholds will lead to noisy reconstructions (see Fig~\ref{fig:vis-canny} white points). With high thresholds, the result will have more flying pixels in the boundary regions compared to our settings. Our experimental observation indicates that the visual quality is generally stable with a large range of parameters around the ``Mid'' .}

\begin{table}
\centering
\caption{Ablation studies on depth prediction on the NYUD-V2 dataset. {\ourmodelshortoriginal}: depth-to-normal and normal-to-depth modules. Full Model: all the modules with iterative inference.}
\label{tab:ablation-depth}
\setlength{\tabcolsep}{2pt}
\scalebox{0.95}{
\begin{tabular}{c|ccc|ccc}
\toprule
\multirow{1}{*}{Method}&
\multicolumn{3}{c|}{{Error}}&
\multicolumn{3}{c}{{Accuracy}} \\
& {{rmse}} & {{$\log10$}} & {{rel}} & {{$1.25$}} & {{$1.25^2$}} & {{$ 1.25^3$}} \\
\midrule
\midrule
Our Baseline  &{0.626}&{0.068}&{0.155}&{0.768}&{0.951}&{0.988}\\
With {\ourmodelshortoriginal}\cite{qi2018geonet}&{{0.608}} & {{0.065}}&{{0.149}} & {0.786} &{0.956} &{{0.990}}\\
With Ensemble &0.605&0.064&0.147&0.789&0.957&0.990\\
With Edge-aware &0.605&0.064&0.146&0.789&0.958&0.990\\
Full Model&\textbf{0.600}&\textbf{0.063}&\textbf{0.144}&\textbf{0.791}&\textbf{0.960}&\textbf{0.991}\\
\midrule \midrule
\textcolor{black}{Canny (Low)} &\textcolor{black}{0.606}&\textcolor{black}{0.064}&\textcolor{black}{0.147}&\textcolor{black}{0.789}&\textcolor{black}{0.957}&\textcolor{black}{0.990}\\
\textcolor{black}{Canny (Mid)} &\textcolor{black}{0.605}&\textcolor{black}{0.064}&\textcolor{black}{0.146}&\textcolor{black}{0.789}&\textcolor{black}{0.958}&\textcolor{black}{0.990}\\
\textcolor{black}{Canny (High)} &\textcolor{black}{0.605}&\textcolor{black}{0.064}&\textcolor{black}{0.147}&\textcolor{black}{0.790}&\textcolor{black}{0.957}&\textcolor{black}{0.990}\\ \bottomrule
\end{tabular}}
\end{table}

\begin{table}
\centering
\caption{Ablation studies regarding 3DGM on the NYUD-V2 dataset. {\ourmodelshortoriginal}: depth-to-normal and normal-to-depth modules. Full Model: all the modules with iterative inference.}
\label{tab:com-depth-to-norm}
\setlength{\tabcolsep}{2pt}
\scalebox{0.95}{
\begin{tabular}{c|ccc|ccc}
\toprule
\multirow{1}{*}{} &
\multicolumn{3}{c|}{{Error}}  &
\multicolumn{3}{c}{{Accuracy}} \\
& {{mean}} & {{median}} & {{rmse}} & {{$11.25^\circ$}} & {{$22.5^\circ$}} & {{$30^\circ$}} \\
\midrule
\midrule
Our Baseline  & 42.39 & 37.61 & 50.81 & 12.09 & 28.97 & 39.68 \\
With {\ourmodelshortoriginal}\cite{qi2018geonet}&35.02 & 29.12&43.33 &17.60 &39.04 &51.36\\
With Ensemble &35.16 & 28.78&43.82 &17.95 &39.58 &51.87\\
With Edge-aware &34.96&29.14&43.09&17.05&38.87&51.37\\
Full Model & \textbf{33.24} & \textbf{26.28} &\textbf{42.24} & \textbf{19.60} & \textbf{43.19} & \textbf{56.09} \\

\bottomrule
\end{tabular}}

\end{table}

\begin{table}
\centering
\caption{Performance evaluation of depth-to-normal on the NYUD-V2 test set. VGG stands for the VGG-16 network. ``LS'' means our least square module. ``D-N'' is our depth-to-normal network without the last $1 \times 1$ convolution layer. Ground-truth depth maps are used as input.}
\label{tab:geom-comp}
\setlength{\tabcolsep}{4pt}
\scalebox{1.0}{
\begin{tabular}{c|ccc|ccc}
\toprule
\multirow{1}{*}{} &
\multicolumn{3}{c|}{{Error}}  &
\multicolumn{3}{c}{{Accuracy}} \\
& {{mean}} & {{median}} & {{rmse}} & {{$11.25^\circ$}} & {{$22.5^\circ$}} & {{$30^\circ$}} \\
\midrule
\midrule
{4-layer} & {39.5} & {37.6} & {44.0} & {6.1} & {21.4} & {35.5} \\
{7-layer}  & {39.8} & {38.2} & {44.3} & {6.5} & {21.0} & {34.2} \\
{VGG} & {47.8} & {47.3} & {52.1} & {2.8} & {11.8} & {20.7} \\ \hline
{LS}& {11.5}&{6.4}&{18.8}&{70.0}&{86.7}&{91.3}\\
{D-N} & \textbf{8.2} & \textbf{3.0} & \textbf{15.5} & \textbf{80.0} & \textbf{90.3} & \textbf{93.5}\\
\bottomrule
\end{tabular}}

\end{table}
\section{CNNs and Geometric Constraints}\label{sect:naive-solution}

In this section, we verify our motivation by testing if CNNs can directly learn the mapping from depth to surface normal, \ie, implicitly learn the geometric constraints, so that the generated depth naturally produces high-quality surface reconstructions.
To this end, we train CNNs, which take the ground-truth depth and surface normal maps as inputs and supervision respectively.
We tried architectures including the first $4$ layers, the first $7$ layers, and the full version of VGG-16 network.
Before feeding to the above networks, the depth map is transformed into a $3$-channel image encoding $\{x, y, z\}$ coordinates respectively.

We provide the test performance on the NYUD-V2 dataset in Tab.~\ref{tab:geom-comp}.
All variants of CNNs converge to very poor local minima. 
We also show the test performance of the surface normal predicted by our depth-to-normal network.
In particular, since the depth-to-normal module contains least-square and residual modules, we also show the surface normal map predicted by the least square module {\it only} denoted as ``LS''.
Tab.~\ref{tab:geom-comp} reveals that the ``LS'' module alone is significantly better than the vanilla CNN baselines in all metrics. 
Moreover, with the residual module, the performance of our module gets further boosted.

These preliminary experiments lead to the following important findings:
\begin{enumerate}
	\item Learning a mapping from depth to normal directly via vanilla CNNs hardly respects the underlying geometric relation.
	\item Despite its simplicity, the least square module is very effective in incorporating geometric constraints into neural networks, thus achieving better performance.
	\item Our depth-to-normal network further improves the quality compared to the least-square module alone.
\end{enumerate}

\begin{table}
\centering
\caption{Depth-to-normal consistency evaluation on the NYUD-V2 test set.
``Pred'' means that we transform predicted depth to surface normal and compare it with the predicted surface normal.
``GT'' means that we transform predicted depth to surface normal and compare it with the ground-truth surface normal.
``Baseline'' and ``{\ourmodelshortoriginal}'' indicate that predictions are from baseline and our model respectively.
The backbone network of our baseline is VGG-16.
}
\label{tab:normal-consistency}
\setlength{\tabcolsep}{3.0pt}
\scalebox{1.0}{
\begin{tabular}{c|ccc|ccc}
\toprule
\multirow{1}{*}{} &
\multicolumn{3}{c|}{{Error}}  &
\multicolumn{3}{c}{{Accuracy}} \\
& {{mean}} & {{median}} & {{rmse}} & {{$11.25^\circ$}} & {{$22.5^\circ$}} & {{$30^\circ$}} \\
\midrule
\midrule
Pred-Baseline & 42.2 & 39.8 & 48.9 & 9.8 & 25.2 & 35.9 \\
Pred-{\ourmodelshortoriginal}  & \textbf{34.9} & \textbf{31.4} & \textbf{41.4} & \textbf{15.3} & \textbf{35.0} & \textbf{47.7} \\ \hline
GT-Baseline & 47.8 & 47.3 & 52.1 & 2.8 & 11.8 & 20.7 \\
GT-{\ourmodelshortoriginal} & \textbf{36.8} & \textbf{32.1} & \textbf{44.5} & \textbf{15.0} & \textbf{34.5} & \textbf{46.7} \\
\bottomrule
\end{tabular}}
\end{table}
\vspace{-0.1in}
\subsection{Geometric Consistency}

We verify if the depth and surface normal maps predicted by our geometric model, {\ie}, {\ourmodelshortoriginal}, are consistent.
To this end, we first train a standard depth-to-normal module using ground-truth depth and surface normal maps and regard it as an accurate transformation.
Given the predicted depth map, we compute the transformed surface normal map using this trained standard module.

With these preparations, we conduct experiments under the following $4$ settings.
(1) Comparison between transformed normal and predicted normal both generated by the baseline network.
(2) Comparison between transformed normal and predicted normal both generated by our {\ourmodelshortoriginal}.
(3) Comparison between transformed normal generated by the baseline network and the ground-truth normal.
(4) Comparison between transformed normal generated by~{\ourmodelshortoriginal} and the ground-truth normal.
Here we also use the VGG-16 network as the backbone network.

These results are shown in Tab.~\ref{tab:normal-consistency}.
The ``Pred'' columns of the table show that our {\ourmodelshort} can generate predictions of depth and surface normal which are more consistent than those of the baseline CNN. 
From the ``GT'' columns of the table, it is also obvious that compared to the baseline CNN, the predictions yielded from our {\ourmodelshort} are consistently closer to the ground-truth.

\section{Conclusion}\label{sect:conclusion}

We have proposed {\ourmodel} ({\ourmodelshort}) to jointly predict depth and surface normal from a single image. Our {\ourmodelshort} involves depth-to-normal and normal-to-depth modules. It effectively enforces the geometric constraints that the prediction should obey regarding depth and surface normal. 
They make the final prediction geometrically consistent and more accurate. 
The ensemble network slightly adjusts the results by fusing geometric refined predictions and initial predictions from backbone networks.
The edge-aware refinement network updates predictions in the planar and boundary regions.
The iterative inference is finally adopted to improve the prediction.
Our extensive experiments show that {\ourmodelshort} achieves state-of-the-art results in terms of both 2D metrics and a newly proposed 3D geometric metric.
In the future, we would like to apply our {\ourmodelshort} to geometric estimation tasks, such as 3D reconstruction, stereo matching, and SLAM.

\section*{Acknowledgements}

The work was supported in part by HKU Start-up Fund, Seed Fund for Basic Research, the ERC grant ERC-2012-AdG321162-HELIOS, EPSRC grant Seebibyte EP/M013774/1 and EPSRC/MURI grant EP/N019474/1. We would also like to thank the Royal Academy of Engineering and FiveAI. 
RL was supported by Connaught International Scholarship and RBC Fellowship.
\ifCLASSOPTIONcaptionsoff
  \newpage
\fi

\bibliographystyle{IEEEtran}
\bibliography{geo_net}

%

\begin{IEEEbiography} [{\includegraphics[width=1in,height=1.25in,clip,keepaspectratio]{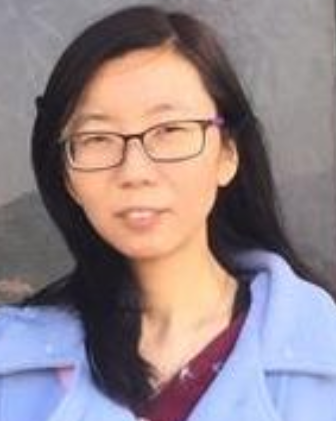}}]{Xiaojuan Qi} received her B.Eng degree in Electronic Science and Technology at Shanghai Jiao Tong University (SJTU) in 2014, and the Ph.D. degree in Computer Science and Engineering from the Chinese University of Hong Kong in 2018. She was a postdoc at the University of Oxford. She is now an assistant professor at the University of Hong Kong.
\end{IEEEbiography}
\vspace{-0.4in}
\begin{IEEEbiography} [{\includegraphics[width=1in,height=1.25in,clip,keepaspectratio]{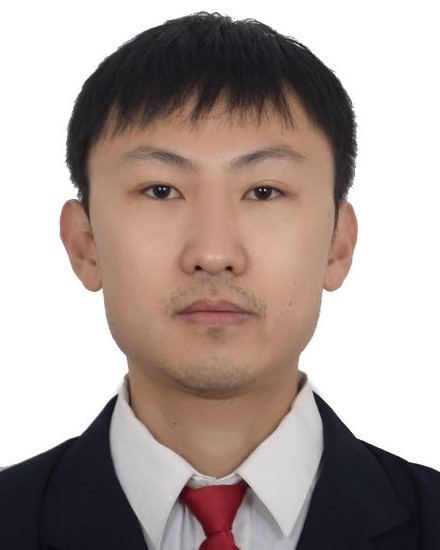}}]{Zhengzhe Liu} received his B.Eng degree in Information Engineering at Shanghai Jiao Tong University (SJTU) in 2014, and the MPhil degree in Computer Science and Engineering from the Chinese University of Hong Kong in 2017. 
\end{IEEEbiography}
\vspace{-0.4in}
\begin{IEEEbiography} [{\includegraphics[width=1in,height=1.25in,clip,keepaspectratio]{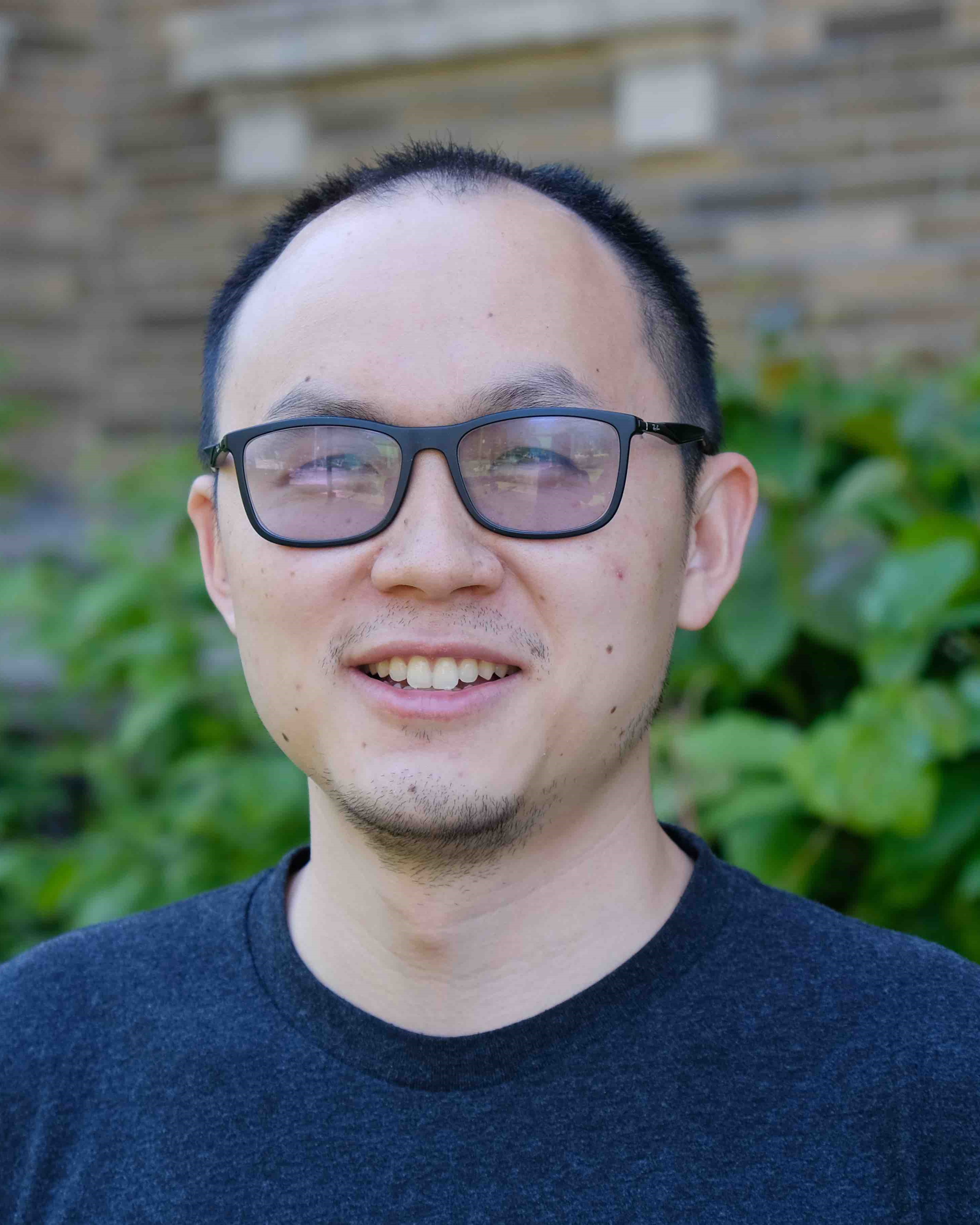}}]{Renjie Liao} received his B.Eng degree from School of Automation Science and Electrical Engineering at Beihang University (former Beijing University of Aeronautics and Astronautics), and an  MPhil degree in Computer Science and Engineering from the Chinese University of Hong Kong. He is now a Ph.D. student in the Department of Computer Science, University of Toronto.  
\end{IEEEbiography}
\vspace{-0.4in}
\begin{IEEEbiography} [{\includegraphics[width=1in,height=1.25in,clip,keepaspectratio]{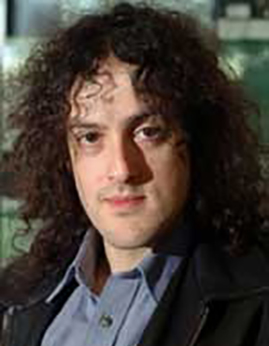}}]{Philip H.S. Torr}received his Ph.D. degree from Oxford University. After working for another three years at  Oxford as a research fellow,  he worked for six years in Microsoft  Research,  first in  Redmond,  then in Cambridge, founding the vision side of the Machine Learning and Perception Group. He then became a Professor in Computer Vision and Machine Learning at Oxford Brookes University. He is now a  professor at  Oxford  University. He is a fellow of the Royal Academy engineering and an Ellis Fellow.
\end{IEEEbiography}
\vspace{-0.4in}
\begin{IEEEbiography}
[{\includegraphics[width=1in,height=1.25in,clip,keepaspectratio]{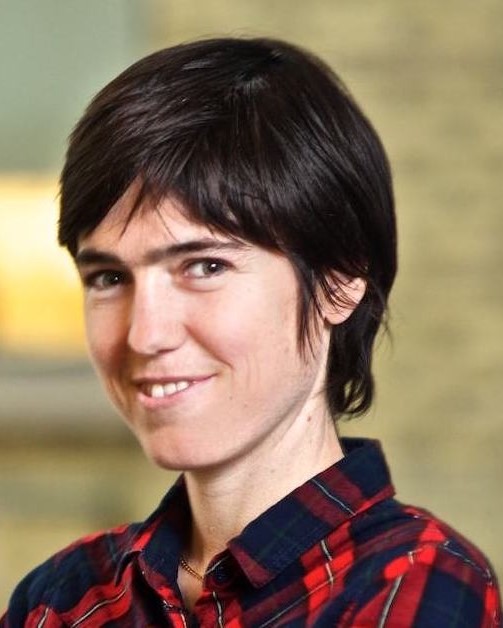}}]{Raquel Urtasun} her Ph.D. degree from the Computer Science department at Ecole Polytechnique Federal de Lausanne (EPFL) in 2006 and did her postdoc at MIT and UC Berkeley. She was also a visiting professor at ETH Zurich during the spring semester of 2010. Then, she was an Assistant Professor at the Toyota Technological Institute at Chicago (TTIC). She is currently Uber ATG Chief Scientist and the Head of Uber ATG Toronto. She is also an Associate Professor in the Department of Computer Science at the University of Toronto, a Canada Research Chair in Machine Learning and Computer Vision, and a co-founder of the Vector Institute for AI.
\end{IEEEbiography}
\vspace{-0.4in}
\begin{IEEEbiography} [{\includegraphics[width=1in,height=1.25in,clip,keepaspectratio]{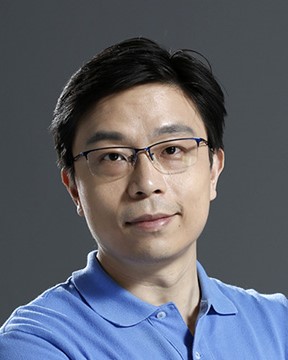}}]{Jiaya Jia} received his Ph.D. degree in Computer Science from Hong Kong University of Science and Technology in 2004.  From March 2003 to August 2004, he was a visiting scholar at Microsoft. Then, he joined the Department of Computer Science and Engineering at The Chinese University of Hong Kong (CUHK) in 2004 as an assistant professor and was promoted to Associate Professor in 2010. He conducted collaborative research at Adobe Research in 2007. He was promoted to Professor in 2015. He was the Distinguished Scientist and Founding Executive Director of Tencent YouTu X-Lab. He is an IEEE Fellow.
\end{IEEEbiography}

\end{document}